\documentclass{article}

     \usepackage[preprint,nonatbib]{neurips_2024}

\usepackage{microtype}
\usepackage{graphicx}
\usepackage{subfigure}
\usepackage{booktabs} % for professional tables

\usepackage{hyperref}

\usepackage[utf8]{inputenc} % allow utf-8 input
\usepackage[T1]{fontenc}    % use 8-bit T1 fonts
\usepackage{hyperref}       % hyperlinks
\usepackage{url}            % simple URL typesetting
\usepackage{booktabs}       % professional-quality tables
\usepackage{amsfonts}       % blackboard math symbols
\usepackage{nicefrac}       % compact symbols for 1/2, etc.
\usepackage{microtype}      % microtypography
\usepackage{xcolor}         % colors

\usepackage{amsmath}
\usepackage{amssymb}
\usepackage{mathtools}
\usepackage{amsthm}

\usepackage{wrapfig}
\usepackage{algorithm}
\usepackage{algorithmic}
\usepackage[capitalize,noabbrev]{cleveref}

\theoremstyle{plain}

\theoremstyle{definition}

\theoremstyle{remark}

\usepackage[textsize=tiny]{todonotes}

\usepackage{amssymb}
\newcommand{\Pre}{\textrm{PRE}}
\newcommand{\Add}{\textrm{ADD}}
\newcommand{\Del}{\textrm{DEL}}

\newcommand{\ignore}[1]{{}}
\newcommand{\ours}{{\tt LLMs4Plan}}

\title{On the Roles of LLMs in Planning: Embedding LLMs into Planning Graphs}

\author{%
  Hankz Hankui Zhuo\thanks{Corresponding author. homepage: https://www.xplan-lab.org} \\
  Sun Yat-sen University, Guangzhou, China \\
  \texttt{zhuohank@gmail.com} \\
   \And
   Xin Chen \\
  Sun Yat-sen University, Guangzhou, China \\
  \texttt{chenx287@mail2.sysu.edu.cn} \\
  \AND
  Rong Pan \\
  Sun Yat-sen University, Guangzhou, China \\
   \texttt{panr@sysu.edu.cn} \\
}

\begin{document}

\maketitle

\begin{abstract}
Plan synthesis aims to generate a course of actions or policies to transit given initial states to goal states, provided domain models that could be designed by experts or learnt from training data or interactions with the world.
Intrigued by the claims of emergent planning capabilities in large language models (LLMs), works have been proposed to investigate the planning effectiveness of LLMs, without considering any utilization of off-the-shelf planning techniques in LLMs. 
In this paper, we aim to further study the insight of the planning capability of LLMs by investigating the roles of LLMs in off-the-shelf planning frameworks. To do this, we investigate the effectiveness of embedding LLMs into one of the well-known planning frameworks, graph-based planning, proposing a novel LLMs-based planning framework with LLMs embedded in two levels of planning graphs, i.e., mutual constraints generation level and constraints solving level. We empirically exhibit the effectiveness of our proposed framework in various planning domains. 
\end{abstract}

\section{Introduction}
Plan synthesis aims to generate a course of actions or policies to transit given initial states to goal states, provided domain models that could be designed by experts or learnt from training data \cite{DBLP:journals/ai/AinetoCO19} or interactions with the world \cite{DBLP:conf/ijcai/LamannaFGSSST23,DBLP:conf/aaai/JinMJZCY22}. It is a time- and space-consuming open issue in the planning community \cite{book/Ghallab04}.
Intrigued by the claims of emergent planning capabilities in large language models (LLMs), works have been proposed to investigate the planning effectiveness of LLMs, without considering any utilization of off-the-shelf planning techniques in LLMs \cite{neurIPS23-Valmeekam}. As demonstrated by \cite{neurIPS23-Valmeekam}, even in a seemingly simple common-sense domain like Blocksworld that humans usually find easy to solve, LLMs are evaluated to be quite ineffective in planning autonomously.

An interesting result shown by \cite{neurIPS23-Valmeekam} is when taking the solution generated by LLMs, which is incorrect, as a seed plan to be repaired by an off-the-shelf planner, e.g., LPG \cite{DBLP:conf/aips/GereviniS02}, a significant improvement in search steps can be attained over the result when an empty plan provided as a seed plan for the planner. This indicates that LLMs can indeed provide some helpful information (e.g., in some sense of heuristics) for planning, even though they cannot solve planning problems solely. Inspired by the result of loosely using plans generated by LLMs as seed plans, we are curious if it is possible to ``dig'' more helpful information from LLMs to assist planning deeply, e.g., by inserting LLMs into planning frameworks. By doing this, we aim to answer the question: \textbf{what roles can be played exactly by LLMs in planning?} Indeed, there have been attempts to explore off-the-shelf planning techniques to help LLMs solving planning problems \cite{DBLP:journals/corr/abs-2304-11477}. Similar to \cite{neurIPS23-Valmeekam}, they only view planners as black-boxes without deepening the integration of LLMs in planning frameworks. 

%Specifically, we aim to further study the insight of the planning capability of LLMs by investigating the roles of LLMs in off-the-shelf planning frameworks. 
To do this, we investigate the effectiveness of embedding LLMs into one of the well-known planning frameworks, graph-based planning \cite{DBLP:journals/ai/BlumF97}. We propose a novel LLMs-based planning framework with LLMs embedded in two phases of the planning framework (namely \ours). The first phase is to propose promising actions in ``action-levels'' of the planning graph using LLMs. The second phase is to propose non-mutual action sets using LLMs when backtracking the planning graph. Note that the two phases correspond to two critical steps that influence the efficiency and effectiveness in graph planning. 

\begin{wrapfigure}{r}{0.5\textwidth}
    %\centering
    %\subfigure[Expantion]{
\includegraphics[width=0.5\textwidth]{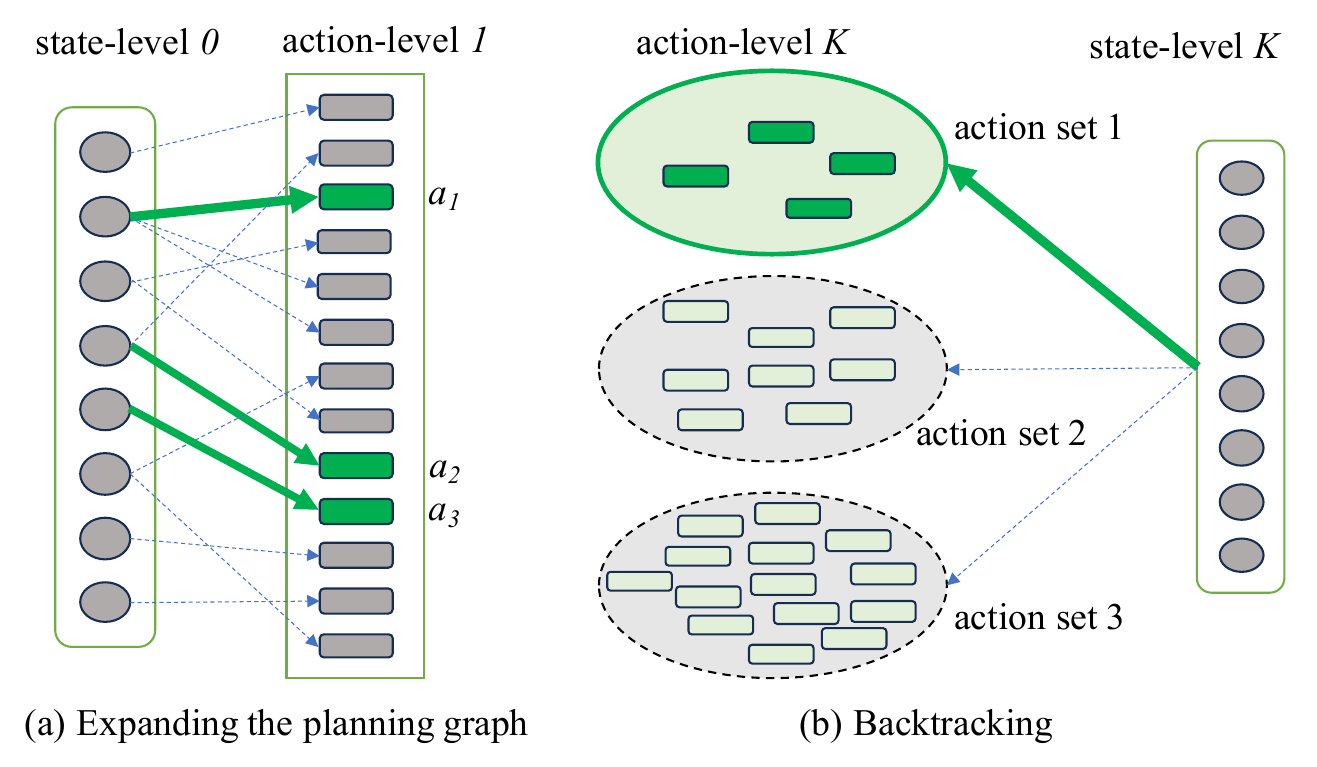}
%\caption{}
%}
%    \subfigure[Backtracking]{
%    \label{fig:motivation2}
%\includegraphics[width=0.25\textwidth]{motivation2.pdf}
%\caption{}
%}
    \caption{Two critical steps in graph planning}
    \label{fig:motivation}
\end{wrapfigure}
\emph{For example, as shown in Figure \ref{fig:motivation}(a), there could be a large number of actions in ``action-level 1'' when expanding ``state-label 0'' with Graphplan \cite{DBLP:journals/ai/BlumF97}. We aim to exploit LLMs to help select a small subset of promising actions, e.g., $\{a_1,a_2,a_3\}$ are selected in Figure \ref{fig:motivation}(a). In Figure \ref{fig:motivation}(b), when backtracking from ``state-level $K$'' that includes goals, there could be a large number of candidate sets of actions to be explored (e.g., ``action set 1'', ``action set 2'', ``action set 3'') --- actions in each candidate set are not mutually exclusive with each other (two actions are mutually \textbf{exclusive} if they are not allowed to be executed at the same time, e.g., actions ``pick up object A'' and ``put down object A'' are mutually exclusive). It is particularly time-consuming to search all of the valid candidate sets based on \emph{mutual constraints} in each action-level. We expect LLMs are capable of selecting a small number of candidate sets to be backtracked, e.g., only ``action set 1'' is selected to be backtracked by LLMs as shown in Figure \ref{fig:motivation}(b).}

Specifically, in {\ours}, for each action-level $i$, we first automatically generate prompts based on \emph{propositions} in state-level $i-1$, \emph{goals} and \emph{domain models}, and feed the prompts to LLMs to select actions for generating action-level $i$. After that, when backtracking from the last state-level $K$ in the expanded planning graph, which includes the goal, we automatically generate prompts based on propositions in state-level $K$, propositions in the initial state (i.e., state-level 0), mutual constraints in action-level $K$, and \emph{domain models}, and feed the prompts to LLMs to select action sets for backtracking. We embed the above two components into one of the well-known off-the-shelf graph planners, Graphplan \cite{DBLP:journals/ai/BlumF97}. We study the effectiveness of different cases of adding or removing the above one or two components in Graphplan to see the significance of roles LLMs play in the graph planning framework. 

Through this study, we provide new clues for how to \emph{deeply} embed LLMs into off-the-shelf planning frameworks, i.e., first identifying critical steps (generally time-consuming ones) in specific planning frameworks, and then designing proper prompt generation to be embedded into the frameworks. We verify that soly relying on LLMs to do planning is far from a good option, while leveraging LLMs to help deal with some critical steps in the graph planning framework is possible.

\section{Related work}
\textbf{LLMs as planners:}
%LLMs have been used as a few-shot policy for language-conditioned task planning \cite{DBLP:conf/icml/HuangAPM22}, or a policy for multi-step reasoning problems \cite{DBLP:conf/iclr/ShiSF0SVCTRZ0W23,DBLP:journals/corr/abs-2305-10601}. However, recent research \cite{DBLP:journals/corr/abs-2305-18654} also suggests that transformer LLMs are inherently limited for solving multi-step reasoning problems.
There have beem works leveraging LLMs to assisting planning tasks, such as Chain-of-Thoughts \cite{DBLP:conf/nips/Wei0SBIXCLZ22}, Tree-of-Thoughts \cite{DBLP:journals/corr/abs-2305-10601} and Zero-Shot Planner \cite{DBLP:conf/icml/HuangAPM22}, which utilize prompts to guide LLMs for generating action sequences for planning related tasks. Approaches such as HuggingGPT \cite{DBLP:journals/corr/abs-2303-17580} and Chameleon \cite{DBLP:journals/corr/abs-2304-09842} aim to generate initial plans using different tools and then call the corresponding APIs for execution by augmenting LLMs with plug-and-play modules for compositional reasoning. They designed an LLM-based planner to assemble a sequence of tools to execute to generate final solution plans for planning tasks.
There are also some works that prompt LLMs to compose plans in the form of PDDL (Planning Domain Definition Language) in planning community \cite{DBLP:journals/corr/abs-2304-11477,DBLP:conf/ijcai/PallaganiMSRHML23}. All of the above-mentioned approaches generally synthesize plans without consideration of feedback from external environments.
In order to consider environmental feedback, SayCan \cite{DBLP:conf/corl/IchterBCFHHHIIJ22}, ReAct \cite{DBLP:conf/iclr/YaoZYDSN023}, Reflexion \cite{DBLP:journals/corr/abs-2303-11366} and Inner Monologue \cite{DBLP:conf/corl/HuangXXCLFZTMCS22} allow LLMs to take single-step actions according to the environmental feedback, assuming LLM-generated initial plans are correct without any adaptation of them. They only adjust the immediate action being executed and are prone to fall into local sub-optimal actions without considering long-term goals. In consideration of removing this assumption, the evaluation of planning capabilities in the literature involves the user incrementally interacting with LLMs, and re-prompting it to point out flaws in its plans, with the hope that the LLM eventually reaches an executable plan \cite{DBLP:conf/corl/HuangXXCLFZTMCS22,DBLP:conf/iclr/YaoZYDSN023,DBLP:journals/corr/abs-2211-09935,DBLP:journals/corr/abs-2302-01560,DBLP:conf/nips/Sun23,DBLP:journals/corr/abs-2305-11014}. Such evaluations are notoriously with the actual planning being done by humans in the loop rather than LLMs themselves. Instead of considering human-in-the-loop,
\cite{neurIPS23-Valmeekam} built another framework for automatically evaluating planning capabilities of LLMs by leveraging automated planning models and tools to generate the queries and validate answers from LLMs. All of the above-mentioned approaches view LLMs as a sole planner rather than being leveraged as components and embedded into off-the-shelf planning frameworks, which means the planning capability of planning frameworks is not leveraged by them. 

\ignore{
\textbf{LLMs finetuned to be Planners:}
%\cite{DBLP:journals/corr/abs-2305-14314} uses a novel high-precision technique to quantize a pretrained model to 4-bit, then adds a small set of learnable Low-rank Adapter weights that are tuned by backpropagating gradients through the quantized weights.
Another issue of broad interest in LLMs is to investigate how to finetune LLMs to be capable of planning. Some works \cite{DBLP:journals/tmlr/ReedZPCNBGSKSEBREHCHVBF22,DBLP:conf/rss/BrohanBCCDFGHHH23} collected paired visual, language and action data and trained an enormous neural networks for tackling long-horizon tasks. Collecting those data, however, is often expensive and difficult to scale up. Different from training LLMs from scratch, finetuning off-the-shelf LLMs \cite{DBLP:conf/icml/DriessXSLCIWTVY23,DBLP:conf/nips/LiPPDWF0HAAAM0Z22} to deal with visual and language related tasks using task-specific robot demonstrations is an option to alleviate issues of data collection. It is, however, still problematic since finetuning most high-performing language models, such as GPT3.5/4 \cite{DBLP:conf/nips/Ouyang0JAWMZASR22} and PaLM \cite{DBLP:journals/jmlr/Chowd}, is currently impossible, as the model weights are not open-sourced. Besides, paired vision and language robotics demonstrations, which are still needed for finetuning, are not readily available in practice. Considering this, \cite{DBLP:journals/corr/abs-2309-08587} propose compositional foundation models for hierarchical planning, where a foundation model is a composition of different expert models trained on language, vision, and action data individually. Given an abstract language instruction describing the desired task, they use a large language model to generate a sequence of sub-tasks and then use a large video diffusion model to capture geometric and physical information about the world and generate a more detailed plan. Finetuning LLMs to be a planner indicates training data are required to be available and the planning capability of finetuned LLMs to be transferred to other tasks is still an open issue. Instead of finetuning LLMs, we aim to leverage both benefits of LLMs and planning frameworks to solve planning tasks without any training data.
}

\textbf{LLMs as components in planners:}
%There have been efforts which mostly depended on LLMs as “translators" of natural language problem/goal specification into formal specifications, which are then thrown over to external planners \cite{DBLP:journals/corr/abs-2302-05128,DBLP:journals/corr/abs-2304-11477}. Such efforts do not shed any light on the internal planning capabilities of the LLMs themselves. %Investigating whether GPT-4 \cite{DBLP:journals/corr/abs-2303-08774} can be used to write a domain-specific Python program that solves a set of tasks in a planning domain \cite{DBLP:journals/corr/abs-2305-11014}.
%\cite{neurIPS23-Guan} investigated two use cases of the generated PDDL action models for downstream planning tasks. For one, by utilizing an LLM to translate user instructions into goal specifications in PDDL [58, 30], we can use any standard domain-independent planner to search for a plan. On the other hand, the extracted PDDL model can be used to validate plans suggested by an LLM planner and to provide corrective feedback in the form of unmet preconditions or goal conditions. In this case, the PDDL model is essentially serving as an inexpensive high-level simulator or a human proxy to ensure plan correctness.
There are also works using LLMs as heuristics \cite{DBLP:journals/corr/abs-2303-05510} or transition function \cite{DBLP:conf/emnlp/HaoGMHWWH23} in MCTS, boosting the performance in coding or small-scale reasoning. %Utilizing LLMs as a world model in depth, however, is not discussed \cite{DBLP:journals/corr/abs-2305-14078}.
LLMs can also provide a commonsense model of the world in addition to a policy that acts on it. The world model and the policy can be combined into Monte Carlo Tree Search (MCTS) to scale up task planning \cite{DBLP:journals/corr/abs-2305-14078}, which is demonstrated to be effective by embedding LLMs in the MCTS framework. The idea is similar to our work in the sense that LLMs are embedded into an off-the-shelf framework in depth. However, they aim to learn MCTS policies from interactions with environments, while our work aims to solve planning problems with LLMs embedded into an off-the-shelf planner, planning-graph planner, without any interactions or learning from environments. \cite{neurIPS23-Guan} investigated effectiveness of the generated PDDL action models from LLMs for downstream planning tasks. Likewise, it is different from our work in the sense that we assume the PDDL action models are already known and focus on exploring LLMs to speed up the planning efficiency. 
%}
\section{Problem Formulation}
In this work we consider classical planning problems specified in the form of STRIPS \cite{DBLP:journals/ai/FikesN71}. Similar ideas can be extended into more expressive planning language such as PDDL \cite{PDDL}. Let $\mathcal{L}$ be a set of atoms, each of which is composed of a predicate with zero or more parameters (e.g., \emph{clean(room)} is an atom indicating \emph{room} is \emph{clean}).
A STRIPS domain is composed of a set of action models $\mathcal{A}$, each of which is a quadruple
$\langle a,\Pre(a), \Add(a), \Del(a) \rangle$, where $a$ is an action
name with zero or more parameters, $\Pre(a)\subseteq \mathcal{L}$ is a precondition list
indicating the conditions under which $a$ can be applied, $\Add(a)\subseteq \mathcal{L}$ is
an adding list and $\Del(a)\subseteq \mathcal{L}$ is a deleting list indicating the effects
of $a$. Let $\mathcal{R}$ be a set of propositions, which are instances of atoms in $\mathcal{L}$. We define a planning problem as
$\mathcal{P} = \langle \mathcal{R}, s_0, g, \mathcal{A} \rangle$, where $s_0 \subseteq
\mathcal{R}$ is an initial state and $g \subseteq \mathcal{R}$ is a
goal. A solution $\pi$ to the planning problem is a sequence of actions that transit initial state $s_0$ to goal $g$.
\ignore{
A solution plan $\pi=\langle A, C\rangle$ to problem $\mathcal{P}$ consists of a set of actions $A$ and a set of partial order constraints $C$ among actions $A$. Note that some actions in $A$ are allowed to be executed in parallel so long as there are no \emph{conflicts} among them. Intuitively, we say there are \emph{conflicts} between two actions if one deletes a precondition or an adding effect of the other (more detailed description of emph{conflicts} will be given in Section 4.2). A partial order plan can be ``stretched'' into a many total order linear plans, each of which can successfully transit initial state $s_0$ to goal $g$.
}
An intuitive example of our planning problem is as shown below.\\
%=========== \emph{Beginning of the example} ============\\
%\textbf{\emph{Example 1:}}
\emph{Suppose we would like to clean a bedroom using a vacuum which is placed in a tool room. We can formulate the problem $\mathcal{P}=\langle \mathcal{R}, s_0,g,\mathcal{A}\rangle$ in the form of STRIPS (note that we assume there is no parameter for each predicate and action for simplicity since there is only one tool, one bedroom and one toolroom). 
The set of propositions $\mathcal{R}$ is represented by $\mathcal{R}=\{dirty(),toolroom(),clean(),bedroom()\}$.
Initial state $s_0$ is represented by $s_0$ = \{dirty(), toolroom()\}, which indicates the ``bedroom'' is dirty, and the tool ``vacuum'' is in the tool room (i.e., ``toolroom''). 
The goal $g$ is represented by $g=\{clean(), toolroom()\}$, which indicates the ``bedroom'' is clean, and the tool ``vacuum'' is back to the tool room. 
The set of action models $\mathcal{A}$ is represented as follows: 
}
\begin{center}
    \begin{tabular}{|l l l|}
    \hline\hline
        Action & Preconditions & Effects \\
        \hline
        $vacuum()$ & $dirty()$, $bedroom()$ & $clean()$, $\neg dirty()$ \\
       % & $bedroom()$ &  \\
        \hline
        $move2tr()$ & $bedroom()$ & $toolroom()$, $\neg bedroom()$ \\
        % & &  \\
        \hline
        $move2br()$ & $toolroom()$ & 	  $bedroom()$, $\neg toolroom()$ \\
        %& &  \\
        \hline
        \hline
    \end{tabular} 
\end{center}
\emph{Action $vacuum()$ aims to vacuuming ``bedroom'', the preconditions of which are ``bedroom'' is dirty and the vacuum-cleaner is in ``bedroom''. The effects of $vacuum()$ are adding $clean()$ to the state where $vacuum()$ is executed, indicating ``bedroom'' is clean, and deleting $dirty()$ (i.e., $\neg dirty()$) from the state, indicating ``bedroom'' is not dirty anymore. Action $move2tr()$ aims to move the vacuum-cleaner to ``toolroom'', the precondition of which is $bedroom()$ indicating the vacuum-cleaner is in ``bedroom''. The effects are adding $toolroom()$ indicating the vacuum-cleaner is in ``toolroom'', and deleting $bedroom()$, indicating the vacuum-cleaner is not in ``bedroom''. Similarly, action $move2br$ aims to move the vacuum-cleaner to ``bedroom'', the precondition of which is $toolroom()$, indicating the vacuum-cleaner is in ``toolroom''. The effects are the vacuum-cleaner is adding $bedroom()$, indicating the vacuum-cleaner is in ``bedroom'', deleting $toolroom()$, indicating the vacuum-cleaner is not in ``toolroom''.
}
A solution $\pi$ to the problem $\mathcal{P}$ is $move2br(),vacuum(),move2tr()$. 

\section{Our {\ours} approach} 
An overview of our {\ours} approach is shown in Algorithm \ref{lmm4planning}. In Step 3, the pruning possibility $\kappa_i$ is decreased as the exponent $i$ increasing. In Step 5, we expand planning graph $PG^r$ with one more level using LLMs to prune actions based on pruning possibility $\kappa_i$ and planning problem $\mathcal{P}$. In Steps 7, if goal $g$ is not included by the last state-level in $PG^r$, i.e., $Satisfied(g,PG^r)$ is false, we continue to Step 4. In Step 8, we build a set of mutual constraints $\mathcal{C}$ based on $PG^r$, i.e., $buildConstraints(PG^r)$. In Step 9, we sort sets of actions based on constraints $\mathcal{C}$ using LLMs, i.e., $sortActionsLLMs(PG^r,\mathcal{C}$. In Step 10, we search solution $\pi$ based on the sorted action sets $\mathbb{A}$ using depth-first search. 
\ignore{
If the number of levels in $PG$ is less than the maximal levels $MaxIter$, we build mutual constraints $\mathcal{C}$ and sort the candidate sets of actions, resulting in the list of candidate sets $\mathbb{A}$ with the help of LLMs, each of which satisfies $\mathcal{C}$. After that we make a depth-first search based on $\mathbb{A}$ to get $\pi$. If $\pi$ is not \emph{Failure}, it is returned as the solution plan; otherwise, we expand $PG$ with one more level using LLMs and repeat Steps 2-11 of the algorithm until the maximal level $MaxIter$ is reached or a solution plan $\pi$ is found and returned in Step 7.
} 
In the following subsections, we will address our {\ours} in detail.
% 关于完备性方面，我们实验发现，从深层开始放弃剪枝比从浅层开始放弃剪枝效果会更好。
\begin{algorithm}
    \caption{An overview of our {\ours}}\label{lmm4planning}
    \textbf{Input:} Planning problem $\mathcal{P}$, pruning possibility $\kappa_0$ \\
    \textbf{Output:} Solution $\pi$
    \begin{algorithmic}[1]
        \STATE $PG^r=\emptyset$
        \FOR{$i=1$ to $N$}
            \STATE $\kappa_i=(\kappa_0)^i$, $k=1$
            \WHILE{$k < K$} 
                \STATE $PG^r\leftarrow expandGraphLLMs(PG^r,\mathcal{P},\kappa_i)$
                \STATE $k=k+1$
                \STATE if {$Satisfied(g,PG^r) = false$}, then \textbf{continue}
                \STATE $\mathcal{C}=buildConstraints(PG^r)$
                \STATE $\mathbb{A}= sortActionsLLMs(PG^r, \mathcal{C})$
                \STATE $\pi = depthFirstSearch(\mathbb{A},PG^r)$
                \STATE if {$\pi\neq$ Failure},
                    then \textbf{return} $\pi$
            \ENDWHILE
        \ENDFOR
        \STATE \textbf{return} Failure
    \end{algorithmic}
\end{algorithm}
\ignore{
\begin{algorithm}[!ht]
    \caption{An overview of our {\ours} approach}\label{graphplan}
    \textbf{Input: } Planning problem $\mathcal{P}$ \\
    \textbf{Output: } Solution $\pi$ 
    \begin{algorithmic}[1]
        \STATE Build graph: $(PG, k) =buildGraphLLMs(\mathcal{P})$
        \WHILE{$k < K$}
            \STATE Build constraints: $\mathcal{C}=buildConstraints(PG)$
            \STATE Sort action sets: $\mathbb{A}= sortActionsLLMs(PG, \mathcal{C})$
            \STATE Search solutions: $\pi = depthFirstSearch(\mathbb{A},PG)$
            \IF{$\pi\neq$ Failure}
                \STATE \textbf{return} $\pi$
            \ENDIF
            \STATE Expand graph: $PG\leftarrow expandGraphLLMs(PG)$
            \STATE $k \leftarrow k+1$
        \ENDWHILE
        \STATE \textbf{return} Failure
    \end{algorithmic}
\end{algorithm}
}
\ignore{
\textbf{Theorem 1:} 
If planning problem $\mathcal{P}$ is $K$-step solvable, our {\ours} is capable of searching the solution to $\mathcal{P}$, i.e., our {\ours} is $K$-step \emph{complete}.\\
\emph{Proof. }
}
\subsection{Building Planning Graphs with LLMs}
\begin{figure}[!ht]
    \centering
    %\vspace{-em}
    \includegraphics[width=0.85\textwidth]{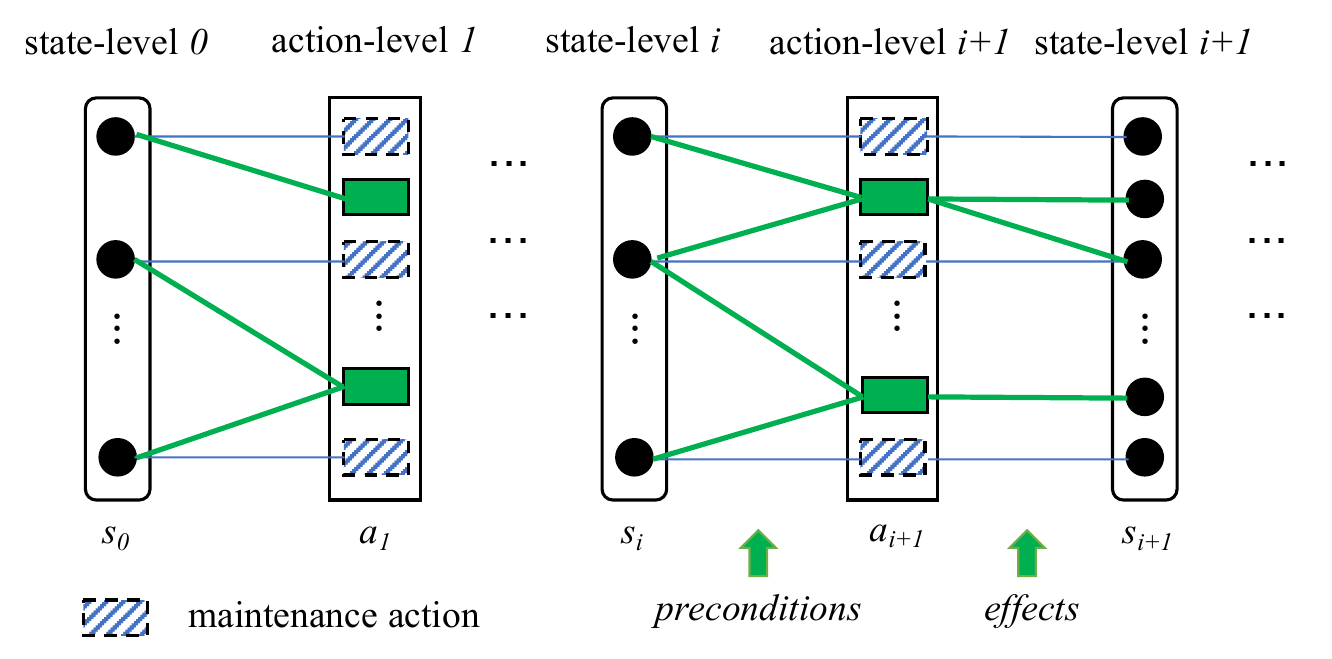}
    \caption{The framework of a planning graph}
    \label{fig:plangraph}
    %\vspace{-2em}
%\end{wrapfigure}
\end{figure}

A planning graph $PG^r$ is the search space for a relaxed version of the planning problem, an intuitive framework of which is shown in Figure \ref{fig:plangraph}. It alternates layers of ground literals and actions. ``Square'' nodes at action-level $i+1$ indicate actions that might be possible to be executed in state $s_i$. Maintenance actions indicate dump operators that keep literals unchanged between state-levels $i$ and $i+1$. ``Black circle'' nodes at state-level $i$ indicate literals that might possibly be true at time $i$. Edges between state-level $i$ and action-level $i$ indicate literals in state-level $i$ are preconditions of actions in action-level $i$, while edges between action-level $i$ and state-level $i+1$ indicate literals in state-level $i+1$ are adding or deleting effects of actions in action-level $i$. The nodes in the first state-level indicate literals that are true in initial state $s_0$. 

%\vspace{-em}
%\begin{wrapfigure}{r}{0.6\textwidth}
The procedure of building the planning graph with LLMs (i.e., \emph{buildGraphLLMs}) based on the given planning problem $\mathcal{P}=\langle \mathcal{R}, s_0, g, \mathcal{A}\rangle$ is as follows:
\begin{enumerate}
    \item All propositions in $s_0$ and negation of propositions in $\mathcal{R}-s_0$ are added into state-level 0.
    \item All actions in $\mathcal{A}$, whose preconditions are satisfied in state-level 0 and \textbf{selected by LLMs}, are added into action-level 1; a maintenance action corresponding to each proposition in state-level 0 is added into action-level 1.
    \item The propositions added or deleted by actions in action-level 1 are added into state-level 1; and all propositions in state-level 0 are added into state-level 1 as well (i.e., which is done by the maintenance action). 
    \item We repeat steps 1-3 by increasing state-level 0 to 1 (or $i$ to $i+1$) until all propositions in goal $g$ are included by state-level $k$. 
\end{enumerate}
\begin{wrapfigure}{r}{0.5\textwidth}
    \centering
    \includegraphics[width=0.45\textwidth]{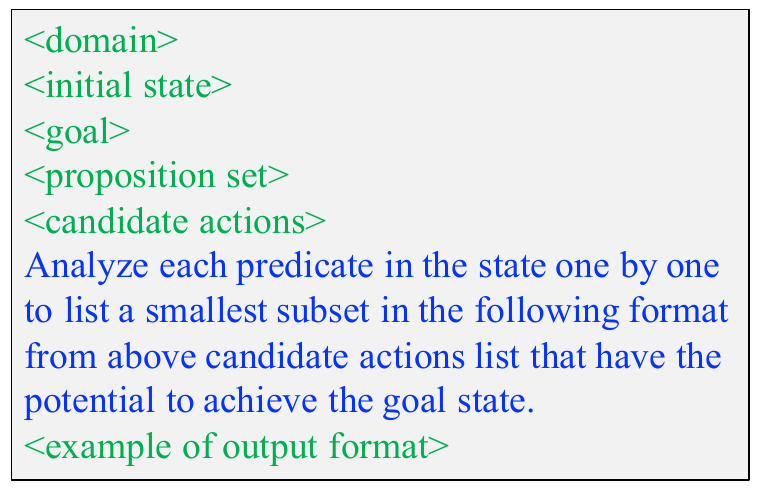}
    \caption{The prompt for pruning actions}
    \label{fig:prompt1}
    \vspace{-3em}
\end{wrapfigure}
In Step 5, we use LLMs to help select actions to build the planning graph. Note that in classical graph-based planning \cite{DBLP:journals/ai/BlumF97}, all of the actions whose preconditions are satisfied will be added into the action-level. 

We design the prompt to consult LLMs as shown in Figure \ref{fig:prompt1}, where ``$\langle$domain$\rangle$", ``$\langle$initial state$\rangle$", ``$\langle$goal$\rangle$", ``$\langle$proposition set$\rangle$", and "$\langle$candidate actions$\rangle$" are action models $\mathcal{A}$, initial state $s_0$, goal $g$, the set of propositions $\mathcal{R}$ and all of the candidate actions whose preconditions are satisfied in $s_0$. The text in BLUE is the prompt used to guide LLMs to select actions. ``$\langle$example of output format$\rangle$'' is used to guide LLMs to output actions in the desired format, e.g., ``move {'?from': 'rooma', '?to': 'roomb'}".

\subsection{Building Mutual Constraints}
\begin{figure}[!ht]
    \centering
    \includegraphics[width=0.8\textwidth]{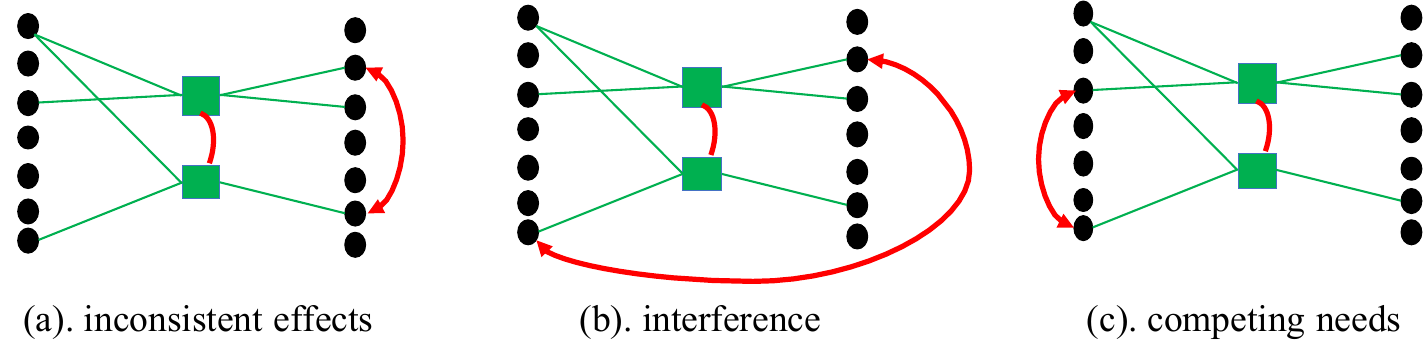}
    \caption{Mutual exclusion of actions}
    \label{fig:mutex}
\end{figure}

Due to the satisfaction of action models being relaxed, actions and/or states in action-levels or state-labels may be inconsistent, i.e., there may be some actions mutually exclusive in action-levels, or some literals mutually exclusive in state-levels. As shown in Figure \ref{fig:mutex}, there are three types of mutual exclusion constraints among actions.
Specifically, two actions at the same action-level are mutex, if they satisfy the following conditions:
\begin{itemize}
    \item An effect of one negates an effect of the other, which is called \emph{inconsistent effects}. 
    \item One deletes a precondition of the other, which is called \emph{interference}.
    \item They have mutually exclusive preconditions, which is called \emph{Competing needs}.
\end{itemize}
Otherwise they do not interfere with each other, i.e., both may appear in a solution plan. 
\ignore{
\begin{figure}[!ht]
    \centering
    \includegraphics[width=0.25\textwidth]{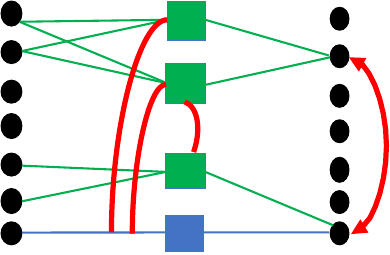}
    \caption{Inconsistent support for propositions}
    \label{fig:incon-sup}
\end{figure}
}
Two literals at the same state-level are mutex if one is the negation of the other, or all ways of achieving them are pairwise mutex, namely \emph{inconsistent support}.

\begin{figure}[!ht]
    \centering
    \includegraphics[width=0.8\textwidth]{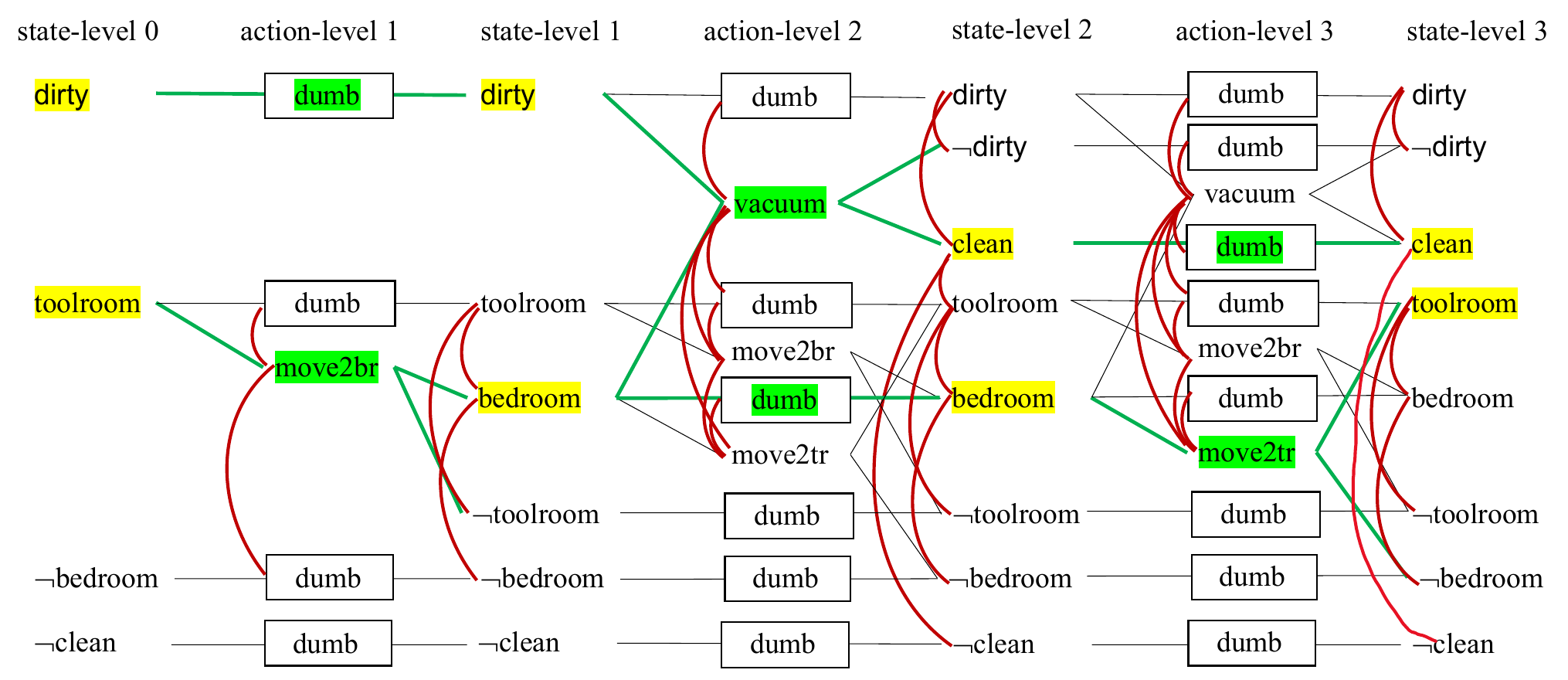}
    \caption{An example of planning graph and mutual constraints indicated in RED arcs}
    \label{fig:plangraph-example}
\end{figure}

An example planning graph corresponding to Example 1 is as shown in Figure \ref{fig:plangraph-example}. 
Action \emph{vacuum} is mutually exclusive with action \emph{dumb} for \emph{toolroom} at \emph{action-level 2} since \emph{vacuum}'s precondition \emph{bedroom} is mutually exclusive with \emph{toolroom} at \emph{state-level 1}.

\subsection{Sort Action Sets with LLMs and Search Solutions}
\begin{wrapfigure}{r}{0.5\textwidth}
    \centering
    \includegraphics[width=0.45\textwidth]{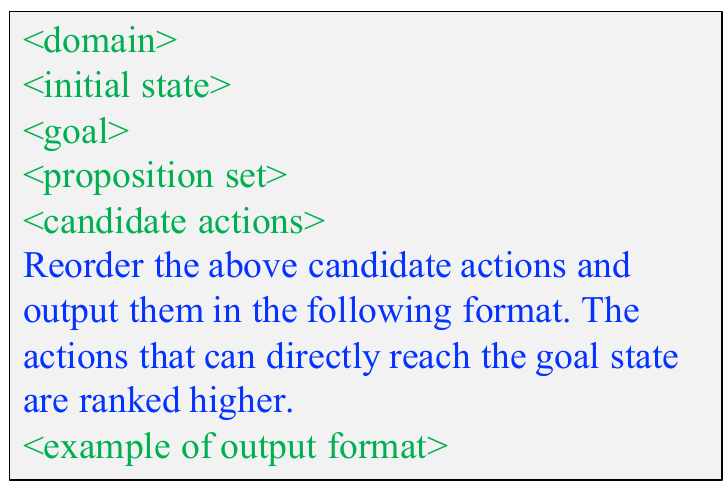}
    \caption{The prompt for sorting action sets}
    \label{fig:prompt2}
    \vspace{-3em}
\end{wrapfigure}
After we build a set of constraints in Step 8 of Algorithm \ref{lmm4planning}, we use off-the-shelf procedure presented in \cite{DBLP:journals/ai/BlumF97} to compute candidate action sets such that there are no conflicts (i.e., satisfying the constraints $\mathcal{C}$) among actions in each action set. After that, in Step 9, we consult LLMs to sort the action sets by designing the prompts as shown in Figure \ref{fig:prompt2}, which is similar to the prompt shown in Figure \ref{fig:prompt1} except the command in BLUE. After we get the sorted action sets $\mathbb{A}$, in Step 10, we conduct the dept-first search procedure as done in \cite{DBLP:journals/ai/BlumF97} by giving the priority of action sets based on the sorted action sets in $\mathbb{A}$.

%In Step 9 of Algorithm \ref{lmm4planning} (i.e., $sortActionsLLMs(PG^r,\mathcal{C})$), we sort action sets $PG^r$ with the same prompt as shown in Figure \ref{fig:prompt1} when selecting actions for expanding one more level of $PG^r$. 

%\subsection{Searching Solution Plans}

\ignore{
\subsection{Plan extraction}
\begin{algorithm}
    \caption{Solution-extraction($g$,$j$)}\label{solutionextraction}
    \textbf{input:} The set of goals $g$ we are trying to achieve; The level of the state $s_j$ \\
    \textbf{output:} solution \\
    \begin{algorithmic}[1]
        \IF {$j=0$} 
            \STATE return the solution
        \ENDIF
        \FOR{each literal $l \in g$}
            \STATE nondeterministically choose an action to use in state $s_{j–1}$ to achieve $l$
            \IF{any pair of chosen actions are mutex}           \STATE backtrack
            \ENDIF
            \STATE $g'$:= {the preconditions of the chosen actions}
            \STATE Solution-extraction($g'$, $j–1$)
        \ENDFOR
    \end{algorithmic}
\end{algorithm}
}

%\subsection{ Search in Planning Graphs}

%How to search for an optimal plan?

%How to do backtrack search based on quantum search?

\section{Experiment}

% 1. 前向剪枝 + 后向剪枝 的平衡: 暴力剪和排序后剪；只有前向；只有后向；
% 2. 实验领域增多到10个
% 3. 与GP算法比搜索节点数
% 4. 与纯LLMs比成功率
% 5. 不同领域的K可以不一样
% 6. 给定 初始状态+动作模型 =》 K = 100，由GP和我们方法的效果而定

\subsection{Experimental Setup}
% 在本次实验中，我们选取了四个不同场景的domain进行实验，每个domain随机选取了10个problem，包括：gripper、miconic、logistics、movie。具体的场景和规模描述如下：
% 1.gripper：利用已有的机器人手臂在不同房间之间进行搬运物品，场景中的object在6~20个之间。
% 2.miconic：利用电梯在楼层之间服务客人，以帮助他们达到想去的楼层，场景中的object在6~50个之间。
% 3.logistics：经典的物流问题之一，在不同城市的不同location之间，利用truck和airplane进行物品之间的运输，场景中的object在10~30个之间。
% 4.movie：模拟看电影期间的一些简单行为，场景中的object在45~155个之间。
% 5.blocks：作为最经典的规划问题之一，通过机械臂操作桌子上的不同木块以完成相应的目标。场景中的方块数量在20~40个之间。
% 6.satellite：携带不同仪器的卫星在空间中执行不同的任务，场景中的仪器和卫星等实体合计多达20~40个。
% 7.zenotravel：该领域涉及乘客坐飞机在不同城市之间旅行的问题，其中还有飞机的燃料管理。场景中的实体合计在20~30个之间。
% 8.driverlog：不同的卡车司机需要利用卡车做平台之间的货物运输调度问题。场景中的实体合计在20~30之间。
% 9.woodworking：木工需要操作车间内不同的机器、木板和部件完成加工的任务。场景中的实体合计在30~40之间。
% 10.openstacks：在流水线中，根据订单任务生产相应数量的产品。场景中的实体合计在10~30之间。

In the experiment, we evaluate {\ours} in ten planning domains with different scenarios, including gripper, miconic, logistics, movie, blocks, satellite, zenotravel, driverlog, woodworking and openstacks. Ten problems are randomly selected for each domain. The specific scenarios and sizes are described in Appendix \ref{domains}.

% 为了证明我们方法的有效性，我们设计了五组对比实验，都是采用python进行实现，将我们的方法和另外四种方法进行了比较，具体的描述如下：
% 1.GP：即是graph planning方法，实现了传统的图规划算法，作为最重要的baseline进行对比。我们直接提供domain.pddl和problem.pddl给规划器进行解决。
% 2.GPT3.5：我们直接将domain.pddl和problem.pddl里面的内容进行简单构造之后，再加入必要的命令提示，构成一个完整的提示输入GPT3.5中，让模型对问题进行直接求解。
% 3.GPT4：流程和GPT3.5相同。
% 4.GP-GPT3.5：我们在拓展层次或者回溯的时候，面对候选动作选择的时候，我们根据section prompt generation描述的一样把prompt构造好，让LLM从中选择一个最小的动作子集，利用这个动作子集进行接下来的算法操作，这里的LLM具体就是GPT3.5。
% 5.GP-GPT4：流程基本和GP-GPT3.5相同，只是LLM模型换成了GPT4。
To demonstrate the effectiveness of our {\ours} approach, we designed five sets of comparison experiments. The methods were implemented using Python. We compared our {\ours} approach with four other methods, which are listed below:
\begin{itemize}
    \item \textbf{GP}: It is the graph-based planning algorithm mentioned above. We implement the traditional graph planning algorithm as the most important baseline for comparison. We directly provide \textit{domain.pddl} and \textit{problem.pddl} to the planner for solving.
    \item \textbf{GPT-3.5}: We simply construct and splice the contents of \textit{domain.pddl} and \textit{problem.pddl} directly. We then add the necessary command prompts to form a complete prompt into GPT3.5 and command the model to solve the problem directly.
    \item \textbf{GPT-4}: The process is the same as GPT-3.5.
    \item \textbf{{\ours}-GPT3.5}: When we are expanding the hierarchy or backtracking, we are faced with a candidate action selection with LLMs. We guide the LLMs to select a minimal subset of actions from them and utilize this subset of actions for the next algorithmic operations, where LLMs are specifically GPT-3.5.
    \item \textbf{{\ours}-GPT4}: Replaced the LLM model with GPT4, otherwise same as {\ours}-GPT3.5.
\end{itemize}

\subsection{Experimental Metrics}
% 在上面实验的设定下，我们计算了5个指标对比不同方法之间的效果，分别为：问题解决的成功率、拓展动作总数、互斥动作总数、拓展层数以及回溯深搜的节点个数。
In the experimental framework described, we employed three distinct metrics to assess the efficacy of various methodologies: the problem-solving success rate, the cumulative count of expansion actions and the node count for backtracking in Depth-First Search (DFS).

% 问题解决的成功率：对于规划问题，问题能否解决毫无疑问是最重要的因素之一。无论是GP还是LLM的方法，我们都让他们能够产生足已解决问题的动作序列，只有问题能够通过这个动作序列从初始状态转移到目标状态，对应的问题才能算是获得了成功解决。除此之外，我们设定了对问题层数的上限，我们已知测试问题的最优动作序列长度，如果获得的解超过了这个层数则表示该方法对该问题的最优动作路径求解失败。设定层数上限的目的在于我们不仅仅要求规划器能解决问题，还要求规划器能够更高效地解决问题，输出的动作序列更加精简准确，尽量避免冗余动作的出现。
\textbf{Problem-solving success rate}. The solvability of a problem is a crucial metric in assessing planning problems. All approaches are required to generate a sequence of actions that is sufficient to solve the problem, and only if the problem can be transferred from the initial state to the goal state through this sequence of actions can the corresponding problem be considered to be successfully solved. Furthermore, we have established an upper bound on the depth of the problem-solving process. The optimal length of the action sequence for the test problem is known to us. Should the solution obtained surpass this predetermined depth, it signifies the inability of the method to successfully ascertain the optimal action path for this particular problem. Setting an upper bound on the depth of the problem-solving process serves the purpose of not only requiring the planner to solve problems but also demanding that it does so more efficiently. This ensures that the output action sequences are more concise and accurate, minimizing the occurrence of redundant actions.

% 拓展动作总数：在graph-based planning算法中，每层拓展动作是一个核心的步骤，其拓展的数量也是一个关键的指标。在能够保留有效动作的前提下，拓展的数量越少，后续生成的互斥动作对数量也就越少，回溯阶段深度搜索的分支也就越少，效率越高。所以在这里，我们计算不同方法在不同domain里面所有问题的每层拓展动作总数的平均数，作为一个重要的对比指标之一。
\textbf{Total number of expansion actions}. In the GP algorithm, the expansion of actions at each layer is a fundamental process, and the number of these expansions serves as a vital metric. Under the premise of preserving effective actions, fewer expansions result in a reduced count of mutually exclusive action pairs and subsequently fewer branches in the deep search phase of backtracking, thereby enhancing efficiency. Consequently, we compute the average total number of action expansions per layer across all problems, applying different methods within various domains, as a significant metric for comparison.

\textbf{Number of nodes for backtracking DFS}. This metric serves as the cornerstone for validating our optimization efforts, as the DFS during backtracking accounts for the majority of the computational load in the GP algorithm, overshadowing the forward expansion phase. Particularly when dealing with increasing expansion depths, the exponentially growing number of DFS poses the most significant challenge for GP algorithms in tackling large-scale problems or complex solution sequences. We primarily utilize this metric to ascertain which method truly enhances the efficiency of the planning process.

% 对于DFS节点统计这个指标，我们只统计了GP和GP-GPT4两种方法的数据。主要的原因有两点。一个是因为这些指标在问题成功解决的前提下才具有意义，如果解决失败则无法统计，所以我们放弃了对GP-GPT3.5的统计，因为其普遍较低的成功率。另外一个是，这些指标只有基于GP框架下才能进行统计，所以直接用GPT3.5和GPT4对问题进行解决，无法对这些数据进行统计。
Regarding the number of nodes for backtracking DFS, our analysis was confined to data from the GP and {\ours}-GPT4 methods, primarily for two reasons. Firstly, the metric is relevant only in scenarios where the problem is successfully solved; failed solutions do not yield countable data. Consequently, we excluded {\ours}-GPT3.5 from our statistical analysis due to its comparatively lower success rate. Secondly, these metrics are inherently calculable within the GP framework alone. Hence, directly solving problems using GPT-3.5 and GPT-4 precludes the possibility of gathering this data, as these methods operate outside the GP framework.

\subsection{Experimental Results}
% 我们将成功率展示在了表1中，将LLM对GP拓展过程中的动作剪枝效果呈现在了图7中，而表2则展示了我们的方法相比起传统的GP算法在DFS指标上的实验结果，以及相关的消融实验。（待翻译）
% We present the success rate in Table \ref{table:success_rate}, the action pruning effect of LLM during the GP expansion process in Figure \ref{fig:logistics_layers}, and the ablation study comparing our method to traditional GP algorithms Table \ref{table:ablation_results}.

We present the success rates in Table \ref{table:success_rate}, depict the pruning effects of action expansion for LLM on GP in Figure \ref{fig:logistics_layers}, and showcase experimental results in Table \ref{table:ablation_results} comparing our approach to traditional GP algorithms in terms of the number of nodes for backtracking DFS metrics, along with relevant ablation studies.

\begin{table*}[!ht] % [!ht]表格在文本中放置的位置参数（努力放在当前位置，实在放不下，将放在下一页的顶部）
\centering % 表格整体居中
\caption{Success rate results. In the table, each row corresponds to a distinct domain, while each column represents a separate approach or method. The values presented within the table indicate the success rate for each combination of domain and approach, with these rates quantified on a scale ranging from 0 to 1.}
\begin{tabular}{cccccc} % 其中，|c|表示文本居中，文本两边有竖直表线。
\hline & \textbf{GPT-3.5} & \textbf{GPT-4} & \textbf{GP} & \textbf{{\ours}-GPT3.5} & \textbf{{\ours}-GPT4}  \\ \hline
\textbf{gripper}   & 0.00 & 0.60 & 0.70 & 0.00 & \textbf{1.00}  \\ 
\textbf{miconic}   & 0.10 & 0.50 & 0.60 & 0.10 & \textbf{1.00} \\ 
\textbf{logistics} & 0.20 & 0.60 & 0.60 & 0.20 & \textbf{1.00} \\ 
\textbf{movie}   & \textbf{1.00} & \textbf{1.00} & \textbf{1.00} & \textbf{1.00} & \textbf{1.00} \\
\textbf{blocks}     & 0.10 & 0.70 & 0.60 & 0.30 & \textbf{1.00} \\
\textbf{satellite}    & 0.00 & 0.50 & 0.90 & 0.10 & \textbf{1.00} \\
\textbf{zenotravel}   & 0.20 & 0.60 & 0.90 & 0.20 & \textbf{1.00} \\
\textbf{driverlog}     & 0.00 & 0.10 & 0.90 & 0.20 & \textbf{1.00} \\
\textbf{woodworking}   & 0.90 & 0.90 & 0.70 & \textbf{1.00} & \textbf{1.00} \\
\textbf{openstacks}    & 0.10 & 0.20 & \textbf{1.00} & 0.20 & \textbf{1.00} \\ \hline
\end{tabular}
\label{table:success_rate}
\end{table*}

% 我们设置了四组实验验证前向剪枝和后向回溯排序的作用，分别为：
% 1.{\our}-pruned-sorted：前向剪枝+回溯排序
% 2.{\our}-pruned-unsorted：前向剪枝+回溯不排序
% 3.{\our}-unpruned-sorted：前向不剪枝+回溯排序
% 4.{\our}-unpruned-unsorted：前向不剪枝+回溯不排序，即GP的结果。
\textbf{Ablation Experiment}: We conducted four ablation experiments to ascertain the effectiveness of forward pruning and backward sorting, detailed as follows:
\begin{enumerate}
    \item \textbf{{\ours}}: This method involves both forward pruning and backward sorting.
    \item \textbf{{\ours}-unsorted}: Here, we implement pruning without sorting.
    \item \textbf{{\ours}-unpruned}: In this approach, sorting is used, but not pruning.
    \item \textbf{GP}: This method involves neither pruning nor sorting.
\end{enumerate}

\begin{table*}[!ht] % [!ht]表格在文本中放置的位置参数（努力放在当前位置，实在放不下，将放在下一页的顶部）
\centering % 表格整体居中
\caption{In the table, each row corresponds to a distinct domain, while each column represents a group of ablation experiments. The values presented within the table indicate the number of nodes for backtracking DFS. Both pruning and sorting effectively enhance search efficiency, leading to a substantial reduction in the number of nodes required for searching. Generally, pruning tends to be slightly more effective than sorting.}
\begin{tabular}{ccccc} % 其中，|c|表示文本居中，文本两边有竖直表线。
\hline & \textbf{{\ours}} & \textbf{{\ours}-unsorted} & \textbf{{\ours}-unpruned} & \textbf{GP}  \\ \hline
\textbf{gripper}   &  \textbf{6839}  &  11294   &  4850376   & 8486698  \\ 
\textbf{miconic}   &  \textbf{11891}   &  47863   &  445145   & 2018484  \\ 
\textbf{logistics} &  \textbf{59}    &  85 &  1226   & 1261250  \\ 
\textbf{movie}     &  \textbf{975163}  &  1211830   &  \textbf{975163}   & 10869160 \\
\textbf{blocks}     &  \textbf{4572}  &  7272   &  83129  & 1205223 \\
\textbf{satellite}     &  \textbf{46619}  &  94811   &  67167049  & 88779785 \\
\textbf{zenotravel}     &  \textbf{1548}  &  12166   &  839527  & 2259283 \\ 
\textbf{driverlog}     &  \textbf{574}  &  1916   &  58311  & 1579486 \\ 
\textbf{woodworking}     &  \textbf{49}  &  3924   &  48553  & 114502 \\ 
\textbf{openstacks}     &  \textbf{107}  &  409   &  13577  & 24267 \\ \hline
\end{tabular}
\label{table:ablation_results}
\end{table*}

\subsection{Experimental Analysis}
% 实验结果分析
% 首先我们从Table \ref{table:success_rate}这个表格可以得出一些结论。GPT3.5基本上只能解决movie这样动作序列长度较短的简单任务，在其他domain里面解决成功率非常低，进而也很难提升GP算法的能力。而GPT4能力相比起GPT3.5有了大幅度的提升，在推理能力特别是对于长动作序列决策上有了很强的提升。在拥有更多推理和常识能力的GPT4加持下，GP能够提升部分domain的解决成功率。我们注意到GP也有一些失败的样例，这是因为我们在测试的时候加入了部分的有损数据，即我们会随机删减domain文件中一定比例的动作前提命题和效果命题，对于极度依赖domain文件完整性的GP算法来说，这些样例在不同领域有不同的影响。我们在另外一个表格中呈现了随机删减对于求解成功率的影响。对于GP算法每层的拓展来说，动作的前提命题和效果命题是必须要提供的。然而，对于有LLM加持的GP算法，面对缺失的动作命题，LLM也仍然能做出合理的动作决策，帮助完成当前的规划任务。因此，在成功率方面，我们的算法对于GP有更强的鲁棒性。

\textbf{Analysis of the success rate of planning}:  From Table \ref{table:success_rate}, several conclusions can be drawn. GPT3.5 exhibits competence primarily in resolving simple problems with short action sequence lengths, such as in the \textbf{movie} domain, while its success rates are notably low in other domains. Consequently, it struggles to enhance the capabilities of GP algorithms. Conversely, GPT4 demonstrates substantial improvements in abilities compared to GPT3.5, particularly in reasoning skills and decision-making involving long action sequences. With the enhanced reasoning and commonsense capabilities of GPT4, GP shows an enhanced success rate in certain domains. We observe instances of failure in GP, attributed to the inclusion of partially corrupt data during testing. Specifically, we introduce a proportion of corrupted data by randomly removing action preconditions and effects propositions from domain files. These instances have varying impacts across different domains, particularly affecting traditional GP algorithms reliant on the completeness of domain files. We detail the influence of random removal on success rates in Table \ref{table:robustness_experiments}. For each layer of GP expansion, the provision of action preconditions and effects propositions is essential. However, in the case of LLM-augmented GP algorithms, {\ours} is capable of making rational action decisions even in the presence of missing action propositions, thereby aiding in the completion of current planning tasks. Consequently, our algorithm exhibits greater robustness in terms of success rates compared to traditional GP approaches. The specific settings for the robustness experiments are detailed more extensively in section \ref{robustness_experiments} of the appendix.

% % 在鲁棒性实验中，我们把domain文件中的动作前提命题和效果命题（两者统称为“条件”），随机删除一定比例（例如10%，20%，30%，40%，50%），测试GP算法能不能输出正确的解。我们对每个domain的每个比例测试五组实验，每组实验测试3次取平均的成功率。这里的失败不仅仅包括前面的定义，由于缺失谓词后GP算法可能会陷入循环导致程序卡死，我们还额外设定了一个较长的时间阈值。当超过这个时间阈值，我们也将其作为失败解决的案例。因为规划的时效性也是非常重要的指标，如果一个问题在正常情况的求解时间的几倍甚至几十倍的时间内也获得不了解决，那么这个规划器也是没有意义的。（可以考虑放到附件中，如果篇幅太长）
% In the robustness experiments in Table \ref{table:robustness_experiments}, we randomly delete a certain proportion (e.g., 10\%, 20\%, 30\%, 40\%, 50\%) of action preconditions and effects propositions (referred to collectively as conditions) from the domain files and assess whether the GP algorithm can produce correct solutions. We conduct five sets of experiments for each proportion in every domain, with each experiment tested three times to obtain the average success rate. In addition to the previously defined criteria, failures also include cases where the GP algorithm may become stuck in a loop or program deadlock due to missing predicates. Therefore, we set an extended time threshold. If the solving process exceeds this threshold, we consider it as a failed solution. Timeliness in planning is crucial; if a problem remains unsolved for several times the normal solving duration, the planner becomes impractical.

\begin{table*}[!ht] % [!ht]表格在文本中放置的位置参数（努力放在当前位置，实在放不下，将放在下一页的顶部）
\centering % 表格整体居中
\caption{This table illustrates experiments on the robustness of missing action predicates.
In the table, each row corresponds to a distinct domain. Each column in the table represents the proportion of predicates we removed. A higher proportion indicates a greater amount of missing information, posing increased difficulty for the planner to solve the problem. The values in the table represent the success rates of GP in solving the problems. In the majority of domains, as the proportion of deleted predicates increases, the success rate of GP planning decreases. Overall, this indicates that GP exhibits poor robustness to missing action predicates.}
\begin{tabular}{cccccc} % 其中，|c|表示文本居中，文本两边有竖直表线。
\hline & \textbf{10\%} & \textbf{20\%} & \textbf{30\%} & \textbf{40\%} & \textbf{50\%}  \\ \hline
\textbf{gripper}   &  0.40  &  0.87   &  0.73   & 0.80 & 0.67  \\ 
\textbf{miconic}   &  0.60  &  0.60   &  0.60   & 0.80   & 0.40  \\ 
\textbf{logistics} &  0.20  &  0.80   &  0.80   & 0.67 & 0.60  \\ 
\textbf{movie}     &  1.00  &  1.00   &  1.00   & 1.00 & 0.60 \\
\textbf{blocks}     &  1.00  &  0.26   &  0.07   & 0.27 & 0.47 \\
\textbf{satellite}     &  0.60  &  0.73   &  1.00   & 0.86 & 0.87 \\
\textbf{zenotravel}    &  0.93  &  0.93   &  1.00   & 0.93 & 0.80 \\ 
\textbf{driverlog}     &  0.80  &  1.00   &  1.00   & 1.00 & 1.00 \\ 
\textbf{woodworking}   &  1.00  &  0.73   &  0.40   & 0.27  & 0.40 \\ 
\textbf{openstacks}   &  1.00  &  1.00   &  1.00   & 1.00 & 1.00 \\ \hline
\end{tabular}
\label{table:robustness_experiments}
\end{table*}

% 在观察具体产生的动作序列时，我们发现，虽然GPT4具有一定的求解成功率，但是单纯用GPT4产生出来的动作序列长度普遍都比GP长，加入了图规划的GPT4，即我们的方法，能够更有效地产生更优的动作序列。
Upon examining the generated action sequences, we observed that although GPT4 achieves a certain level of success in solving problems, the action sequences it produces tend to be longer compared to those generated by GP alone. By integrating GPT4 with graph planning, {\ours} can effectively generate more optimal action sequences.

% 除了规划成功率，我们的方法相比起GP算法，在搜索效率上的提升更是巨大。我们可以从表格{table:ablation_results}中明显看出，在成功输出规划解的问题中，我们极大缩减了搜索节点的代价，优化的幅度是指数级别的。那么，拥有LLM的GP方法是怎么提升搜索效率的？我们深入实验的案例中，发现其优化的本质主要包含两个方面：1.在前向拓展的过程中，LLM对拓展动作进行了高效且有效的剪枝，让拓展动作和互斥动作的总数获得了不同程度的稳定下降，从而前向拓展的计算量相应减少了。2.在回溯的深度搜索过程中，LLM能让搜索优先搜索更接近于规划解的集合，从而更快获得规划解，节省无效搜索的时间。我们给出了一个logistics里面的例子，以证明我们上面的分析。解决该问题的时候，GP和GP-GPT4都拓展了10层，横坐标代表层数，数字越小越接近于初始状态，数字越大越接近于目标状态，纵坐标代表拓展的动作数量，不仅包括domain里面的动作，还包含大量在图规划算法里面特有的空动作，所以实际有效动作的剪枝比例是相当高的。可以看出，利用LLM用来剪枝的效果是非常明显的，对于每层的搜索动作都进行了剪枝。
\textbf{Analysis of search efficiency}: Besides planning success rates, our method significantly improves search efficiency compared to GP algorithms. This enhancement is evident from Table \ref{table:ablation_results}, where, among problems with successful planning outputs, we drastically reduce the cost of search nodes, achieving an exponential level of optimization. So, \textbf{how does the LLM-augmented GP method enhance search efficiency?}

Through in-depth analysis of experimental cases, we identify two main aspects of optimization:
\begin{enumerate}
    \item During forward expansion, LLM efficiently and effectively prunes the expansion actions, leading to varying degrees of stable reduction in the total number of expanded actions and mutually exclusive actions. Consequently, the computational load of forward expansion decreases correspondingly.
    \item During the depth-first search backtracking process, LLM prioritizes searching closer to the set of planning solutions, accelerating the attainment of planning solutions and saving time by avoiding ineffective searches.
\end{enumerate}

\begin{figure}[!ht]
    \includegraphics[width=.9\textwidth]{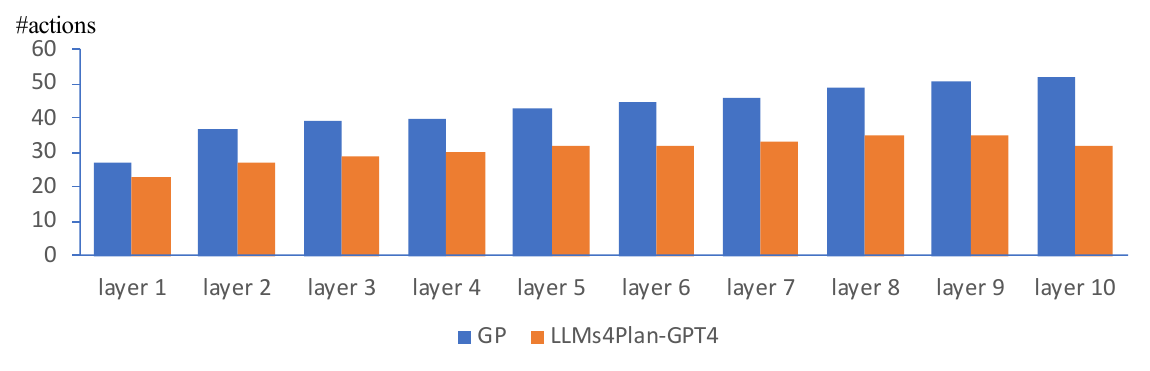}
    \caption{An example of action pruning. Both the GP and {\ours}-GPT4 methods expanded through 10 layers. The horizontal axis represents the layer number, with lower numbers indicating proximity to the initial state and higher numbers nearing the goal state. The vertical axis shows the count of expanded actions, including both domain-specific actions and numerous empty actions, characteristic of the graph planning algorithm. This implies a high pruning ratio for genuinely effective actions. LLMs prune almost every layer of expansion actions and the data in the table also contains many empty actions.}
    \label{fig:logistics_layers}
\end{figure}

% 我们给出了一个logistics里面的例子，以证明我们上面的分析。除此之外，我们还在附件中给出了对于互斥动作和拓展动作总数的进一步实验结果和分析。
We provide an example from the 'logistics' domain to illustrate our analysis in Figure \ref{fig:logistics_layers}, where we compare the number of expansion actions before and after LLM pruning. The application of LLM for pruning demonstrates significant efficacy across all layers. In addition, we have provided further experimental results and analysis on the total number of mutually exclusive actions and expanded actions in section \ref{add_experiments} of the supplementary materials.

% 对于剪枝来说，最大的风险就是会把必要的动作减去导致问题无法求解。在实验中，我们发现LLM会在部分层中把关键动作减去导致问题无法获得有效的求解，不过我们加入了剪枝概率能至少保证算法的完备性。实验证明，即使我们会通过剪枝概率纠正LLM的错误剪枝行为的过程中，会让算法增加一些拓展和搜索步骤，但是将这些搜索步骤的代价累积起来，消耗的搜索代价也是远远低于了单纯用GP算法解决问题的代价，我们在消融实验的表中，展示的就是具有完备性的我们的方法和GP算法的对比结果。
For pruning, the greatest risk is removing necessary actions, rendering the problem unsolvable. In our experiments, we observed instances where LLM prunes crucial actions in certain layers, resulting in the inability to obtain effective solutions. However, we introduced pruning probabilities to ensure algorithm completeness. Experimental results demonstrate that although the process of correcting LLM's erroneous pruning behavior through pruning probabilities may introduce additional expansion and search steps, the cumulative cost of these search steps remains significantly lower than the cost of solely using GP algorithms to solve problems. The results presented in Table \ref{table:ablation_results} compare the outcomes of our method {\ours}, which ensures completeness, with those of GP algorithms.

% 消融实验分析：从table（ablation_results）可以看出，剪枝和排序都起到了提升搜索效率的作用，两者的结合也进一步加强了这种提升效果。相比较而言，剪枝比排序时候的效果作用略好一些，分析原因后，我们认为LLM在剪枝的时候也会对剩下的动作进行合理的排序，而排序的话，由于动作较多，文本较长，LLM容易出现局部排序错误。在实验中，我们还发现，针对图规划中的空动作，LLM并不会给出很高的优先级，更倾向于将其排序到后面去。我们分析，非空动作的优先探索确实更有意义，因为一层如果不做任何动作，该层其实就没有存在的必要了。
\textbf{Analysis of ablation experiments}: Table \ref{table:ablation_results} reveals that both pruning and sorting contribute to enhanced search efficiency, with their combination amplifying this effect. Notably, pruning appears slightly more effective than sorting. This is likely because LLM, while pruning, also organizes the remaining actions logically. In contrast, sorting may lead to minor errors due to the multitude of actions and lengthy text. In this regard, we require further optimization of natural language processing techniques tailored for handling extremely long texts to enhance our framework's capability in solving more complex problems. The experiments also indicate that LLM tends not to prioritize empty actions in graph planning, favoring their later arrangement. This aligns with our analysis suggesting that prioritizing non-empty actions is more productive, as a layer without any action is essentially redundant.

% 我们在分析失败的例子的时候，发现在使用GPT4对候选动作剪枝的时候，拓展的层次越深，GPT4越容易出现剪枝错误，具体表现为将有效动作舍弃掉了，从而使拓展的层数加深，更不利于解决问题。我们分析，具体原因主要有两点。第一，拓展层次越深，谓词集合和候选动作集合越来越大，相对应输入的prompt越来越长，长文本让GPT4对前面的信息记忆逐渐丧失，从而决策失误。第二，由于图规划里面的谓词集合和传统规划的当前状态有所不同，会导致LLM无法对谓词集合做正确的分析，从而筛选掉了有效动作。具体到miconic这个问题而言，一般来说电梯只有一部，在传统的基于搜索的规划中，它只可能出现在一个楼层，但是图规划的状态集合中，却能允许它处于多个楼层的状态同时出现。GPT4对此的做法就是随机保留其中一个状态，那一旦选择失误，有效动作就被删除了。我们称之为，LLM缺乏对谓词集合的组合能力。
\textbf{Analysis of the advantages and disadvantages of {\ours}}: Upon analyzing examples where solutions failed, we observed that GPT4 is more prone to pruning errors at deeper expansion levels. This results in the discarding of effective actions, thereby unnecessarily increasing the expansion layers and hindering problem resolution. We attribute this to two primary factors. Firstly, as the expansion level deepens, both the predicate set and the candidate action set expand, leading to increasingly lengthy input prompts. This prolonged text can cause GPT-4 to gradually lose track of previous information, resulting in decision-making errors. Secondly, the nature of the predicate set in graph planning diverges from traditional planning's current state representation. This discrepancy impairs LLM's ability to accurately analyze the predicate set, leading to the erroneous elimination of effective actions. LLM lacks capacity to analyze complex predicate set combinations.
% 这个具体例子可以考虑一下加不加。加了更好说明LLM缺乏谓词组合分析的能力，但受限于篇幅又可能不加。
% A specific instance is observed in the miconic domain, where typically only one elevator is present. In traditional search-based planning, the elevator is confined to a single floor, but graph planning allows it to be simultaneously on multiple floors. GPT4, in attempting to reconcile this, might randomly retain one state, but an incorrect selection here can eliminate a valid action. This issue underscores LLM's limited capacity to analyze complex predicate set combinations.

% 我们在分析更多失败的例子中，总结认为图规划的优势在于，能够高效解决并行执行大量不互斥动作的问题，因为这样可以把大量不互斥的动作集中在同一层里面。相对而言，其劣势就在于，不适合执行高度复杂且超长动作序列的问题，因为如果一个问题最优的解决序列都非常长，图规划必然要拓展非常深的层次，就算拥有了非常有效的剪枝，也无法解决由于深度带来的回溯深搜时的指数级复杂度的增长。
Our analysis of additional failure examples indicates that graph planning excels in efficiently handling numerous non-mutually exclusive actions in parallel, due to its ability to group these actions within the same layer. However, its limitation becomes apparent in scenarios requiring the execution of highly complex and extremely long action sequences. If a problem's optimal solution sequences are lengthy, the planning graph must be expanded considerably deeper. Despite effective pruning, this does not resolve the issue of exponential complexity growth in backtracking DFS caused by increased depth.

\section{Conclusion and Future Work}
% 结论
% 综上所述，相比起原来的图规划算法graph-based planning algorithm，我们的方法不仅提升了问题解决的成功率，而且对于原来本来就能够解决的问题上，大大提升了搜索的效率，指数级地降低了计算的复杂度。
% 总的来说，我们通过在多个domain上进行了对比实验，能够验证我们方法的有效性，即利用LLM嵌入图规划算法中，能够大幅度提升图规划算法的求解能力。我们的方法不仅提升了问题解决的成功率，而且对于原来本来就能够解决的问题上，大大提升了搜索的效率，指数级地降低了计算的复杂度。在LLM和GP的组合中，LLM为GP提供非常有效的先验知识帮助GP避免掉多余的计算，更有效率地输出解决问题的动作序列；GP则通过提供候选动作集合等信息，能够帮助LLM产生更优的解决动作路径。
% In summary, we have demonstrated the effectiveness of our approach by conducting comparative experiments on multiple domains. our {\ours} approach can substantially improve the solving ability of graph planning algorithms. Specifically, our {\ours} approach not only improves the success rate of problem-solving but also greatly improves the efficiency of the search and exponentially reduces the computational complexity. In the combination of LLM and GP, LLM provides GP with very effective a priori knowledge to help GP avoid redundant computations and output problem-solving action sequences more efficiently, while GP, by providing information such as the set of candidate actions, can help LLM to generate a better solution action path.

Our comparative experiments in multiple domains demonstrated the efficacy of our {\ours} in significantly enhancing the problem-solving capabilities of graph planning algorithms. Notably, {\ours} boosts not just the success rate of problem resolution but also markedly enhances search efficiency and substantially reduces computational complexity. %In the synergy between LLMs and GP, LLMs impart crucial a priori knowledge to GP, aiding in the avoidance of superfluous computations and facilitating more efficient generation of solution action sequences. Concurrently, GP, with its provision of elements like the candidate action set, assists LLM in generating a better solution action path.
% 未来工作
% 本篇工作的基础上，还有许多问题需要探讨。
% 1.本篇的LLM仅仅使用了GPT3.5和GPT4，其实还有很多大模型可以选择。
% 2.并非每层用LLM都能利于问题的求解效率，具体哪几层用，层数和层的位置选择，是一个需要探讨的问题。
% 3.我们分析GPT4缺乏了一定的组合问题，后续能不能通过其他方法解决这个问题，比如利用思维链的方式探索更多有效的prompt。
% 4.我们实验中失败的例子大多是大规模的问题，我们是否能进一步用LLM进一步解决大规模的图规划问题，让图规划器能够解决大规模的问题。
% 5.我们的实验主要采用zero-shot的方式，如果能用few-shot或者one-shot的话，效果或许能进一步提升。
\ignore{
Based on our current research, several key issues have emerged that require further exploration and resolution in future work. We summarize these as follows:
% \begin{enumerate}
%     \item The LLM in this paper only uses GPT3.5 and GPT4, but there are many other large models to be selected.
%     \item Not every layer of LLM can benefit the problem-solving efficiency. The choice of the number of layers and the location of the layers is an issue that needs to be explored.
%     \item We analyzed that GPT4 lacks combinatorial analysis capability. Whether this problem can be solved by other means, such as exploring more effective prompts by using the chain-of-thought approach, is a future direction.
%     \item Most of the failed examples in our experiments are large-scale problems. Can we further use LLM to solve large-scale graph planning problems, so that the graph planner has the ability to solve large-scale problems?
%     \item Our experiments currently only use zero-shot for prompt. if we could use few-shot or one-shot, the effect might be further improved.
% \end{enumerate}
\begin{enumerate}
    \item Determining the optimal layers for LLM application is critical for problem-solving efficiency. Identifying which layers to use requires further investigation.
    \item Exploring solutions to GPT-4's limited combinatorial analysis capability is a future direction. This could include developing more effective prompts or employing the chain-of-thought approach.
    \item Enhancing LLM's capability to address large-scale graph planning problems, as most failures in our experiments involved such challenges, is a key future exploration.
\end{enumerate}
}
%\section{Final Remarks}
The runtime of {\ours} is currently hindered by multiple LLMs calls. While our method requires multiple LLMs calls, it provides substantially improved results. There are also various ways to enhance runtime performance like using smaller LLMs like Llama \cite{DBLP:journals/corr/abs-2302-13971} or distilling LLMs' knowledge into a smaller model \cite{DBLP:conf/acl/ShridharSS23,DBLP:conf/acl/HsiehLYNFRKLP23,DBLP:conf/icml/LiangZZHCZ23}. Those are interesting avenues for future research. Instead of leveraging LLMs to assist planning, it would also be possible to study acquring action models \cite{DBLP:journals/ai/ZhuoK17} %\cite{DBLP:conf/aips/ZhuoYPL11,DBLP:conf/aaai/Zhuo15,DBLP:journals/ai/ZhuoK17} 
and more planning frameworks \cite{DBLP:journals/ai/JinZXWK22} with the help of LLMs.
\ignore{
\footnote{For the purpose of algorithm comparison we consider the ratio of worst-case time-complexity upper bounds. Specifically, an algorithm with time-complexity bound $\tilde{O}(g(n))$ is said to improve upon an $\tilde{O}(f(n))$ algorithm by a factor of $\frac{f(n)}{g(n)}$.}
\footnote{We use the notation $\tilde{O}(f(n)) = O(f(n)polylogf(n))$ to suppress factors that are polylogarithmic in the dominant part of the scaling.}
}

%\section*{Impact Statements}
%This paper presents work whose goal is to advance the field of Machine Learning. There are many potential societal consequences of our work, none which we feel must be specifically highlighted here.
\newpage
\bibliographystyle{plain}
\bibliography{neurIPS2024}

%%%%%%%%%%%%%%%%%%%%%%%%%%%%%%%%%%%%%%%%%%%%%%%%%%%%%%%%%%%%

\appendix

\section{Appendix / supplemental material}

\subsection{Description of Testing Domains}\label{domains}
In the experiment, we evaluate {\ours} in ten planning domains with different scenarios, including gripper, miconic, logistics, movie, blocks, satellite, zenotravel, driverlog, woodworking and openstacks, which are:
\begin{itemize}
    \item \textbf{Gripper}: Tasks utilize an existing robot arm to move objects between rooms, with between 6 and 20 objects in the scene.
    \item \textbf{Miconic}: Tasks uses elevators to serve guests between floors and help them reach the floor they want to go to, with between 6 and 50 objects in the scene.
    \item \textbf{Logistics}: One of the classic logistics problems. Transportation of items between different locations in different cities using trucks and airplanes, with between 10 and 30 objects in the scenario.
    \item \textbf{Movie}: Simulate some simple behaviors while watching a movie with between 45 and 155 objects in the scene.
    \item \textbf{Blocks}: As one of the most classic planning problems, it involves manipulating various blocks on a table using a robotic arm to achieve specific goals. The number of blocks in the scenario ranges between 20 and 40.
    \item \textbf{Satellite}: Satellites equipped with various instruments perform different tasks in space. The total number of entities, including instruments and satellites, ranges from 20 to 40.
    \item \textbf{Zenotravel}: This field involves the problem of passengers traveling between different cities by airplane, including the management of aircraft fuel. The total number of entities in the scenario ranges from 20 to 30.
    \item \textbf{Driverlog}: Different truck drivers need to coordinate the transportation of goods between platforms using trucks. The total number of entities in the scenario ranges from 20 to 30.
    \item \textbf{Woodworking}: A carpenter needs to operate various machines, wooden boards, and components in the workshop to complete processing tasks. The total number of entities in the scenario ranges from 30 to 40.
    \item \textbf{Openstacks}: In an assembly line, the corresponding number of products are produced based on order tasks. The total number of entities in the scenario ranges from 10 to 30.
\end{itemize}

% Optionally include supplemental material (complete proofs, additional experiments and plots) in appendix.
% All such materials \textbf{SHOULD be included in the main submission.}

\subsection{Robustness Experimental Setup} \label{robustness_experiments}
% 在鲁棒性实验中，我们把domain文件中的动作前提命题和效果命题（两者统称为“条件”），随机删除一定比例（例如10%，20%，30%，40%，50%），测试GP算法能不能输出正确的解。我们对每个domain的每个比例测试五组实验，每组实验测试3次取平均的成功率。这里的失败不仅仅包括前面的定义，由于缺失谓词后GP算法可能会陷入循环导致程序卡死，我们还额外设定了一个较长的时间阈值。当超过这个时间阈值，我们也将其作为失败解决的案例。因为规划的时效性也是非常重要的指标，如果一个问题在正常情况的求解时间的几倍甚至几十倍的时间内也获得不了解决，那么这个规划器也是没有意义的。（可以考虑放到附件中，如果篇幅太长）
In the robustness experiments in Table \ref{table:robustness_experiments}, we randomly delete a certain proportion (e.g., 10\%, 20\%, 30\%, 40\%, 50\%) of action preconditions and effects propositions (referred to collectively as conditions) from the domain files and assess whether the GP algorithm can produce correct solutions. We conduct five sets of experiments for each proportion in every domain, with each experiment tested three times to obtain the average success rate. In addition to the previously defined criteria, failures also include cases where the GP algorithm may become stuck in a loop or program deadlock due to missing predicates. Therefore, we set an extended time threshold. If the solving process exceeds this threshold, we consider it as a failed solution. Timeliness in planning is crucial; if a problem remains unsolved for several times the normal solving duration, the planner becomes impractical.

\subsection{Additional Experiments} \label{add_experiments}
% 在本部分中，我们主要讨论我们的算法对于GP算法高效的本质原因。首先我们先解释两个评估指标。
In this section, our primary focus lies in elucidating the fundamental reasons behind the efficiency of our algorithm compared to the GP algorithm. In this section, our primary objective is to delve into the fundamental reasons underlying the efficiency of our algorithm in contrast to the GP algorithm. To commence, we will elucidate two evaluation metrics: the total number of expansion actions and the total number of mutually exclusive actions.

% 拓展动作总数：在graph-based planning算法中，每层拓展动作是一个核心的步骤，其拓展的数量也是一个关键的指标。在能够保留有效动作的前提下，拓展的数量越少，后续生成的互斥动作对数量也就越少，回溯阶段深度搜索的分支也就越少，效率越高。所以在这里，我们计算不同方法在不同domain里面所有问题的每层拓展动作总数的平均数，作为一个重要的对比指标之一。
\textbf{Total number of expansion actions}. In the GP algorithm, the expansion of actions at each layer is a fundamental process, and the number of these expansions serves as a vital metric. Under the premise of preserving effective actions, fewer expansions result in a reduced count of mutually exclusive action pairs and subsequently fewer branches in the deep search phase of backtracking, thereby enhancing efficiency. Consequently, we compute the average total number of action expansions per layer across all problems, applying different methods within various domains, as a significant metric for comparison.

% 互斥动作总数：互斥动作是由候选拓展动作集根据多种不同的互斥条件生成出来的，这个指标的意义在于，不管是前向拓展的时候互斥动作集合的生成，还是回溯的时候我们会根据互斥动作集对回溯时的候选动作进行筛选判断，互斥动作总数越少，能够节省一定程度的计算时间。
\textbf{Total number of mutually exclusive actions}. Mutually exclusive actions are generated from the set of candidate expansion actions based on many different mutually exclusive conditions. The critical aspect of this metric lies in the fact that a lower total number of mutually exclusive actions translates into significant computational time savings. This efficiency is evident both during the generation of the mutually exclusive action set in the forward expansion phase and in the filtering of candidate actions based on this set during the backtracking process.

% 我们对一些domain进行了相关指标的统计，结果呈现在表格1中。从表格中数据我们不难看出，在拥有了LLM的知识之后，拓展动作和互斥动作的总数获得了不同程度的稳定下降，前向拓展的计算量相应减少了。综合而言，这组数据验证了我们在实验分析中对于我们的方法的高效性的分析。
We conducted statistical analysis on relevant metrics for several domains, and the results are presented in Table \ref{table:more_results}. The data indicates that both the total number of expansion actions and mutually exclusive actions experience a consistent decline with LLM integration. This suggests a reduction in the computational effort required for forward expansion. Overall, these data corroborate the conclusion drawn from our experimental analysis regarding the efficiency of {\ours}.

\begin{table*}[!ht] % [!ht]表格在文本中放置的位置参数（努力放在当前位置，实在放不下，将放在下一页的顶部）
\centering % 表格整体居中
\caption{LLM pruning results. In the table, each column signifies a distinct domain. It displays the statistical outcomes of the two methods across two indicators. Smaller numbers in the table denote lower computational resource usage and greater effectiveness.}
    \begin{tabular}{ccccc} % 其中，|c|表示文本居中，文本两边有竖直表线。
        \hline & \multicolumn{2}{c}{\textbf{Expansion Actions}} & \multicolumn{2}{c}{\textbf{Mutex Actions}}  \\ 
        \cmidrule(lr){2-3} \cmidrule(lr){4-5} 
        \textbf{Domain} & GP & {\ours} & GP & {\ours}  \\ \hline
        \textbf{gripper}   & 157 & \textbf{128} & 1656 & \textbf{1179}  \\
        \textbf{miconic}   & 181 & \textbf{176} & 792  & \textbf{668} \\ 
        \textbf{logistics} & 294 & \textbf{167} & 1726 & \textbf{161} \\
        \textbf{movie}     & 401 & \textbf{308} & 7    & 7    \\ \hline
    \end{tabular}
\label{table:more_results}
\end{table*}

\subsection{Experimental Case Presentation} \label{case_presentation}
% 在本部分中，我们将展示一个具体的案例，展示我们是如何将LLM嵌入到GP算法当中的，这个具体的案例取自于logistics 领域之中的一个问题。具体而言，我们将展示在解决这个问题的过程中，每一层的三类动作集合。第一类动作是图规划拓展和回溯的动作集合，我们称之为GP-ACTION-SET，第二类动作是LLM拓展的动作集合，我们称之为LLM-EP-ACTION-SET，第三类动作是LLM回溯的动作集合，我们称之为LLM-BP-ACTION-SET。
In this section, we will present a specific case to illustrate how we integrate LLM into the GP algorithm. This particular case is drawn from a problem in logistics domain. In our demonstration, we will present three sets of actions at each layer involved in addressing this problem. The first set of actions pertains to graph planning expansion and backtracking, denoted as the GP-ACTION-SET. The second set of actions corresponds to the expansion process of the Large Language Model (LLM), designated as the LLM-EP-ACTION-SET. The third set of actions relates to the backtracking process of the LLM, termed as the LLM-BP-ACTION-SET.

% logistics的domain用PDDL语言描述如下所示：
The domain of logistics is described in PDDL language as follows:
\begin{center}
\fcolorbox{black}{gray!10}{\parbox{.9\linewidth}{
(define (domain logistics-strips) \\
  (:requirements :strips) \\
  (:predicates  (OBJ ?obj)
          (TRUCK ?truck)
                (LOCATION ?loc)
    (AIRPLANE ?airplane)
                (CITY ?city)
                (AIRPORT ?airport)
    (at ?obj ?loc)
    (in ?obj1 ?obj2)
    (in-city ?obj ?city))\\

(:action LOAD-TRUCK\\
  :parameters
   (?obj
    ?truck
    ?loc)\\
  :precondition
   (and (OBJ ?obj) (TRUCK ?truck) (LOCATION ?loc)
   (at ?truck ?loc) (at ?obj ?loc))\\
  :effect
   (and (not (at ?obj ?loc)) (in ?obj ?truck)))\\

(:action LOAD-AIRPLANE\\
  :parameters
   (?obj
    ?airplane
    ?loc)\\
  :precondition
   (and (OBJ ?obj) (AIRPLANE ?airplane) (LOCATION ?loc)
   (at ?obj ?loc) (at ?airplane ?loc))\\
  :effect
   (and (not (at ?obj ?loc)) (in ?obj ?airplane)))\\

(:action UNLOAD-TRUCK\\
  :parameters
   (?obj
    ?truck
    ?loc)\\
  :precondition
   (and (OBJ ?obj) (TRUCK ?truck) (LOCATION ?loc)
        (at ?truck ?loc) (in ?obj ?truck))\\
  :effect
   (and (not (in ?obj ?truck)) (at ?obj ?loc)))\\

(:action UNLOAD-AIRPLANE\\
  :parameters
   (?obj
    ?airplane
    ?loc)\\
  :precondition
   (and (OBJ ?obj) (AIRPLANE ?airplane) (LOCATION ?loc)
        (in ?obj ?airplane) (at ?airplane ?loc))\\
  :effect
   (and (not (in ?obj ?airplane)) (at ?obj ?loc)))\\

(:action DRIVE-TRUCK\\
  :parameters
   (?truck
    ?loc-from
    ?loc-to
    ?city)\\
  :precondition
   (and (TRUCK ?truck) (LOCATION ?loc-from) (LOCATION ?loc-to) (CITY ?city)
   (at ?truck ?loc-from)
   (in-city ?loc-from ?city)
   (in-city ?loc-to ?city))\\
  :effect
   (and (not (at ?truck ?loc-from)) (at ?truck ?loc-to)))\\

(:action FLY-AIRPLANE\\
  :parameters
   (?airplane
    ?loc-from
    ?loc-to)\\
  :precondition
   (and (AIRPLANE ?airplane) (AIRPORT ?loc-from) (AIRPORT ?loc-to)
  (at ?airplane ?loc-from))\\
  :effect
   (and (not (at ?airplane ?loc-from)) (at ?airplane ?loc-to))))

}}
\end{center}

The problem of logistics is described in PDDL language as follows:
\begin{center}
\fcolorbox{black}{gray!10}{\parbox{.9\linewidth}{
(define (problem logistics-02)\\
(:domain logistics-strips)\\
(:objects \\a0\\
          c0 c1\\
          t0 t1\\
          l00 l01 l10 l11\\
          p0 p1 \\
)\\
(:init\\
(AIRPLANE a0)
(CITY c0)
(CITY c1)
(TRUCK t0)
(TRUCK t1)
(LOCATION l00)
(in-city  l00 c0)
(LOCATION l01)
(in-city  l01 c0)
(LOCATION l10)
(in-city  l10 c1)
(LOCATION l11)
(in-city  l11 c1)
(AIRPORT l00)
(AIRPORT l10)
(at a0 l00)
(OBJ p0)
(OBJ p1)
(at t0 l00)
(at t1 l10)
(at p0 l01)
)\\
(:goal\\
(and
(at p0 l11)
)
)\\
)

}}
\end{center}

% 下面我们将逐层展示上述的三类动作的集合，并且对最终输出的规划解进行字体加粗标注。第一层表示最开始拓展的那个动作层，以此类推到最终的目标层，一共展示10层。我们可以从这个具体案例可以观察到以下几点：1.虽然LLM会进行剪枝，但是正确的规划解总是存在于LLM-EP-ACTION-SET中，删去的其实都是一些多余的动作；2.LLM-EP-ACTION-SET的大小都是小于GP-ACTION-SET的；3.在LLM-BP-ACTION-SET中，LLM总是能把正确的规划解动作放在其他大部分动作的前面，虽然并非每次都是第一，但是排位也是相当靠前的。
We will demonstrate the sets of the aforementioned three types of actions layer by layer, and annotate the final output of the planning solution with bold fonts. The first layer represents the initial expansion action layer, and so on up to the final goal layer, totaling 10 layers.From this specific case, we can observe the following points:

\begin{enumerate}
    \item Although LLM undergoes pruning, the correct planning solution always exists within the LLM-EP-ACTION-SET, with the removed actions being redundant. 
    \item The size of the LLM-EP-ACTION-SET is always smaller than that of the GP-ACTION-SET.
    \item In the LLM-BP-ACTION-SET, LLM consistently positions the correct planning solution actions towards the front among most other actions. Although not always the first, they are generally placed near the beginning.
\end{enumerate}

\newpage
\begin{enumerate}
    \item Layer 1\\
    GP-ACTION-SET
        \begin{center}
        \fcolorbox{black}{gray!10}{\parbox{.9\linewidth}{
            DRIVE-TRUCK \{'?truck': 't1', '?loc-from': 'l10', '?loc-to': 'l11', '?city': 'c1'\} \{('at', 't1', 'l11')\} \\
            NoOp \{\} \{('LOCATION', 'l10')\} \\
            NoOp \{\} \{('in-city', 'l01', 'c0')\}\\
            NoOp \{\} \{('CITY', 'c1')\}\\
            NoOp \{\} \{('OBJ', 'p1')\}\\
            NoOp \{\} \{('AIRPLANE', 'a0')\}\\
            NoOp \{\} \{('LOCATION', 'l11')\}\\
            NoOp \{\} \{('at', 't1', 'l10')\}\\
            NoOp \{\} \{('in-city', 'l00', 'c0')\}\\
            DRIVE-TRUCK \{'?truck': 't1', '?loc-from': 'l10', '?loc-to': 'l10', '?city': 'c1'\} \{('at', 't1', 'l10')\}\\
            DRIVE-TRUCK \{'?truck': 't0', '?loc-from': 'l00', '?loc-to': 'l00', '?city': 'c0'\} \{('at', 't0', 'l00')\}\\
            FLY-AIRPLANE \{'?airplane': 'a0', '?loc-from': 'l00', '?loc-to': 'l00'\} \{('at', 'a0', 'l00')\}\\
            NoOp \{\} \{('LOCATION', 'l01')\}\\
            NoOp \{\} \{('in-city', 'l11', 'c1')\}\\
            NoOp \{\} \{('AIRPORT', 'l10')\}\\
            \textbf{DRIVE-TRUCK \{'?truck': 't0', '?loc-from': 'l00', '?loc-to': 'l01', '?city': 'c0'\} \{('at', 't0', 'l01')\}}\\
            NoOp \{\} \{('CITY', 'c0')\}\\
            FLY-AIRPLANE \{'?airplane': 'a0', '?loc-from': 'l00', '?loc-to': 'l10'\} \{('at', 'a0', 'l10')\}\\
            NoOp \{\} \{('at', 'a0', 'l00')\}\\
            NoOp \{\} \{('LOCATION', 'l00')\}\\
            NoOp \{\} \{('TRUCK', 't1')\}\\
            NoOp \{\} \{('OBJ', 'p0')\}\\
            NoOp \{\} \{('TRUCK', 't0')\}\\
            NoOp \{\} \{('AIRPORT', 'l00')\}\\
            NoOp \{\} \{('at', 't0', 'l00')\}\\
            NoOp \{\} \{('in-city', 'l10', 'c1')\}\\
            NoOp \{\} \{('at', 'p0', 'l01')\}
        }}\end{center}
    LLM-EP-ACTION-SET
        \begin{center}
            \fcolorbox{black}{gray!10}{\parbox{.9\linewidth}{
                DRIVE-TRUCK \{'?truck': 't1', '?loc-from': 'l10', '?loc-to': 'l11', '?city': 'c1'\} \{('at', 't1', 'l11')\}\\
                NoOp \{\} \{('LOCATION', 'l10')\}\\
                NoOp \{\} \{('in-city', 'l01', 'c0')\}\\
                NoOp \{\} \{('CITY', 'c1')\}\\
                NoOp \{\} \{('OBJ', 'p1')\}\\
                NoOp \{\} \{('AIRPLANE', 'a0')\}\\
                NoOp \{\} \{('LOCATION', 'l11')\}\\
                NoOp \{\} \{('at', 't1', 'l10')\}\\
                NoOp \{\} \{('in-city', 'l00', 'c0')\}\\
                NoOp \{\} \{('LOCATION', 'l01')\}\\
                NoOp \{\} \{('in-city', 'l11', 'c1')\}\\
                NoOp \{\} \{('AIRPORT', 'l10')\}\\
                \textbf{DRIVE-TRUCK \{'?truck': 't0', '?loc-from': 'l00', '?loc-to': 'l01', '?city': 'c0'\} \{('at', 't0', 'l01')\}}\\
                NoOp \{\} \{('CITY', 'c0')\}\\
                NoOp \{\} \{('at', 'a0', 'l00')\}\\
                NoOp \{\} \{('LOCATION', 'l00')\}\\
                NoOp \{\} \{('TRUCK', 't1')\}\\
                NoOp \{\} \{('OBJ', 'p0')\}\\
                NoOp \{\} \{('TRUCK', 't0')\}\\
                NoOp \{\} \{('AIRPORT', 'l00')\}\\
                NoOp \{\} \{('at', 't0', 'l00')\}\\
                NoOp \{\} \{('in-city', 'l10', 'c1')\}\\
                NoOp \{\} \{('at', 'p0', 'l01')\}\\
            }}\end{center}
    LLM-BP-ACTION-SET
        \begin{center}
            \fcolorbox{black}{gray!10}{\parbox{.9\linewidth}{
                DRIVE-TRUCK \{'?truck': 't1', '?loc-from': 'l10', '?loc-to': 'l11', '?city': 'c1'\} \{('at', 't1', 'l11')\}\\
                \textbf{DRIVE-TRUCK \{'?truck': 't0', '?loc-from': 'l00', '?loc-to': 'l01', '?city': 'c0'\} \{('at', 't0', 'l01')\}}\\
                NoOp \{\} \{('LOCATION', 'l10')\}\\
                NoOp \{\} \{('in-city', 'l01', 'c0')\}\\
                NoOp \{\} \{('CITY', 'c1')\}\\
                NoOp \{\} \{('OBJ', 'p1')\}\\
                NoOp \{\} \{('AIRPLANE', 'a0')\}\\
                NoOp \{\} \{('LOCATION', 'l11')\}\\
                NoOp \{\} \{('at', 't1', 'l10')\}\\
                NoOp \{\} \{('in-city', 'l00', 'c0')\}\\
                NoOp \{\} \{('LOCATION', 'l01')\}\\
                NoOp \{\} \{('in-city', 'l11', 'c1')\}\\
                NoOp \{\} \{('AIRPORT', 'l10')\}\\
                NoOp \{\} \{('CITY', 'c0')\}\\
                NoOp \{\} \{('at', 'a0', 'l00')\}\\
                NoOp \{\} \{('LOCATION', 'l00')\}\\
                NoOp \{\} \{('TRUCK', 't1')\}\\
                NoOp \{\} \{('OBJ', 'p0')\}\\
                NoOp \{\} \{('TRUCK', 't0')\}\\
                NoOp \{\} \{('AIRPORT', 'l00')\}\\
                NoOp \{\} \{('at', 't0', 'l00')\}\\
                NoOp \{\} \{('in-city', 'l10', 'c1')\}\\
                NoOp \{\} \{('at', 'p0', 'l01')\}\\
            }}\end{center}
             
    \item Layer 2\\
    GP-ACTION-SET
        \begin{center}
        \fcolorbox{black}{gray!10}{\parbox{.9\linewidth}{
            NoOp \{\} \{('LOCATION', 'l10')\}\\
            NoOp \{\} \{('at', 'p0', 'l01')\}\\
            NoOp \{\} \{('TRUCK', 't0')\}\\
            FLY-AIRPLANE \{'?airplane': 'a0', '?loc-from': 'l00', '?loc-to': 'l10'\} \{('at', 'a0', 'l10')\}\\
            NoOp \{\} \{('LOCATION', 'l11')\}\\
            NoOp \{\} \{('TRUCK', 't1')\}\\
            NoOp \{\} \{('LOCATION', 'l01')\}\\
            NoOp \{\} \{('OBJ', 'p0')\}\\
            FLY-AIRPLANE \{'?airplane': 'a0', '?loc-from': 'l00', '?loc-to': 'l00'\} \{('at', 'a0', 'l00')\}\\
            DRIVE-TRUCK \{'?truck': 't0', '?loc-from': 'l00', '?loc-to': 'l00', '?city': 'c0'\} \{('at', 't0', 'l00')\}\\
            NoOp \{\} \{('at', 't0', 'l00')\}\\
            DRIVE-TRUCK \{'?truck': 't1', '?loc-from': 'l10', '?loc-to': 'l11', '?city': 'c1'\} \{('at', 't1', 'l11')\}\\
            NoOp \{\} \{('AIRPORT', 'l10')\}\\
            DRIVE-TRUCK \{'?truck': 't1', '?loc-from': 'l11', '?loc-to': 'l10', '?city': 'c1'\} \{('at', 't1', 'l10')\}\\
            DRIVE-TRUCK \{'?truck': 't1', '?loc-from': 'l10', '?loc-to': 'l10', '?city': 'c1'\} \{('at', 't1', 'l10')\}\\
            NoOp \{\} \{('in-city', 'l11', 'c1')\}\\
            NoOp \{\} \{('in-city', 'l00', 'c0')\}\\
            NoOp \{\} \{('LOCATION', 'l00')\}\\
            DRIVE-TRUCK \{'?truck': 't0', '?loc-from': 'l01', '?loc-to': 'l01', '?city': 'c0'\} \{('at', 't0', 'l01')\}\\
            NoOp \{\} \{('at', 'a0', 'l00')\}\\
            NoOp \{\} \{('CITY', 'c0')\}\\
            NoOp \{\} \{('AIRPLANE', 'a0')\}\\
            NoOp \{\} \{('in-city', 'l10', 'c1')\}\\
            \textbf{LOAD-TRUCK \{'?obj': 'p0', '?truck': 't0', '?loc': 'l01'\} \{('in', 'p0', 't0')\}} \\
            NoOp \{\} \{('CITY', 'c1')\}\\
            DRIVE-TRUCK \{'?truck': 't0', '?loc-from': 'l00', '?loc-to': 'l01', '?city': 'c0'\} \{('at', 't0', 'l01')\}\\
            NoOp \{\} \{('OBJ', 'p1')\}\\
            NoOp \{\} \{('AIRPORT', 'l00')\}\\
            NoOp \{\} \{('at', 't0', 'l01')\}\\
            NoOp \{\} \{('at', 't1', 'l11')\}\\
            NoOp \{\} \{('in-city', 'l01', 'c0')\}\\
            NoOp \{\} \{('at', 't1', 'l10')\}\\
            DRIVE-TRUCK \{'?truck': 't0', '?loc-from': 'l01', '?loc-to': 'l00', '?city': 'c0'\} \{('at', 't0', 'l00')\}\\
            DRIVE-TRUCK \{'?truck': 't1', '?loc-from': 'l11', '?loc-to': 'l11', '?city': 'c1'\} \{('at', 't1', 'l11')\}\\
        }}\end{center}
        
    LLM-EP-ACTION-SET
        \begin{center}
        \fcolorbox{black}{gray!10}{\parbox{.9\linewidth}{
            NoOp \{\} \{('LOCATION', 'l10')\}\\
            NoOp \{\} \{('at', 'p0', 'l01')\}\\
            NoOp \{\} \{('TRUCK', 't0')\}\\
            FLY-AIRPLANE \{'?airplane': 'a0', '?loc-from': 'l00', '?loc-to': 'l10'\} \{('at', 'a0', 'l10')\}\\
            NoOp \{\} \{('LOCATION', 'l11')\}\\
            NoOp \{\} \{('TRUCK', 't1')\}\\
            NoOp \{\} \{('LOCATION', 'l01')\}\\
            NoOp \{\} \{('OBJ', 'p0')\}\\
            NoOp \{\} \{('at', 't0', 'l00')\}\\
            DRIVE-TRUCK \{'?truck': 't1', '?loc-from': 'l10', '?loc-to': 'l11', '?city': 'c1'\} \{('at', 't1', 'l11')\}\\
            NoOp \{\} \{('AIRPORT', 'l10')\}\\
            NoOp \{\} \{('in-city', 'l11', 'c1')\}\\
            NoOp \{\} \{('in-city', 'l00', 'c0')\}\\
            NoOp \{\} \{('LOCATION', 'l00')\}\\
            NoOp \{\} \{('at', 'a0', 'l00')\}\\
            NoOp \{\} \{('CITY', 'c0')\}\\
            NoOp \{\} \{('AIRPLANE', 'a0')\}\\
            NoOp \{\} \{('in-city', 'l10', 'c1')\}\\
            \textbf{LOAD-TRUCK \{'?obj': 'p0', '?truck': 't0', '?loc': 'l01'\} \{('in', 'p0', 't0')\}} \\
            NoOp \{\} \{('CITY', 'c1')\}\\
            NoOp \{\} \{('OBJ', 'p1')\}\\
            NoOp \{\} \{('AIRPORT', 'l00')\}\\
            NoOp \{\} \{('at', 't0', 'l01')\}\\
            NoOp \{\} \{('at', 't1', 'l11')\}\\
            NoOp \{\} \{('in-city', 'l01', 'c0')\}\\
            NoOp \{\} \{('at', 't1', 'l10')\}\\
            DRIVE-TRUCK \{'?truck': 't0', '?loc-from': 'l01', '?loc-to': 'l00', '?city': 'c0'\} \{('at', 't0', 'l00')\}\\
        }}\end{center}
        
    LLM-BP-ACTION-SET
        \begin{center}
        \fcolorbox{black}{gray!10}{\parbox{.9\linewidth}{
            \textbf{LOAD-TRUCK \{'?obj': 'p0', '?truck': 't0', '?loc': 'l01'\} \{('in', 'p0', 't0')\}} \\
            DRIVE-TRUCK \{'?truck': 't1', '?loc-from': 'l10', '?loc-to': 'l11', '?city': 'c1'\} \{('at', 't1', 'l11')\}\\
            FLY-AIRPLANE \{'?airplane': 'a0', '?loc-from': 'l00', '?loc-to': 'l10'\} \{('at', 'a0', 'l10')\}\\
            DRIVE-TRUCK \{'?truck': 't0', '?loc-from': 'l01', '?loc-to': 'l00', '?city': 'c0'\} \{('at', 't0', 'l00')\}\\
            NoOp \{\} \{('LOCATION', 'l10')\}\\
            NoOp \{\} \{('at', 'p0', 'l01')\}\\
            NoOp \{\} \{('TRUCK', 't0')\}\\
            NoOp \{\} \{('LOCATION', 'l11')\}\\
            NoOp \{\} \{('TRUCK', 't1')\}\\
            NoOp \{\} \{('LOCATION', 'l01')\}\\
            NoOp \{\} \{('OBJ', 'p0')\}\\
            NoOp \{\} \{('at', 't0', 'l00')\}\\
            NoOp \{\} \{('AIRPORT', 'l10')\}\\
            NoOp \{\} \{('in-city', 'l11', 'c1')\}\\
            NoOp \{\} \{('in-city', 'l00', 'c0')\}\\
            NoOp \{\} \{('LOCATION', 'l00')\}\\
            NoOp \{\} \{('at', 'a0', 'l00')\}\\
            NoOp \{\} \{('CITY', 'c0')\}\\
            NoOp \{\} \{('AIRPLANE', 'a0')\}\\
            NoOp \{\} \{('in-city', 'l10', 'c1')\}\\
            NoOp \{\} \{('CITY', 'c1')\}\\
            NoOp \{\} \{('OBJ', 'p1')\}\\
            NoOp \{\} \{('AIRPORT', 'l00')\}\\
            NoOp \{\} \{('at', 't0', 'l01')\}\\
            NoOp \{\} \{('at', 't1', 'l11')\}\\
            NoOp \{\} \{('in-city', 'l01', 'c0')\}\\
            NoOp \{\} \{('at', 't1', 'l10')\}\\
        }}\end{center}

    \item Layer 3\\
    GP-ACTION-SET
        \begin{center}
        \fcolorbox{black}{gray!10}{\parbox{.9\linewidth}{
            NoOp \{\} \{('AIRPORT', 'l10')\}\\
            NoOp \{\} \{('in', 'p0', 't0')\}\\
            NoOp \{\} \{('at', 'a0', 'l10')\}\\
            NoOp \{\} \{('in-city', 'l11', 'c1')\}\\
            UNLOAD-TRUCK \{'?obj': 'p0', '?truck': 't0', '?loc': 'l01'\} \{('at', 'p0', 'l01')\}\\
            NoOp \{\} \{('OBJ', 'p1')\}\\
            NoOp \{\} \{('OBJ', 'p0')\}\\
            DRIVE-TRUCK \{'?truck': 't1', '?loc-from': 'l11', '?loc-to': 'l10', '?city': 'c1'\} \{('at', 't1', 'l10')\}\\
            DRIVE-TRUCK \{'?truck': 't1', '?loc-from': 'l10', '?loc-to': 'l11', '?city': 'c1'\} \{('at', 't1', 'l11')\}\\
            NoOp \{\} \{('at', 't0', 'l00')\}\\
            NoOp \{\} \{('LOCATION', 'l00')\}\\
            NoOp \{\} \{('AIRPLANE', 'a0')\}\\
            FLY-AIRPLANE \{'?airplane': 'a0', '?loc-from': 'l10', '?loc-to': 'l00'\} \{('at', 'a0', 'l00')\}\\
            NoOp \{\} \{('at', 'p0', 'l01')\}\\
            NoOp \{\} \{('CITY', 'c0')\}\\
            NoOp \{\} \{('LOCATION', 'l10')\}\\
            NoOp \{\} \{('in-city', 'l00', 'c0')\}\\
            DRIVE-TRUCK \{'?truck': 't0', '?loc-from': 'l00', '?loc-to': 'l01', '?city': 'c0'\} \{('at', 't0', 'l01')\}\\
            DRIVE-TRUCK \{'?truck': 't0', '?loc-from': 'l01', '?loc-to': 'l01', '?city': 'c0'\} \{('at', 't0', 'l01')\}\\
            DRIVE-TRUCK \{'?truck': 't1', '?loc-from': 'l11', '?loc-to': 'l11', '?city': 'c1'\} \{('at', 't1', 'l11')\}\\
            DRIVE-TRUCK \{'?truck': 't0', '?loc-from': 'l00', '?loc-to': 'l00', '?city': 'c0'\} \{('at', 't0', 'l00')\}\\
            \textbf{DRIVE-TRUCK \{'?truck': 't0', '?loc-from': 'l01', '?loc-to': 'l00', '?city': 'c0'\} \{('at', 't0', 'l00')\}} \\
            NoOp \{\} \{('LOCATION', 'l01')\}\\
            LOAD-TRUCK \{'?obj': 'p0', '?truck': 't0', '?loc': 'l01'\} \{('in', 'p0', 't0')\}\\
            FLY-AIRPLANE \{'?airplane': 'a0', '?loc-from': 'l00', '?loc-to': 'l10'\} \{('at', 'a0', 'l10')\}\\
            NoOp \{\} \{('at', 't0', 'l01')\}\\
            NoOp \{\} \{('at', 'a0', 'l00')\}\\
            NoOp \{\} \{('AIRPORT', 'l00')\}\\
            NoOp \{\} \{('in-city', 'l01', 'c0')\}\\
            NoOp \{\} \{('CITY', 'c1')\}\\
            FLY-AIRPLANE \{'?airplane': 'a0', '?loc-from': 'l00', '?loc-to': 'l00'\} \{('at', 'a0', 'l00')\}\\
            NoOp \{\} \{('LOCATION', 'l11')\}\\
            NoOp \{\} \{('at', 't1', 'l11')\}\\
            NoOp \{\} \{('TRUCK', 't0')\}\\
            FLY-AIRPLANE \{'?airplane': 'a0', '?loc-from': 'l10', '?loc-to': 'l10'\} \{('at', 'a0', 'l10')\}\\
            NoOp \{\} \{('at', 't1', 'l10')\}\\
            DRIVE-TRUCK \{'?truck': 't1', '?loc-from': 'l10', '?loc-to': 'l10', '?city': 'c1'\} \{('at', 't1', 'l10')\}\\
            NoOp \{\} \{('TRUCK', 't1')\}\\
            NoOp \{\} \{('in-city', 'l10', 'c1')\}\\
        }}\end{center}
    LLM-EP-ACTION-SET
        \begin{center}
        \fcolorbox{black}{gray!10}{\parbox{.9\linewidth}{
            NoOp \{\} \{('AIRPORT', 'l10')\}\\
            NoOp \{\} \{('in', 'p0', 't0')\}\\
            NoOp \{\} \{('at', 'a0', 'l10')\}\\
            NoOp \{\} \{('in-city', 'l11', 'c1')\}\\
            NoOp \{\} \{('OBJ', 'p1')\}\\
            NoOp \{\} \{('OBJ', 'p0')\}\\
            DRIVE-TRUCK \{'?truck': 't1', '?loc-from': 'l10', '?loc-to': 'l11', '?city': 'c1'\} \{('at', 't1', 'l11')\}\\
            NoOp \{\} \{('at', 't0', 'l00')\}\\
            NoOp \{\} \{('LOCATION', 'l00')\}\\
            NoOp \{\} \{('AIRPLANE', 'a0')\}\\
            NoOp \{\} \{('at', 'p0', 'l01')\}\\
            NoOp \{\} \{('CITY', 'c0')\}\\
            NoOp \{\} \{('LOCATION', 'l10')\}\\
            NoOp \{\} \{('in-city', 'l00', 'c0')\}\\
            \textbf{DRIVE-TRUCK \{'?truck': 't0', '?loc-from': 'l01', '?loc-to': 'l00', '?city': 'c0'\} \{('at', 't0', 'l00')\}} \\
            NoOp \{\} \{('LOCATION', 'l01')\}\\
            LOAD-TRUCK \{'?obj': 'p0', '?truck': 't0', '?loc': 'l01'\} \{('in', 'p0', 't0')\}\\
            FLY-AIRPLANE \{'?airplane': 'a0', '?loc-from': 'l00', '?loc-to': 'l10'\} \{('at', 'a0', 'l10')\}\\
            NoOp \{\} \{('at', 't0', 'l01')\}\\
            NoOp \{\} \{('at', 'a0', 'l00')\}\\
            NoOp \{\} \{('AIRPORT', 'l00')\}\\
            NoOp \{\} \{('in-city', 'l01', 'c0')\}\\
            NoOp \{\} \{('CITY', 'c1')\}\\
            NoOp \{\} \{('LOCATION', 'l11')\}\\
            NoOp \{\} \{('at', 't1', 'l11')\}\\
            NoOp \{\} \{('TRUCK', 't0')\}\\
            NoOp \{\} \{('at', 't1', 'l10')\}\\
            NoOp \{\} \{('TRUCK', 't1')\}\\
            NoOp \{\} \{('in-city', 'l10', 'c1')\}\\
        }}\end{center}
    LLM-BP-ACTION-SET
        \begin{center}
        \fcolorbox{black}{gray!10}{\parbox{.9\linewidth}{
            DRIVE-TRUCK \{'?truck': 't1', '?loc-from': 'l10', '?loc-to': 'l11', '?city': 'c1'\} \{('at', 't1', 'l11')\}\\
            \textbf{DRIVE-TRUCK \{'?truck': 't0', '?loc-from': 'l01', '?loc-to': 'l00', '?city': 'c0'\} \{('at', 't0', 'l00')\}} \\
            LOAD-TRUCK \{'?obj': 'p0', '?truck': 't0', '?loc': 'l01'\} \{('in', 'p0', 't0')\}\\
            FLY-AIRPLANE \{'?airplane': 'a0', '?loc-from': 'l00', '?loc-to': 'l10'\} \{('at', 'a0', 'l10')\}\\
            NoOp \{\} \{('AIRPORT', 'l10')\}\\
            NoOp \{\} \{('in', 'p0', 't0')\}\\
            NoOp \{\} \{('at', 'a0', 'l10')\}\\
            NoOp \{\} \{('in-city', 'l11', 'c1')\}\\
            NoOp \{\} \{('OBJ', 'p1')\}\\
            NoOp \{\} \{('OBJ', 'p0')\}\\
            NoOp \{\} \{('at', 't0', 'l00')\}\\
            NoOp \{\} \{('LOCATION', 'l00')\}\\
            NoOp \{\} \{('AIRPLANE', 'a0')\}\\
            NoOp \{\} \{('at', 'p0', 'l01')\}\\
            NoOp \{\} \{('CITY', 'c0')\}\\
            NoOp \{\} \{('LOCATION', 'l10')\}\\
            NoOp \{\} \{('in-city', 'l00', 'c0')\}\\
            NoOp \{\} \{('LOCATION', 'l01')\}\\
            NoOp \{\} \{('at', 't0', 'l01')\}\\
            NoOp \{\} \{('at', 'a0', 'l00')\}\\
            NoOp \{\} \{('AIRPORT', 'l00')\}\\
            NoOp \{\} \{('in-city', 'l01', 'c0')\}\\
            NoOp \{\} \{('CITY', 'c1')\}\\
            NoOp \{\} \{('LOCATION', 'l11')\}\\
            NoOp \{\} \{('at', 't1', 'l11')\}\\
            NoOp \{\} \{('TRUCK', 't0')\}\\
            NoOp \{\} \{('at', 't1', 'l10')\}\\
            NoOp \{\} \{('TRUCK', 't1')\}\\
            NoOp \{\} \{('in-city', 'l10', 'c1')\}\\
        }}\end{center}

    \item Layer 4\\
    GP-ACTION-SET
        \begin{center}
        \fcolorbox{black}{gray!10}{\parbox{.9\linewidth}{
            NoOp \{\} \{('in-city', 'l10', 'c1')\}\\
            NoOp \{\} \{('LOCATION', 'l10')\}\\
            NoOp \{\} \{('in-city', 'l11', 'c1')\}\\
            FLY-AIRPLANE \{'?airplane': 'a0', '?loc-from': 'l10', '?loc-to': 'l00'\} \{('at', 'a0', 'l00')\}\\
            NoOp \{\} \{('TRUCK', 't1')\}\\
            NoOp \{\} \{('LOCATION', 'l11')\}\\
            NoOp \{\} \{('at', 't0', 'l00')\}\\
            NoOp \{\} \{('AIRPORT', 'l00')\}\\
            DRIVE-TRUCK \{'?truck': 't0', '?loc-from': 'l00', '?loc-to': 'l01', '?city': 'c0'\} \{('at', 't0', 'l01')\}\\
            NoOp \{\} \{('OBJ', 'p0')\}\\
            NoOp \{\} \{('in-city', 'l01', 'c0')\}\\
            NoOp \{\} \{('in-city', 'l00', 'c0')\}\\
            NoOp \{\} \{('at', 'a0', 'l10')\}\\
            DRIVE-TRUCK \{'?truck': 't1', '?loc-from': 'l10', '?loc-to': 'l11', '?city': 'c1'\} \{('at', 't1', 'l11')\}\\
            NoOp \{\} \{('CITY', 'c1')\}\\
            FLY-AIRPLANE \{'?airplane': 'a0', '?loc-from': 'l10', '?loc-to': 'l10'\} \{('at', 'a0', 'l10')\}\\
            UNLOAD-TRUCK \{'?obj': 'p0', '?truck': 't0', '?loc': 'l01'\} \{('at', 'p0', 'l01')\}\\
            DRIVE-TRUCK \{'?truck': 't0', '?loc-from': 'l01', '?loc-to': 'l00', '?city': 'c0'\} \{('at', 't0', 'l00')\}\\
            NoOp \{\} \{('at', 't1', 'l11')\}\\
            DRIVE-TRUCK \{'?truck': 't0', '?loc-from': 'l00', '?loc-to': 'l00', '?city': 'c0'\} \{('at', 't0', 'l00')\}\\
            NoOp \{\} \{('OBJ', 'p1')\}\\
            NoOp \{\} \{('at', 't0', 'l01')\}\\
            NoOp \{\} \{('TRUCK', 't0')\}\\
            DRIVE-TRUCK \{'?truck': 't1', '?loc-from': 'l11', '?loc-to': 'l10', '?city': 'c1'\} \{('at', 't1', 'l10')\}\\
            NoOp \{\} \{('AIRPORT', 'l10')\}\\
            LOAD-TRUCK \{'?obj': 'p0', '?truck': 't0', '?loc': 'l01'\} \{('in', 'p0', 't0')\}\\
            NoOp \{\} \{('CITY', 'c0')\}\\
            FLY-AIRPLANE \{'?airplane': 'a0', '?loc-from': 'l00', '?loc-to': 'l00'\} \{('at', 'a0', 'l00')\}\\
            NoOp \{\} \{('LOCATION', 'l01')\}\\
            NoOp \{\} \{('at', 'a0', 'l00')\}\\
            NoOp \{\} \{('at', 'p0', 'l01')\}\\
            DRIVE-TRUCK \{'?truck': 't0', '?loc-from': 'l01', '?loc-to': 'l01', '?city': 'c0'\} \{('at', 't0', 'l01')\}\\
            \textbf{UNLOAD-TRUCK \{'?obj': 'p0', '?truck': 't0', '?loc': 'l00'\} \{('at', 'p0', 'l00')\}} \\
            NoOp \{\} \{('in', 'p0', 't0')\}\\
            NoOp \{\} \{('AIRPLANE', 'a0')\}\\
            NoOp \{\} \{('at', 't1', 'l10')\}\\
            DRIVE-TRUCK \{'?truck': 't1', '?loc-from': 'l11', '?loc-to': 'l11', '?city': 'c1'\} \{('at', 't1', 'l11')\}\\
            NoOp \{\} \{('LOCATION', 'l00')\}\\
            FLY-AIRPLANE \{'?airplane': 'a0', '?loc-from': 'l00', '?loc-to': 'l10'\} \{('at', 'a0', 'l10')\}\\
            DRIVE-TRUCK \{'?truck': 't1', '?loc-from': 'l10', '?loc-to': 'l10', '?city': 'c1'\} \{('at', 't1', 'l10')\}\\
        }}\end{center}
    LLM-EP-ACTION-SET
        \begin{center}
        \fcolorbox{black}{gray!10}{\parbox{.9\linewidth}{
            NoOp \{\} \{('in-city', 'l10', 'c1')\}\\
            NoOp \{\} \{('LOCATION', 'l10')\}\\
            NoOp \{\} \{('in-city', 'l11', 'c1')\}\\
            NoOp \{\} \{('TRUCK', 't1')\}\\
            NoOp \{\} \{('LOCATION', 'l11')\}\\
            NoOp \{\} \{('at', 't0', 'l00')\}\\
            NoOp \{\} \{('AIRPORT', 'l00')\}\\
            NoOp \{\} \{('OBJ', 'p0')\}\\
            NoOp \{\} \{('in-city', 'l01', 'c0')\}\\
            NoOp \{\} \{('in-city', 'l00', 'c0')\}\\
            NoOp \{\} \{('at', 'a0', 'l10')\}\\
            DRIVE-TRUCK \{'?truck': 't1', '?loc-from': 'l10', '?loc-to': 'l11', '?city': 'c1'\} \{('at', 't1', 'l11')\}\\
            NoOp \{\} \{('CITY', 'c1')\}\\
            DRIVE-TRUCK \{'?truck': 't0', '?loc-from': 'l01', '?loc-to': 'l00', '?city': 'c0'\} \{('at', 't0', 'l00')\}\\
            NoOp \{\} \{('at', 't1', 'l11')\}\\
            NoOp \{\} \{('OBJ', 'p1')\}\\
            NoOp \{\} \{('at', 't0', 'l01')\}\\
            NoOp \{\} \{('TRUCK', 't0')\}\\
            NoOp \{\} \{('AIRPORT', 'l10')\}\\
            LOAD-TRUCK \{'?obj': 'p0', '?truck': 't0', '?loc': 'l01'\} \{('in', 'p0', 't0')\}\\
            NoOp \{\} \{('CITY', 'c0')\}\\
            NoOp \{\} \{('LOCATION', 'l01')\}\\
            NoOp \{\} \{('at', 'a0', 'l00')\}\\
            NoOp \{\} \{('at', 'p0', 'l01')\}\\
            \textbf{UNLOAD-TRUCK \{'?obj': 'p0', '?truck': 't0', '?loc': 'l00'\} \{('at', 'p0', 'l00')\}} \\
            NoOp \{\} \{('in', 'p0', 't0')\}\\
            NoOp \{\} \{('AIRPLANE', 'a0')\}\\
            NoOp \{\} \{('at', 't1', 'l10')\}\\
            NoOp \{\} \{('LOCATION', 'l00')\}\\
            FLY-AIRPLANE \{'?airplane': 'a0', '?loc-from': 'l00', '?loc-to': 'l10'\} \{('at', 'a0', 'l10')\}\\
        }}\end{center}
    LLM-BP-ACTION-SET
        \begin{center}
        \fcolorbox{black}{gray!10}{\parbox{.9\linewidth}{
            DRIVE-TRUCK \{'?truck': 't1', '?loc-from': 'l10', '?loc-to': 'l11', '?city': 'c1'\} \{('at', 't1', 'l11')\}\\
            LOAD-TRUCK \{'?obj': 'p0', '?truck': 't0', '?loc': 'l01'\} \{('in', 'p0', 't0')\}\\
            DRIVE-TRUCK \{'?truck': 't0', '?loc-from': 'l01', '?loc-to': 'l00', '?city': 'c0'\} \{('at', 't0', 'l00')\}\\
            \textbf{UNLOAD-TRUCK \{'?obj': 'p0', '?truck': 't0', '?loc': 'l00'\} \{('at', 'p0', 'l00')\}} \\
            FLY-AIRPLANE \{'?airplane': 'a0', '?loc-from': 'l00', '?loc-to': 'l10'\} \{('at', 'a0', 'l10')\}\\
            NoOp \{\} \{('in-city', 'l10', 'c1')\}\\
            NoOp \{\} \{('LOCATION', 'l10')\}\\
            NoOp \{\} \{('in-city', 'l11', 'c1')\}\\
            NoOp \{\} \{('TRUCK', 't1')\}\\
            NoOp \{\} \{('LOCATION', 'l11')\}\\
            NoOp \{\} \{('at', 't0', 'l00')\}\\
            NoOp \{\} \{('AIRPORT', 'l00')\}\\
            NoOp \{\} \{('OBJ', 'p0')\}\\
            NoOp \{\} \{('in-city', 'l01', 'c0')\}\\
            NoOp \{\} \{('in-city', 'l00', 'c0')\}\\
            NoOp \{\} \{('at', 'a0', 'l10')\}\\
            NoOp \{\} \{('CITY', 'c1')\}\\
            NoOp \{\} \{('at', 't1', 'l11')\}\\
            NoOp \{\} \{('OBJ', 'p1')\}\\
            NoOp \{\} \{('at', 't0', 'l01')\}\\
            NoOp \{\} \{('TRUCK', 't0')\}\\
            NoOp \{\} \{('AIRPORT', 'l10')\}\\
            NoOp \{\} \{('CITY', 'c0')\}\\
            NoOp \{\} \{('LOCATION', 'l01')\}\\
            NoOp \{\} \{('at', 'a0', 'l00')\}\\
            NoOp \{\} \{('at', 'p0', 'l01')\}\\
            NoOp \{\} \{('in', 'p0', 't0')\}\\
            NoOp \{\} \{('AIRPLANE', 'a0')\}\\
            NoOp \{\} \{('at', 't1', 'l10')\}\\
            NoOp \{\} \{('LOCATION', 'l00')\}\\
        }}\end{center}

    \item Layer 5\\
    GP-ACTION-SET
        \begin{center}
        \fcolorbox{black}{gray!10}{\parbox{.9\linewidth}{
            UNLOAD-TRUCK \{'?obj': 'p0', '?truck': 't0', '?loc': 'l01'\} \{('at', 'p0', 'l01')\}\\
            NoOp \{\} \{('AIRPORT', 'l00')\}\\
            NoOp \{\} \{('LOCATION', 'l10')\}\\
            LOAD-TRUCK \{'?obj': 'p0', '?truck': 't0', '?loc': 'l01'\} \{('in', 'p0', 't0')\}\\
            NoOp \{\} \{('in', 'p0', 't0')\}\\
            NoOp \{\} \{('in-city', 'l00', 'c0')\}\\
            NoOp \{\} \{('CITY', 'c0')\}\\
            NoOp \{\} \{('at', 't1', 'l11')\}\\
            NoOp \{\} \{('LOCATION', 'l00')\}\\
            NoOp \{\} \{('at', 'a0', 'l10')\}\\
            NoOp \{\} \{('AIRPORT', 'l10')\}\\
            NoOp \{\} \{('at', 'p0', 'l00')\}\\
            DRIVE-TRUCK \{'?truck': 't1', '?loc-from': 'l11', '?loc-to': 'l11', '?city': 'c1'\} \{('at', 't1', 'l11')\}\\
            NoOp \{\} \{('TRUCK', 't1')\}\\
            NoOp \{\} \{('at', 't1', 'l10')\}\\
            DRIVE-TRUCK \{'?truck': 't1', '?loc-from': 'l10', '?loc-to': 'l11', '?city': 'c1'\} \{('at', 't1', 'l11')\}\\
            NoOp \{\} \{('in-city', 'l11', 'c1')\}\\
            FLY-AIRPLANE \{'?airplane': 'a0', '?loc-from': 'l10', '?loc-to': 'l10'\} \{('at', 'a0', 'l10')\}\\
            NoOp \{\} \{('OBJ', 'p0')\}\\
            FLY-AIRPLANE \{'?airplane': 'a0', '?loc-from': 'l00', '?loc-to': 'l10'\} \{('at', 'a0', 'l10')\}\\
            UNLOAD-TRUCK \{'?obj': 'p0', '?truck': 't0', '?loc': 'l00'\} \{('at', 'p0', 'l00')\}\\
            DRIVE-TRUCK \{'?truck': 't0', '?loc-from': 'l01', '?loc-to': 'l00', '?city': 'c0'\} \{('at', 't0', 'l00')\}\\
            LOAD-TRUCK \{'?obj': 'p0', '?truck': 't0', '?loc': 'l00'\} \{('in', 'p0', 't0')\}\\
            DRIVE-TRUCK \{'?truck': 't1', '?loc-from': 'l10', '?loc-to': 'l10', '?city': 'c1'\} \{('at', 't1', 'l10')\}\\
            DRIVE-TRUCK \{'?truck': 't0', '?loc-from': 'l00', '?loc-to': 'l00', '?city': 'c0'\} \{('at', 't0', 'l00')\}\\
            NoOp \{\} \{('at', 'a0', 'l00')\}\\
            NoOp \{\} \{('at', 'p0', 'l01')\}\\
            NoOp \{\} \{('at', 't0', 'l00')\}\\
            NoOp \{\} \{('at', 't0', 'l01')\}\\
            NoOp \{\} \{('LOCATION', 'l11')\}\\
            FLY-AIRPLANE \{'?airplane': 'a0', '?loc-from': 'l00', '?loc-to': 'l00'\} \{('at', 'a0', 'l00')\}\\
            DRIVE-TRUCK \{'?truck': 't0', '?loc-from': 'l01', '?loc-to': 'l01', '?city': 'c0'\} \{('at', 't0', 'l01')\}\\
            NoOp \{\} \{('in-city', 'l01', 'c0')\}\\
            NoOp \{\} \{('OBJ', 'p1')\}\\
            FLY-AIRPLANE \{'?airplane': 'a0', '?loc-from': 'l10', '?loc-to': 'l00'\} \{('at', 'a0', 'l00')\}\\
            NoOp \{\} \{('in-city', 'l10', 'c1')\}\\
            NoOp \{\} \{('AIRPLANE', 'a0')\}\\
            \textbf{LOAD-AIRPLANE \{'?obj': 'p0', '?airplane': 'a0', '?loc': 'l00'\} \{('in', 'p0', 'a0')\}} \\
            DRIVE-TRUCK \{'?truck': 't0', '?loc-from': 'l00', '?loc-to': 'l01', '?city': 'c0'\} \{('at', 't0', 'l01')\}\\
            NoOp \{\} \{('LOCATION', 'l01')\}\\
            DRIVE-TRUCK \{'?truck': 't1', '?loc-from': 'l11', '?loc-to': 'l10', '?city': 'c1'\} \{('at', 't1', 'l10')\}\\
            NoOp \{\} \{('TRUCK', 't0')\}\\
            NoOp \{\} \{('CITY', 'c1')\}\\
        }}\end{center}
    LLM-EP-ACTION-SET
        \begin{center}
        \fcolorbox{black}{gray!10}{\parbox{.9\linewidth}{
            NoOp \{\} \{('AIRPORT', 'l00')\}\\
            NoOp \{\} \{('LOCATION', 'l10')\}\\
            LOAD-TRUCK \{'?obj': 'p0', '?truck': 't0', '?loc': 'l01'\} \{('in', 'p0', 't0')\}\\
            NoOp \{\} \{('in', 'p0', 't0')\}\\
            NoOp \{\} \{('in-city', 'l00', 'c0')\}\\
            NoOp \{\} \{('CITY', 'c0')\}\\
            NoOp \{\} \{('at', 't1', 'l11')\}\\
            NoOp \{\} \{('LOCATION', 'l00')\}\\
            NoOp \{\} \{('at', 'a0', 'l10')\}\\
            NoOp \{\} \{('AIRPORT', 'l10')\}\\
            NoOp \{\} \{('at', 'p0', 'l00')\}\\
            NoOp \{\} \{('TRUCK', 't1')\}\\
            NoOp \{\} \{('at', 't1', 'l10')\}\\
            DRIVE-TRUCK \{'?truck': 't1', '?loc-from': 'l10', '?loc-to': 'l11', '?city': 'c1'\} \{('at', 't1', 'l11')\}\\
            NoOp \{\} \{('in-city', 'l11', 'c1')\}\\
            NoOp \{\} \{('OBJ', 'p0')\}\\
            FLY-AIRPLANE \{'?airplane': 'a0', '?loc-from': 'l00', '?loc-to': 'l10'\} \{('at', 'a0', 'l10')\}\\
            UNLOAD-TRUCK \{'?obj': 'p0', '?truck': 't0', '?loc': 'l00'\} \{('at', 'p0', 'l00')\}\\
            DRIVE-TRUCK \{'?truck': 't0', '?loc-from': 'l01', '?loc-to': 'l00', '?city': 'c0'\} \{('at', 't0', 'l00')\}\\
            NoOp \{\} \{('at', 'a0', 'l00')\}\\
            NoOp \{\} \{('at', 'p0', 'l01')\}\\
            NoOp \{\} \{('at', 't0', 'l00')\}\\
            NoOp \{\} \{('at', 't0', 'l01')\}\\
            NoOp \{\} \{('LOCATION', 'l11')\}\\
            NoOp \{\} \{('in-city', 'l01', 'c0')\}\\
            NoOp \{\} \{('OBJ', 'p1')\}\\
            NoOp \{\} \{('in-city', 'l10', 'c1')\}\\
            NoOp \{\} \{('AIRPLANE', 'a0')\}\\
            \textbf{LOAD-AIRPLANE \{'?obj': 'p0', '?airplane': 'a0', '?loc': 'l00'\} \{('in', 'p0', 'a0')\}} \\
            NoOp \{\} \{('LOCATION', 'l01')\}\\
            NoOp \{\} \{('TRUCK', 't0')\}\\
            NoOp \{\} \{('CITY', 'c1')\}\\
        }}\end{center}
    LLM-BP-ACTION-SET
        \begin{center}
        \fcolorbox{black}{gray!10}{\parbox{.9\linewidth}{
            DRIVE-TRUCK \{'?truck': 't1', '?loc-from': 'l10', '?loc-to': 'l11', '?city': 'c1'\} \{('at', 't1', 'l11')\}\\
            DRIVE-TRUCK \{'?truck': 't0', '?loc-from': 'l01', '?loc-to': 'l00', '?city': 'c0'\} \{('at', 't0', 'l00')\}\\
            UNLOAD-TRUCK \{'?obj': 'p0', '?truck': 't0', '?loc': 'l00'\} \{('at', 'p0', 'l00')\}\\
            LOAD-TRUCK \{'?obj': 'p0', '?truck': 't0', '?loc': 'l01'\} \{('in', 'p0', 't0')\}\\
            FLY-AIRPLANE \{'?airplane': 'a0', '?loc-from': 'l00', '?loc-to': 'l10'\} \{('at', 'a0', 'l10')\}\\
            \textbf{LOAD-AIRPLANE \{'?obj': 'p0', '?airplane': 'a0', '?loc': 'l00'\} \{('in', 'p0', 'a0')\}} \\
            NoOp \{\} \{('AIRPORT', 'l00')\}\\
            NoOp \{\} \{('LOCATION', 'l10')\}\\
            NoOp \{\} \{('in', 'p0', 't0')\}\\
            NoOp \{\} \{('in-city', 'l00', 'c0')\}\\
            NoOp \{\} \{('CITY', 'c0')\}\\
            NoOp \{\} \{('at', 't1', 'l11')\}\\
            NoOp \{\} \{('LOCATION', 'l00')\}\\
            NoOp \{\} \{('at', 'a0', 'l10')\}\\
            NoOp \{\} \{('AIRPORT', 'l10')\}\\
            NoOp \{\} \{('at', 'p0', 'l00')\}\\
            NoOp \{\} \{('TRUCK', 't1')\}\\
            NoOp \{\} \{('at', 't1', 'l10')\}\\
            NoOp \{\} \{('in-city', 'l11', 'c1')\}\\
            NoOp \{\} \{('OBJ', 'p0')\}\\
            NoOp \{\} \{('at', 'a0', 'l00')\}\\
            NoOp \{\} \{('at', 'p0', 'l01')\}\\
            NoOp \{\} \{('at', 't0', 'l00')\}\\
            NoOp \{\} \{('at', 't0', 'l01')\}\\
            NoOp \{\} \{('LOCATION', 'l11')\}\\
            NoOp \{\} \{('in-city', 'l01', 'c0')\}\\
            NoOp \{\} \{('OBJ', 'p1')\}\\
            NoOp \{\} \{('in-city', 'l10', 'c1')\}\\
            NoOp \{\} \{('AIRPLANE', 'a0')\}\\
            NoOp \{\} \{('LOCATION', 'l01')\}\\
            NoOp \{\} \{('TRUCK', 't0')\}\\
            NoOp \{\} \{('CITY', 'c1')\}\\
        }}\end{center}

    \item Layer 6\\
    GP-ACTION-SET
        \begin{center}
        \fcolorbox{black}{gray!10}{\parbox{.9\linewidth}{
            FLY-AIRPLANE \{'?airplane': 'a0', '?loc-from': 'l00', '?loc-to': 'l00'\} \{('at', 'a0', 'l00')\}\\
            LOAD-AIRPLANE \{'?obj': 'p0', '?airplane': 'a0', '?loc': 'l00'\} \{('in', 'p0', 'a0')\}\\
            NoOp \{\} \{('LOCATION', 'l01')\}\\
            LOAD-TRUCK \{'?obj': 'p0', '?truck': 't0', '?loc': 'l01'\} \{('in', 'p0', 't0')\}\\
            DRIVE-TRUCK \{'?truck': 't0', '?loc-from': 'l00', '?loc-to': 'l01', '?city': 'c0'\} \{('at', 't0', 'l01')\}\\
            DRIVE-TRUCK \{'?truck': 't0', '?loc-from': 'l01', '?loc-to': 'l00', '?city': 'c0'\} \{('at', 't0', 'l00')\}\\
            NoOp \{\} \{('AIRPLANE', 'a0')\}\\
            NoOp \{\} \{('at', 't1', 'l10')\}\\
            NoOp \{\} \{('in-city', 'l10', 'c1')\}\\
            DRIVE-TRUCK \{'?truck': 't1', '?loc-from': 'l11', '?loc-to': 'l11', '?city': 'c1'\} \{('at', 't1', 'l11')\}\\
            NoOp \{\} \{('at', 't1', 'l11')\}\\
            NoOp \{\} \{('at', 'p0', 'l01')\}\\
            NoOp \{\} \{('in-city', 'l00', 'c0')\}\\
            NoOp \{\} \{('at', 't0', 'l00')\}\\
            NoOp \{\} \{('LOCATION', 'l00')\}\\
            NoOp \{\} \{('AIRPORT', 'l10')\}\\
            NoOp \{\} \{('at', 'p0', 'l00')\}\\
            FLY-AIRPLANE \{'?airplane': 'a0', '?loc-from': 'l10', '?loc-to': 'l00'\} \{('at', 'a0', 'l00')\}\\
            UNLOAD-AIRPLANE \{'?obj': 'p0', '?airplane': 'a0', '?loc': 'l00'\} \{('at', 'p0', 'l00')\}\\
            NoOp \{\} \{('AIRPORT', 'l00')\}\\
            DRIVE-TRUCK \{'?truck': 't0', '?loc-from': 'l01', '?loc-to': 'l01', '?city': 'c0'\} \{('at', 't0', 'l01')\}\\
            NoOp \{\} \{('in', 'p0', 'a0')\}\\
            NoOp \{\} \{('TRUCK', 't0')\}\\
            NoOp \{\} \{('in', 'p0', 't0')\}\\
            NoOp \{\} \{('at', 'a0', 'l00')\}\\
            NoOp \{\} \{('in-city', 'l11', 'c1')\}\\
            NoOp \{\} \{('at', 'a0', 'l10')\}\\
            NoOp \{\} \{('LOCATION', 'l10')\}\\
            NoOp \{\} \{('OBJ', 'p1')\}\\
            UNLOAD-TRUCK \{'?obj': 'p0', '?truck': 't0', '?loc': 'l01'\} \{('at', 'p0', 'l01')\}\\
            NoOp \{\} \{('in-city', 'l01', 'c0')\}\\
            LOAD-TRUCK \{'?obj': 'p0', '?truck': 't0', '?loc': 'l00'\} \{('in', 'p0', 't0')\}\\
            UNLOAD-TRUCK \{'?obj': 'p0', '?truck': 't0', '?loc': 'l00'\} \{('at', 'p0', 'l00')\}\\
            DRIVE-TRUCK \{'?truck': 't0', '?loc-from': 'l00', '?loc-to': 'l00', '?city': 'c0'\} \{('at', 't0', 'l00')\}\\
            NoOp \{\} \{('TRUCK', 't1')\}\\
            FLY-AIRPLANE \{'?airplane': 'a0', '?loc-from': 'l10', '?loc-to': 'l10'\} \{('at', 'a0', 'l10')\}\\
            NoOp \{\} \{('OBJ', 'p0')\}\\
            NoOp \{\} \{('LOCATION', 'l11')\}\\
            DRIVE-TRUCK \{'?truck': 't1', '?loc-from': 'l11', '?loc-to': 'l10', '?city': 'c1'\} \{('at', 't1', 'l10')\}\\
            NoOp \{\} \{('CITY', 'c0')\}\\
            NoOp \{\} \{('CITY', 'c1')\}\\
            DRIVE-TRUCK \{'?truck': 't1', '?loc-from': 'l10', '?loc-to': 'l11', '?city': 'c1'\} \{('at', 't1', 'l11')\}\\
            \textbf{FLY-AIRPLANE \{'?airplane': 'a0', '?loc-from': 'l00', '?loc-to': 'l10'\} \{('at', 'a0', 'l10')\}} \\
            DRIVE-TRUCK \{'?truck': 't1', '?loc-from': 'l10', '?loc-to': 'l10', '?city': 'c1'\} \{('at', 't1', 'l10')\}\\
            NoOp \{\} \{('at', 't0', 'l01')\}\\
        }}\end{center}
    LLM-EP-ACTION-SET
        \begin{center}
        \fcolorbox{black}{gray!10}{\parbox{.9\linewidth}{
            LOAD-AIRPLANE \{'?obj': 'p0', '?airplane': 'a0', '?loc': 'l00'\} \{('in', 'p0', 'a0')\}\\
            NoOp \{\} \{('LOCATION', 'l01')\}\\
            DRIVE-TRUCK \{'?truck': 't0', '?loc-from': 'l01', '?loc-to': 'l00', '?city': 'c0'\} \{('at', 't0', 'l00')\}\\
            NoOp \{\} \{('AIRPLANE', 'a0')\}\\
            NoOp \{\} \{('at', 't1', 'l10')\}\\
            NoOp \{\} \{('in-city', 'l10', 'c1')\}\\
            NoOp \{\} \{('at', 't1', 'l11')\}\\
            NoOp \{\} \{('at', 'p0', 'l01')\}\\
            NoOp \{\} \{('in-city', 'l00', 'c0')\}\\
            NoOp \{\} \{('at', 't0', 'l00')\}\\
            NoOp \{\} \{('LOCATION', 'l00')\}\\
            NoOp \{\} \{('AIRPORT', 'l10')\}\\
            NoOp \{\} \{('at', 'p0', 'l00')\}\\
            NoOp \{\} \{('AIRPORT', 'l00')\}\\
            NoOp \{\} \{('in', 'p0', 'a0')\}\\
            NoOp \{\} \{('TRUCK', 't0')\}\\
            NoOp \{\} \{('in', 'p0', 't0')\}\\
            NoOp \{\} \{('at', 'a0', 'l00')\}\\
            NoOp \{\} \{('in-city', 'l11', 'c1')\}\\
            NoOp \{\} \{('at', 'a0', 'l10')\}\\
            NoOp \{\} \{('LOCATION', 'l10')\}\\
            NoOp \{\} \{('OBJ', 'p1')\}\\
            NoOp \{\} \{('in-city', 'l01', 'c0')\}\\
            UNLOAD-TRUCK \{'?obj': 'p0', '?truck': 't0', '?loc': 'l00'\} \{('at', 'p0', 'l00')\}\\
            NoOp \{\} \{('TRUCK', 't1')\}\\
            NoOp \{\} \{('OBJ', 'p0')\}\\
            NoOp \{\} \{('LOCATION', 'l11')\}\\
            NoOp \{\} \{('CITY', 'c0')\}\\
            NoOp \{\} \{('CITY', 'c1')\}\\
            DRIVE-TRUCK \{'?truck': 't1', '?loc-from': 'l10', '?loc-to': 'l11', '?city': 'c1'\} \{('at', 't1', 'l11')\}\\
            \textbf{FLY-AIRPLANE \{'?airplane': 'a0', '?loc-from': 'l00', '?loc-to': 'l10'\} \{('at', 'a0', 'l10')\}} \\
            NoOp \{\} \{('at', 't0', 'l01')\}\\
        }}\end{center}
    LLM-BP-ACTION-SET
        \begin{center}
        \fcolorbox{black}{gray!10}{\parbox{.9\linewidth}{
            LOAD-AIRPLANE \{'?obj': 'p0', '?airplane': 'a0', '?loc': 'l00'\} \{('in', 'p0', 'a0')\}\\
            \textbf{FLY-AIRPLANE \{'?airplane': 'a0', '?loc-from': 'l00', '?loc-to': 'l10'\} \{('at', 'a0', 'l10')\}} \\
            DRIVE-TRUCK \{'?truck': 't0', '?loc-from': 'l01', '?loc-to': 'l00', '?city': 'c0'\} \{('at', 't0', 'l00')\}\\
            DRIVE-TRUCK \{'?truck': 't1', '?loc-from': 'l10', '?loc-to': 'l11', '?city': 'c1'\} \{('at', 't1', 'l11')\}\\
            UNLOAD-TRUCK \{'?obj': 'p0', '?truck': 't0', '?loc': 'l00'\} \{('at', 'p0', 'l00')\}\\
            NoOp \{\} \{('LOCATION', 'l01')\}\\
            NoOp \{\} \{('AIRPLANE', 'a0')\}\\
            NoOp \{\} \{('at', 't1', 'l10')\}\\
            NoOp \{\} \{('in-city', 'l10', 'c1')\}\\
            NoOp \{\} \{('at', 't1', 'l11')\}\\
            NoOp \{\} \{('at', 'p0', 'l01')\}\\
            NoOp \{\} \{('in-city', 'l00', 'c0')\}\\
            NoOp \{\} \{('at', 't0', 'l00')\}\\
            NoOp \{\} \{('LOCATION', 'l00')\}\\
            NoOp \{\} \{('AIRPORT', 'l10')\}\\
            NoOp \{\} \{('at', 'p0', 'l00')\}\\
            NoOp \{\} \{('AIRPORT', 'l00')\}\\
            NoOp \{\} \{('in', 'p0', 'a0')\}\\
            NoOp \{\} \{('TRUCK', 't0')\}\\
            NoOp \{\} \{('in', 'p0', 't0')\}\\
            NoOp \{\} \{('at', 'a0', 'l00')\}\\
            NoOp \{\} \{('in-city', 'l11', 'c1')\}\\
            NoOp \{\} \{('at', 'a0', 'l10')\}\\
            NoOp \{\} \{('LOCATION', 'l10')\}\\
            NoOp \{\} \{('OBJ', 'p1')\}\\
            NoOp \{\} \{('in-city', 'l01', 'c0')\}\\
            NoOp \{\} \{('TRUCK', 't1')\}\\
            NoOp \{\} \{('OBJ', 'p0')\}\\
            NoOp \{\} \{('LOCATION', 'l11')\}\\
            NoOp \{\} \{('CITY', 'c0')\}\\
            NoOp \{\} \{('CITY', 'c1')\}\\
            NoOp \{\} \{('at', 't0', 'l01')\}\\
        }}\end{center}

    \item Layer 7\\
    GP-ACTION-SET
        \begin{center}
        \fcolorbox{black}{gray!10}{\parbox{.9\linewidth}{
            DRIVE-TRUCK \{'?truck': 't1', '?loc-from': 'l11', '?loc-to': 'l10', '?city': 'c1'\} \{('at', 't1', 'l10')\}\\
            FLY-AIRPLANE \{'?airplane': 'a0', '?loc-from': 'l10', '?loc-to': 'l10'\} \{('at', 'a0', 'l10')\}\\
            NoOp \{\} \{('CITY', 'c0')\}\\
            DRIVE-TRUCK \{'?truck': 't0', '?loc-from': 'l00', '?loc-to': 'l00', '?city': 'c0'\} \{('at', 't0', 'l00')\}\\
            NoOp \{\} \{('in-city', 'l01', 'c0')\}\\
            NoOp \{\} \{('AIRPLANE', 'a0')\}\\
            DRIVE-TRUCK \{'?truck': 't0', '?loc-from': 'l00', '?loc-to': 'l01', '?city': 'c0'\} \{('at', 't0', 'l01')\}\\
            NoOp \{\} \{('LOCATION', 'l00')\}\\
            DRIVE-TRUCK \{'?truck': 't0', '?loc-from': 'l01', '?loc-to': 'l01', '?city': 'c0'\} \{('at', 't0', 'l01')\}\\
            UNLOAD-TRUCK \{'?obj': 'p0', '?truck': 't0', '?loc': 'l01'\} \{('at', 'p0', 'l01')\}\\
            FLY-AIRPLANE \{'?airplane': 'a0', '?loc-from': 'l00', '?loc-to': 'l10'\} \{('at', 'a0', 'l10')\}\\
            NoOp \{\} \{('in', 'p0', 't0')\}\\
            NoOp \{\} \{('in-city', 'l10', 'c1')\}\\
            NoOp \{\} \{('LOCATION', 'l01')\}\\
            DRIVE-TRUCK \{'?truck': 't1', '?loc-from': 'l10', '?loc-to': 'l11', '?city': 'c1'\} \{('at', 't1', 'l11')\}\\
            NoOp \{\} \{('AIRPORT', 'l00')\}\\
            NoOp \{\} \{('LOCATION', 'l11')\}\\
            NoOp \{\} \{('TRUCK', 't1')\}\\
            NoOp \{\} \{('in', 'p0', 'a0')\}\\
            UNLOAD-TRUCK \{'?obj': 'p0', '?truck': 't0', '?loc': 'l00'\} \{('at', 'p0', 'l00')\}\\
            NoOp \{\} \{('LOCATION', 'l10')\}\\
            NoOp \{\} \{('at', 'p0', 'l01')\}\\
            NoOp \{\} \{('CITY', 'c1')\}\\
            LOAD-TRUCK \{'?obj': 'p0', '?truck': 't0', '?loc': 'l00'\} \{('in', 'p0', 't0')\}\\
            NoOp \{\} \{('at', 't0', 'l00')\}\\
            NoOp \{\} \{('TRUCK', 't0')\}\\
            NoOp \{\} \{('at', 't0', 'l01')\}\\
            NoOp \{\} \{('at', 't1', 'l10')\}\\
            NoOp \{\} \{('AIRPORT', 'l10')\}\\
            NoOp \{\} \{('OBJ', 'p0')\}\\
            \textbf{UNLOAD-AIRPLANE \{'?obj': 'p0', '?airplane': 'a0', '?loc': 'l10'\} \{('at', 'p0', 'l10')\}} \\
        }}\end{center}
        \begin{center}
        \fcolorbox{black}{gray!10}{\parbox{.9\linewidth}{
            NoOp \{\} \{('at', 't1', 'l11')\}\\
            DRIVE-TRUCK \{'?truck': 't0', '?loc-from': 'l01', '?loc-to': 'l00', '?city': 'c0'\} \{('at', 't0', 'l00')\}\\
            UNLOAD-AIRPLANE \{'?obj': 'p0', '?airplane': 'a0', '?loc': 'l00'\} \{('at', 'p0', 'l00')\} \\
            NoOp \{\} \{('at', 'a0', 'l10')\}\\
            DRIVE-TRUCK \{'?truck': 't1', '?loc-from': 'l10', '?loc-to': 'l10', '?city': 'c1'\} \{('at', 't1', 'l10')\} \\
            NoOp \{\} \{('at', 'p0', 'l00')\}\\
            NoOp \{\} \{('at', 'a0', 'l00')\}\\
            NoOp \{\} \{('OBJ', 'p1')\}\\
            DRIVE-TRUCK \{'?truck': 't1', '?loc-from': 'l11', '?loc-to': 'l11', '?city': 'c1'\} \{('at', 't1', 'l11')\}\\
            LOAD-TRUCK \{'?obj': 'p0', '?truck': 't0', '?loc': 'l01'\} \{('in', 'p0', 't0')\}\\
            FLY-AIRPLANE \{'?airplane': 'a0', '?loc-from': 'l10', '?loc-to': 'l00'\} \{('at', 'a0', 'l00')\}\\
            NoOp \{\} \{('in-city', 'l11', 'c1')\}\\
            NoOp \{\} \{('in-city', 'l00', 'c0')\}\\
            LOAD-AIRPLANE \{'?obj': 'p0', '?airplane': 'a0', '?loc': 'l00'\} \{('in', 'p0', 'a0')\}\\
            FLY-AIRPLANE \{'?airplane': 'a0', '?loc-from': 'l00', '?loc-to': 'l00'\} \{('at', 'a0', 'l00')\}\\
        }}\end{center}

    LLM-EP-ACTION-SET
        \begin{center}
        \fcolorbox{black}{gray!10}{\parbox{.9\linewidth}{
            NoOp \{\} \{('CITY', 'c0')\}\\
            NoOp \{\} \{('in-city', 'l01', 'c0')\}\\
            NoOp \{\} \{('AIRPLANE', 'a0')\}\\
            NoOp \{\} \{('LOCATION', 'l00')\}\\
            FLY-AIRPLANE \{'?airplane': 'a0', '?loc-from': 'l00', '?loc-to': 'l10'\} \{('at', 'a0', 'l10')\}\\
            NoOp \{\} \{('in', 'p0', 't0')\}\\
            NoOp \{\} \{('in-city', 'l10', 'c1')\}\\
            NoOp \{\} \{('LOCATION', 'l01')\}\\
            DRIVE-TRUCK \{'?truck': 't1', '?loc-from': 'l10', '?loc-to': 'l11', '?city': 'c1'\} \{('at', 't1', 'l11')\}\\
            NoOp \{\} \{('AIRPORT', 'l00')\}\\
            NoOp \{\} \{('LOCATION', 'l11')\}\\
            NoOp \{\} \{('TRUCK', 't1')\}\\
            NoOp \{\} \{('in', 'p0', 'a0')\}\\
            UNLOAD-TRUCK \{'?obj': 'p0', '?truck': 't0', '?loc': 'l00'\} \{('at', 'p0', 'l00')\}\\
            NoOp \{\} \{('LOCATION', 'l10')\}\\
            NoOp \{\} \{('at', 'p0', 'l01')\}\\
            NoOp \{\} \{('CITY', 'c1')\}\\
            NoOp \{\} \{('at', 't0', 'l00')\}\\
            NoOp \{\} \{('TRUCK', 't0')\}\\
            NoOp \{\} \{('at', 't0', 'l01')\}\\
            NoOp \{\} \{('at', 't1', 'l10')\}\\
            NoOp \{\} \{('AIRPORT', 'l10')\}\\
            NoOp \{\} \{('OBJ', 'p0')\}\\
            \textbf{UNLOAD-AIRPLANE \{'?obj': 'p0', '?airplane': 'a0', '?loc': 'l10'\} \{('at', 'p0', 'l10')\}} \\
            NoOp \{\} \{('at', 't1', 'l11')\}\\
            DRIVE-TRUCK \{'?truck': 't0', '?loc-from': 'l01', '?loc-to': 'l00', '?city': 'c0'\} \{('at', 't0', 'l00')\}\\
            NoOp \{\} \{('at', 'a0', 'l10')\}\\
            NoOp \{\} \{('at', 'p0', 'l00')\}\\
            NoOp \{\} \{('at', 'a0', 'l00')\}\\
            NoOp \{\} \{('OBJ', 'p1')\}\\
            NoOp \{\} \{('in-city', 'l11', 'c1')\}\\
            NoOp \{\} \{('in-city', 'l00', 'c0')\}\\
            LOAD-AIRPLANE \{'?obj': 'p0', '?airplane': 'a0', '?loc': 'l00'\} \{('in', 'p0', 'a0')\}\\
        }}\end{center}
    LLM-BP-ACTION-SET
        \begin{center}
        \fcolorbox{black}{gray!10}{\parbox{.9\linewidth}{
            \textbf{UNLOAD-AIRPLANE \{'?obj': 'p0', '?airplane': 'a0', '?loc': 'l10'\} \{('at', 'p0', 'l10')\}} \\
            FLY-AIRPLANE \{'?airplane': 'a0', '?loc-from': 'l00', '?loc-to': 'l10'\} \{('at', 'a0', 'l10')\}\\
            LOAD-AIRPLANE \{'?obj': 'p0', '?airplane': 'a0', '?loc': 'l00'\} \{('in', 'p0', 'a0')\}\\
            DRIVE-TRUCK \{'?truck': 't1', '?loc-from': 'l10', '?loc-to': 'l11', '?city': 'c1'\} \{('at', 't1', 'l11')\}\\
            UNLOAD-TRUCK \{'?obj': 'p0', '?truck': 't0', '?loc': 'l00'\} \{('at', 'p0', 'l00')\}\\
            DRIVE-TRUCK \{'?truck': 't0', '?loc-from': 'l01', '?loc-to': 'l00', '?city': 'c0'\} \{('at', 't0', 'l00')\}\\
            NoOp \{\} \{('CITY', 'c0')\}\\
            NoOp \{\} \{('in-city', 'l01', 'c0')\}\\
            NoOp \{\} \{('AIRPLANE', 'a0')\}\\
            NoOp \{\} \{('LOCATION', 'l00')\}\\
            NoOp \{\} \{('in', 'p0', 't0')\}\\
            NoOp \{\} \{('in-city', 'l10', 'c1')\}\\
            NoOp \{\} \{('LOCATION', 'l01')\}\\
            NoOp \{\} \{('AIRPORT', 'l00')\}\\
            NoOp \{\} \{('LOCATION', 'l11')\}\\
            NoOp \{\} \{('TRUCK', 't1')\}\\
            NoOp \{\} \{('in', 'p0', 'a0')\}\\
            NoOp \{\} \{('LOCATION', 'l10')\}\\
            NoOp \{\} \{('at', 'p0', 'l01')\}\\
            NoOp \{\} \{('CITY', 'c1')\}\\
            NoOp \{\} \{('at', 't0', 'l00')\}\\
            NoOp \{\} \{('TRUCK', 't0')\}\\
            NoOp \{\} \{('at', 't0', 'l01')\}\\
            NoOp \{\} \{('at', 't1', 'l10')\}\\
            NoOp \{\} \{('AIRPORT', 'l10')\}\\
            NoOp \{\} \{('OBJ', 'p0')\}\\
            NoOp \{\} \{('at', 't1', 'l11')\}\\
            NoOp \{\} \{('at', 'a0', 'l10')\}\\
            NoOp \{\} \{('at', 'p0', 'l00')\}\\
            NoOp \{\} \{('at', 'a0', 'l00')\}\\
            NoOp \{\} \{('OBJ', 'p1')\}\\
            NoOp \{\} \{('in-city', 'l11', 'c1')\}\\
            NoOp \{\} \{('in-city', 'l00', 'c0')\}\\
        }}\end{center}

    \item Layer 8\\
    GP-ACTION-SET
        \begin{center}
        \fcolorbox{black}{gray!10}{\parbox{.9\linewidth}{
            UNLOAD-TRUCK \{'?obj': 'p0', '?truck': 't0', '?loc': 'l00'\} \{('at', 'p0', 'l00')\}\\
            NoOp \{\} \{('in', 'p0', 'a0')\}\\
            NoOp \{\} \{('CITY', 'c0')\}\\
            DRIVE-TRUCK \{'?truck': 't0', '?loc-from': 'l01', '?loc-to': 'l00', '?city': 'c0'\} \{('at', 't0', 'l00')\}\\
            NoOp \{\} \{('in-city', 'l01', 'c0')\}\\
            UNLOAD-AIRPLANE \{'?obj': 'p0', '?airplane': 'a0', '?loc': 'l00'\} \{('at', 'p0', 'l00')\}\\
            LOAD-TRUCK \{'?obj': 'p0', '?truck': 't0', '?loc': 'l00'\} \{('in', 'p0', 't0')\}\\
            NoOp \{\} \{('AIRPLANE', 'a0')\}\\
            NoOp \{\} \{('at', 't1', 'l10')\}\\
            UNLOAD-TRUCK \{'?obj': 'p0', '?truck': 't0', '?loc': 'l01'\} \{('at', 'p0', 'l01')\}\\
            NoOp \{\} \{('at', 'p0', 'l01')\}\\
            LOAD-TRUCK \{'?obj': 'p0', '?truck': 't0', '?loc': 'l01'\} \{('in', 'p0', 't0')\}\\
            NoOp \{\} \{('in-city', 'l00', 'c0')\}\\
            NoOp \{\} \{('TRUCK', 't0')\}\\
            NoOp \{\} \{('at', 't0', 'l00')\}\\
            NoOp \{\} \{('OBJ', 'p1')\}\\
            DRIVE-TRUCK \{'?truck': 't0', '?loc-from': 'l00', '?loc-to': 'l01', '?city': 'c0'\} \{('at', 't0', 'l01')\}\\
            NoOp \{\} \{('LOCATION', 'l00')\}\\
            NoOp \{\} \{('at', 'p0', 'l00')\}\\
            FLY-AIRPLANE \{'?airplane': 'a0', '?loc-from': 'l00', '?loc-to': 'l00'\} \{('at', 'a0', 'l00')\}\\
            LOAD-AIRPLANE \{'?obj': 'p0', '?airplane': 'a0', '?loc': 'l00'\} \{('in', 'p0', 'a0')\}\\
            NoOp \{\} \{('LOCATION', 'l01')\}\\
            NoOp \{\} \{('LOCATION', 'l10')\}\\
            \textbf{LOAD-TRUCK \{'?obj': 'p0', '?truck': 't1', '?loc': 'l10'\} \{('in', 'p0', 't1')\}} \\
        }}\end{center}
        \begin{center}
        \fcolorbox{black}{gray!10}{\parbox{.9\linewidth}{
            NoOp \{\} \{('at', 't1', 'l11')\}\\
            DRIVE-TRUCK \{'?truck': 't0', '?loc-from': 'l01', '?loc-to': 'l01', '?city': 'c0'\} \{('at', 't0', 'l01')\}\\
            DRIVE-TRUCK \{'?truck': 't1', '?loc-from': 'l11', '?loc-to': 'l10', '?city': 'c1'\} \{('at', 't1', 'l10')\}\\
            FLY-AIRPLANE \{'?airplane': 'a0', '?loc-from': 'l10', '?loc-to': 'l00'\} \{('at', 'a0', 'l00')\}\\
            NoOp \{\} \{('at', 'p0', 'l10')\}\\
            NoOp \{\} \{('at', 'a0', 'l00')\}\\
            FLY-AIRPLANE \{'?airplane': 'a0', '?loc-from': 'l00', '?loc-to': 'l10'\} \{('at', 'a0', 'l10')\}\\
            UNLOAD-AIRPLANE \{'?obj': 'p0', '?airplane': 'a0', '?loc': 'l10'\} \{('at', 'p0', 'l10')\}\\
            NoOp \{\} \{('CITY', 'c1')\}\\
            FLY-AIRPLANE \{'?airplane': 'a0', '?loc-from': 'l10', '?loc-to': 'l10'\} \{('at', 'a0', 'l10')\}\\
            NoOp \{\} \{('AIRPORT', 'l10')\}\\
            NoOp \{\} \{('in', 'p0', 't0')\}\\
            DRIVE-TRUCK \{'?truck': 't0', '?loc-from': 'l00', '?loc-to': 'l00', '?city': 'c0'\} \{('at', 't0', 'l00')\}\\
            DRIVE-TRUCK \{'?truck': 't1', '?loc-from': 'l10', '?loc-to': 'l10', '?city': 'c1'\} \{('at', 't1', 'l10')\}\\
            NoOp \{\} \{('OBJ', 'p0')\}\\
            NoOp \{\} \{('at', 'a0', 'l10')\}\\
            DRIVE-TRUCK \{'?truck': 't1', '?loc-from': 'l10', '?loc-to': 'l11', '?city': 'c1'\} \{('at', 't1', 'l11')\}\\
            NoOp \{\} \{('in-city', 'l10', 'c1')\}\\
            LOAD-AIRPLANE \{'?obj': 'p0', '?airplane': 'a0', '?loc': 'l10'\} \{('in', 'p0', 'a0')\}\\
            NoOp \{\} \{('at', 't0', 'l01')\}\\
            DRIVE-TRUCK \{'?truck': 't1', '?loc-from': 'l11', '?loc-to': 'l11', '?city': 'c1'\} \{('at', 't1', 'l11')\}\\
            NoOp \{\} \{('in-city', 'l11', 'c1')\}\\
            NoOp \{\} \{('AIRPORT', 'l00')\}\\
            NoOp \{\} \{('LOCATION', 'l11')\}\\
            NoOp \{\} \{('TRUCK', 't1')\}\\
        }}\end{center}
    LLM-EP-ACTION-SET
        \begin{center}
        \fcolorbox{black}{gray!10}{\parbox{.9\linewidth}{
            UNLOAD-TRUCK \{'?obj': 'p0', '?truck': 't0', '?loc': 'l00'\} \{('at', 'p0', 'l00')\}\\
            NoOp \{\} \{('in', 'p0', 'a0')\}\\
            NoOp \{\} \{('CITY', 'c0')\}\\
            DRIVE-TRUCK \{'?truck': 't0', '?loc-from': 'l01', '?loc-to': 'l00', '?city': 'c0'\} \{('at', 't0', 'l00')\}\\
            NoOp \{\} \{('in-city', 'l01', 'c0')\}\\
            NoOp \{\} \{('AIRPLANE', 'a0')\}\\
            NoOp \{\} \{('at', 't1', 'l10')\}\\
            NoOp \{\} \{('at', 'p0', 'l01')\}\\
            NoOp \{\} \{('in-city', 'l00', 'c0')\}\\
            NoOp \{\} \{('TRUCK', 't0')\}\\
            NoOp \{\} \{('at', 't0', 'l00')\}\\
            NoOp \{\} \{('OBJ', 'p1')\}\\
            NoOp \{\} \{('LOCATION', 'l00')\}\\
            NoOp \{\} \{('at', 'p0', 'l00')\}\\
            LOAD-AIRPLANE \{'?obj': 'p0', '?airplane': 'a0', '?loc': 'l00'\} \{('in', 'p0', 'a0')\}\\
            NoOp \{\} \{('LOCATION', 'l01')\}\\
            NoOp \{\} \{('LOCATION', 'l10')\}\\
            \textbf{LOAD-TRUCK \{'?obj': 'p0', '?truck': 't1', '?loc': 'l10'\} \{('in', 'p0', 't1')\}} \\
            NoOp \{\} \{('at', 't1', 'l11')\}\\
            NoOp \{\} \{('at', 'p0', 'l10')\}\\
            NoOp \{\} \{('at', 'a0', 'l00')\}\\
            FLY-AIRPLANE \{'?airplane': 'a0', '?loc-from': 'l00', '?loc-to': 'l10'\} \{('at', 'a0', 'l10')\}\\
            UNLOAD-AIRPLANE \{'?obj': 'p0', '?airplane': 'a0', '?loc': 'l10'\} \{('at', 'p0', 'l10')\}\\
            NoOp \{\} \{('CITY', 'c1')\}\\
            NoOp \{\} \{('AIRPORT', 'l10')\}\\
            NoOp \{\} \{('in', 'p0', 't0')\}\\
            NoOp \{\} \{('OBJ', 'p0')\}\\
            NoOp \{\} \{('at', 'a0', 'l10')\}\\
            DRIVE-TRUCK \{'?truck': 't1', '?loc-from': 'l10', '?loc-to': 'l11', '?city': 'c1'\} \{('at', 't1', 'l11')\}\\
            NoOp \{\} \{('in-city', 'l10', 'c1')\}\\
            NoOp \{\} \{('at', 't0', 'l01')\}\\
            NoOp \{\} \{('in-city', 'l11', 'c1')\}\\
            NoOp \{\} \{('AIRPORT', 'l00')\}\\
            NoOp \{\} \{('LOCATION', 'l11')\}\\
            NoOp \{\} \{('TRUCK', 't1')\}\\
        }}\end{center}
    LLM-BP-ACTION-SET
        \begin{center}
        \fcolorbox{black}{gray!10}{\parbox{.9\linewidth}{
            DRIVE-TRUCK \{'?truck': 't1', '?loc-from': 'l10', '?loc-to': 'l11', '?city': 'c1'\} \{('at', 't1', 'l11')\}\\
            \textbf{LOAD-TRUCK \{'?obj': 'p0', '?truck': 't1', '?loc': 'l10'\} \{('in', 'p0', 't1')\}} \\
            UNLOAD-TRUCK \{'?obj': 'p0', '?truck': 't0', '?loc': 'l00'\} \{('at', 'p0', 'l00')\}\\
            DRIVE-TRUCK \{'?truck': 't0', '?loc-from': 'l01', '?loc-to': 'l00', '?city': 'c0'\} \{('at', 't0', 'l00')\}\\
            LOAD-AIRPLANE \{'?obj': 'p0', '?airplane': 'a0', '?loc': 'l00'\} \{('in', 'p0', 'a0')\}\\
            FLY-AIRPLANE \{'?airplane': 'a0', '?loc-from': 'l00', '?loc-to': 'l10'\} \{('at', 'a0', 'l10')\}\\
            UNLOAD-AIRPLANE \{'?obj': 'p0', '?airplane': 'a0', '?loc': 'l10'\} \{('at', 'p0', 'l10')\}\\
            NoOp \{\} \{('in', 'p0', 'a0')\}\\
            NoOp \{\} \{('CITY', 'c0')\}\\
            NoOp \{\} \{('in-city', 'l01', 'c0')\}\\
            NoOp \{\} \{('AIRPLANE', 'a0')\}\\
            NoOp \{\} \{('at', 't1', 'l10')\}\\
            NoOp \{\} \{('at', 'p0', 'l01')\}\\
            NoOp \{\} \{('in-city', 'l00', 'c0')\}\\
            NoOp \{\} \{('TRUCK', 't0')\}\\
            NoOp \{\} \{('at', 't0', 'l00')\}\\
            NoOp \{\} \{('OBJ', 'p1')\}\\
            NoOp \{\} \{('LOCATION', 'l00')\}\\
            NoOp \{\} \{('at', 'p0', 'l00')\}\\
            NoOp \{\} \{('LOCATION', 'l01')\}\\
            NoOp \{\} \{('LOCATION', 'l10')\}\\
            NoOp \{\} \{('at', 't1', 'l11')\}\\
            NoOp \{\} \{('at', 'p0', 'l10')\}\\
            NoOp \{\} \{('at', 'a0', 'l00')\}\\
            NoOp \{\} \{('CITY', 'c1')\}\\
            NoOp \{\} \{('AIRPORT', 'l10')\}\\
            NoOp \{\} \{('in', 'p0', 't0')\}\\
            NoOp \{\} \{('OBJ', 'p0')\}\\
            NoOp \{\} \{('at', 'a0', 'l10')\}\\
            NoOp \{\} \{('in-city', 'l10', 'c1')\}\\
            NoOp \{\} \{('at', 't0', 'l01')\}\\
            NoOp \{\} \{('in-city', 'l11', 'c1')\}\\
            NoOp \{\} \{('AIRPORT', 'l00')\}\\
            NoOp \{\} \{('LOCATION', 'l11')\}\\
            NoOp \{\} \{('TRUCK', 't1')\}\\
        }}\end{center}

    \item Layer 9\\
    GP-ACTION-SET
        \begin{center}
        \fcolorbox{black}{gray!10}{\parbox{.9\linewidth}{
            LOAD-TRUCK \{'?obj': 'p0', '?truck': 't0', '?loc': 'l00'\} \{('in', 'p0', 't0')\}\\
            NoOp \{\} \{('at', 'p0', 'l00')\}\\
            LOAD-TRUCK \{'?obj': 'p0', '?truck': 't0', '?loc': 'l01'\} \{('in', 'p0', 't0')\}\\
            UNLOAD-AIRPLANE \{'?obj': 'p0', '?airplane': 'a0', '?loc': 'l00'\} \{('at', 'p0', 'l00')\}\\
            DRIVE-TRUCK \{'?truck': 't0', '?loc-from': 'l00', '?loc-to': 'l00', '?city': 'c0'\} \{('at', 't0', 'l00')\}\\
            DRIVE-TRUCK \{'?truck': 't0', '?loc-from': 'l01', '?loc-to': 'l01', '?city': 'c0'\} \{('at', 't0', 'l01')\}\\
            NoOp \{\} \{('in', 'p0', 't0')\}\\
            UNLOAD-AIRPLANE \{'?obj': 'p0', '?airplane': 'a0', '?loc': 'l10'\} \{('at', 'p0', 'l10')\}\\
            DRIVE-TRUCK \{'?truck': 't0', '?loc-from': 'l00', '?loc-to': 'l01', '?city': 'c0'\} \{('at', 't0', 'l01')\}\\
            NoOp \{\} \{('at', 'p0', 'l01')\}\\
            NoOp \{\} \{('OBJ', 'p1')\}\\
            NoOp \{\} \{('in-city', 'l11', 'c1')\}\\
            NoOp \{\} \{('at', 't0', 'l00')\}\\
            NoOp \{\} \{('AIRPLANE', 'a0')\}\\
            NoOp \{\} \{('in-city', 'l10', 'c1')\}\\
            DRIVE-TRUCK \{'?truck': 't0', '?loc-from': 'l01', '?loc-to': 'l00', '?city': 'c0'\} \{('at', 't0', 'l00')\}\\
            UNLOAD-TRUCK \{'?obj': 'p0', '?truck': 't0', '?loc': 'l01'\} \{('at', 'p0', 'l01')\}\\
            NoOp \{\} \{('LOCATION', 'l11')\}\\
            NoOp \{\} \{('AIRPORT', 'l10')\}\\
            UNLOAD-TRUCK \{'?obj': 'p0', '?truck': 't1', '?loc': 'l10'\} \{('at', 'p0', 'l10')\}\\
            \textbf{DRIVE-TRUCK \{'?truck': 't1', '?loc-from': 'l10', '?loc-to': 'l11', '?city': 'c1'\} \{('at', 't1', 'l11')\}} \\
        }}\end{center}
        \begin{center}
        \fcolorbox{black}{gray!10}{\parbox{.9\linewidth}{
            NoOp \{\} \{('TRUCK', 't0')\}\\
            NoOp \{\} \{('TRUCK', 't1')\}\\
            NoOp \{\} \{('at', 'p0', 'l10')\}\\
            FLY-AIRPLANE \{'?airplane': 'a0', '?loc-from': 'l00', '?loc-to': 'l10'\} \{('at', 'a0', 'l10')\}\\
            NoOp \{\} \{('at', 't1', 'l11')\}\\
            NoOp \{\} \{('in-city', 'l01', 'c0')\}\\
            LOAD-AIRPLANE \{'?obj': 'p0', '?airplane': 'a0', '?loc': 'l00'\} \{('in', 'p0', 'a0')\}\\
            DRIVE-TRUCK \{'?truck': 't1', '?loc-from': 'l10', '?loc-to': 'l10', '?city': 'c1'\} \{('at', 't1', 'l10')\}\\
            NoOp \{\} \{('at', 't1', 'l10')\}\\
            NoOp \{\} \{('in', 'p0', 't1')\}\\
            NoOp \{\} \{('in-city', 'l00', 'c0')\}\\
            NoOp \{\} \{('CITY', 'c1')\}\\
            NoOp \{\} \{('LOCATION', 'l10')\}\\
            NoOp \{\} \{('in', 'p0', 'a0')\}\\
            NoOp \{\} \{('OBJ', 'p0')\}\\
            LOAD-TRUCK \{'?obj': 'p0', '?truck': 't1', '?loc': 'l10'\} \{('in', 'p0', 't1')\}\\
            NoOp \{\} \{('LOCATION', 'l01')\}\\
            FLY-AIRPLANE \{'?airplane': 'a0', '?loc-from': 'l00', '?loc-to': 'l00'\} \{('at', 'a0', 'l00')\}\\
            NoOp \{\} \{('LOCATION', 'l00')\}\\
            FLY-AIRPLANE \{'?airplane': 'a0', '?loc-from': 'l10', '?loc-to': 'l10'\} \{('at', 'a0', 'l10')\}\\
            DRIVE-TRUCK \{'?truck': 't1', '?loc-from': 'l11', '?loc-to': 'l11', '?city': 'c1'\} \{('at', 't1', 'l11')\}\\
            UNLOAD-TRUCK \{'?obj': 'p0', '?truck': 't0', '?loc': 'l00'\} \{('at', 'p0', 'l00')\}\\
            DRIVE-TRUCK \{'?truck': 't1', '?loc-from': 'l11', '?loc-to': 'l10', '?city': 'c1'\} \{('at', 't1', 'l10')\}\\
            NoOp \{\} \{('AIRPORT', 'l00')\}\\
            NoOp \{\} \{('at', 't0', 'l01')\}\\
            FLY-AIRPLANE \{'?airplane': 'a0', '?loc-from': 'l10', '?loc-to': 'l00'\} \{('at', 'a0', 'l00')\}\\
            NoOp \{\} \{('CITY', 'c0')\}\\
            LOAD-AIRPLANE \{'?obj': 'p0', '?airplane': 'a0', '?loc': 'l10'\} \{('in', 'p0', 'a0')\}\\
            NoOp \{\} \{('at', 'a0', 'l00')\}\\
            NoOp \{\} \{('at', 'a0', 'l10')\}\\
        }}\end{center}
    LLM-EP-ACTION-SET
        \begin{center}
        \fcolorbox{black}{gray!10}{\parbox{.9\linewidth}{
            NoOp \{\} \{('at', 'p0', 'l00')\}\\
            NoOp \{\} \{('in', 'p0', 't0')\}\\
            UNLOAD-AIRPLANE \{'?obj': 'p0', '?airplane': 'a0', '?loc': 'l10'\} \{('at', 'p0', 'l10')\}\\
            NoOp \{\} \{('at', 'p0', 'l01')\}\\
            NoOp \{\} \{('OBJ', 'p1')\}\\
            NoOp \{\} \{('in-city', 'l11', 'c1')\}\\
            NoOp \{\} \{('at', 't0', 'l00')\}\\
            NoOp \{\} \{('AIRPLANE', 'a0')\}\\
            NoOp \{\} \{('in-city', 'l10', 'c1')\}\\
            DRIVE-TRUCK \{'?truck': 't0', '?loc-from': 'l01', '?loc-to': 'l00', '?city': 'c0'\} \{('at', 't0', 'l00')\}\\
            NoOp \{\} \{('LOCATION', 'l11')\}\\
            NoOp \{\} \{('AIRPORT', 'l10')\}\\
            \textbf{DRIVE-TRUCK \{'?truck': 't1', '?loc-from': 'l10', '?loc-to': 'l11', '?city': 'c1'\} \{('at', 't1', 'l11')\}} \\
            NoOp \{\} \{('TRUCK', 't0')\}\\
            NoOp \{\} \{('TRUCK', 't1')\}\\
            NoOp \{\} \{('at', 'p0', 'l10')\}\\
            FLY-AIRPLANE \{'?airplane': 'a0', '?loc-from': 'l00', '?loc-to': 'l10'\} \{('at', 'a0', 'l10')\}\\
            NoOp \{\} \{('at', 't1', 'l11')\}\\
            NoOp \{\} \{('in-city', 'l01', 'c0')\}\\
            LOAD-AIRPLANE \{'?obj': 'p0', '?airplane': 'a0', '?loc': 'l00'\} \{('in', 'p0', 'a0')\}\\
            NoOp \{\} \{('at', 't1', 'l10')\}\\
            NoOp \{\} \{('in', 'p0', 't1')\}\\
            NoOp \{\} \{('in-city', 'l00', 'c0')\}\\
            NoOp \{\} \{('CITY', 'c1')\}\\
            NoOp \{\} \{('LOCATION', 'l10')\}\\
            NoOp \{\} \{('in', 'p0', 'a0')\}\\
            NoOp \{\} \{('OBJ', 'p0')\}\\
            NoOp \{\} \{('LOCATION', 'l01')\}\\
            NoOp \{\} \{('LOCATION', 'l00')\}\\
            UNLOAD-TRUCK \{'?obj': 'p0', '?truck': 't0', '?loc': 'l00'\} \{('at', 'p0', 'l00')\}\\
            NoOp \{\} \{('AIRPORT', 'l00')\}\\
            NoOp \{\} \{('at', 't0', 'l01')\}\\
            NoOp \{\} \{('CITY', 'c0')\}\\
            NoOp \{\} \{('at', 'a0', 'l00')\}\\
            NoOp \{\} \{('at', 'a0', 'l10')\}\\
        }}\end{center}
    LLM-BP-ACTION-SET
        \begin{center}
        \fcolorbox{black}{gray!10}{\parbox{.9\linewidth}{
            \textbf{DRIVE-TRUCK \{'?truck': 't1', '?loc-from': 'l10', '?loc-to': 'l11', '?city': 'c1'\} \{('at', 't1', 'l11')\}} \\
            UNLOAD-AIRPLANE \{'?obj': 'p0', '?airplane': 'a0', '?loc': 'l10'\} \{('at', 'p0', 'l10')\}\\
            DRIVE-TRUCK \{'?truck': 't0', '?loc-from': 'l01', '?loc-to': 'l00', '?city': 'c0'\} \{('at', 't0', 'l00')\}\\
            FLY-AIRPLANE \{'?airplane': 'a0', '?loc-from': 'l00', '?loc-to': 'l10'\} \{('at', 'a0', 'l10')\}\\
            LOAD-AIRPLANE \{'?obj': 'p0', '?airplane': 'a0', '?loc': 'l00'\} \{('in', 'p0', 'a0')\}\\
            UNLOAD-TRUCK \{'?obj': 'p0', '?truck': 't0', '?loc': 'l00'\} \{('at', 'p0', 'l00')\}\\
            NoOp \{\} \{('at', 'p0', 'l00')\}\\
            NoOp \{\} \{('in', 'p0', 't0')\}\\
            NoOp \{\} \{('at', 'p0', 'l01')\}\\
            NoOp \{\} \{('OBJ', 'p1')\}\\
            NoOp \{\} \{('in-city', 'l11', 'c1')\}\\
            NoOp \{\} \{('at', 't0', 'l00')\}\\
            NoOp \{\} \{('AIRPLANE', 'a0')\}\\
            NoOp \{\} \{('in-city', 'l10', 'c1')\}\\
            NoOp \{\} \{('LOCATION', 'l11')\}\\
            NoOp \{\} \{('AIRPORT', 'l10')\}\\
            NoOp \{\} \{('TRUCK', 't0')\}\\
            NoOp \{\} \{('TRUCK', 't1')\}\\
            NoOp \{\} \{('at', 'p0', 'l10')\}\\
            NoOp \{\} \{('at', 't1', 'l11')\}\\
            NoOp \{\} \{('in-city', 'l01', 'c0')\}\\
            NoOp \{\} \{('at', 't1', 'l10')\}\\
            NoOp \{\} \{('in', 'p0', 't1')\}\\
            NoOp \{\} \{('in-city', 'l00', 'c0')\}\\
            NoOp \{\} \{('CITY', 'c1')\}\\
            NoOp \{\} \{('LOCATION', 'l10')\}\\
            NoOp \{\} \{('in', 'p0', 'a0')\}\\
            NoOp \{\} \{('OBJ', 'p0')\}\\
            NoOp \{\} \{('LOCATION', 'l01')\}\\
            NoOp \{\} \{('LOCATION', 'l00')\}\\
            NoOp \{\} \{('AIRPORT', 'l00')\}\\
            NoOp \{\} \{('at', 't0', 'l01')\}\\
            NoOp \{\} \{('CITY', 'c0')\}\\
            NoOp \{\} \{('at', 'a0', 'l00')\}\\
            NoOp \{\} \{('at', 'a0', 'l10')\}\\
        }}\end{center}
        % \newpage

    \item Layer 10\\
    GP-ACTION-SET
        \begin{center}
        \fcolorbox{black}{gray!10}{\parbox{.9\linewidth}{
            NoOp \{\} \{('at', 'a0', 'l00')\}\\
            NoOp \{\} \{('at', 't1', 'l10')\}\\
            DRIVE-TRUCK \{'?truck': 't0', '?loc-from': 'l00', '?loc-to': 'l01', '?city': 'c0'\} \{('at', 't0', 'l01')\}\\
            DRIVE-TRUCK \{'?truck': 't1', '?loc-from': 'l11', '?loc-to': 'l11', '?city': 'c1'\} \{('at', 't1', 'l11')\}\\
            NoOp \{\} \{('OBJ', 'p0')\}\\
            LOAD-AIRPLANE \{'?obj': 'p0', '?airplane': 'a0', '?loc': 'l00'\} \{('in', 'p0', 'a0')\}\\
            NoOp \{\} \{('in', 'p0', 'a0')\}\\
            DRIVE-TRUCK \{'?truck': 't1', '?loc-from': 'l10', '?loc-to': 'l11', '?city': 'c1'\} \{('at', 't1', 'l11')\}\\
            NoOp \{\} \{('AIRPORT', 'l00')\}\\
            DRIVE-TRUCK \{'?truck': 't0', '?loc-from': 'l01', '?loc-to': 'l00', '?city': 'c0'\} \{('at', 't0', 'l00')\}\\
            NoOp \{\} \{('in-city', 'l11', 'c1')\}\\
            LOAD-TRUCK \{'?obj': 'p0', '?truck': 't0', '?loc': 'l01'\} \{('in', 'p0', 't0')\}\\
            NoOp \{\} \{('at', 't1', 'l11')\}\\
            NoOp \{\} \{('at', 'a0', 'l10')\}\\
            \textbf{UNLOAD-TRUCK \{'?obj': 'p0', '?truck': 't1', '?loc': 'l11'\} \{('at', 'p0', 'l11')\}} \\
        }}\end{center}
        \begin{center}
        \fcolorbox{black}{gray!10}{\parbox{.9\linewidth}{
            NoOp \{\} \{('at', 'p0', 'l01')\}\\
            NoOp \{\} \{('in-city', 'l10', 'c1')\}\\
            DRIVE-TRUCK \{'?truck': 't0', '?loc-from': 'l00', '?loc-to': 'l00', '?city': 'c0'\} \{('at', 't0', 'l00')\}\\
            DRIVE-TRUCK \{'?truck': 't1', '?loc-from': 'l11', '?loc-to': 'l10', '?city': 'c1'\} \{('at', 't1', 'l10')\}\\
            FLY-AIRPLANE \{'?airplane': 'a0', '?loc-from': 'l00', '?loc-to': 'l00'\} \{('at', 'a0', 'l00')\}\\
            FLY-AIRPLANE \{'?airplane': 'a0', '?loc-from': 'l00', '?loc-to': 'l10'\} \{('at', 'a0', 'l10')\}\\
            UNLOAD-AIRPLANE \{'?obj': 'p0', '?airplane': 'a0', '?loc': 'l10'\} \{('at', 'p0', 'l10')\}\\
            NoOp \{\} \{('in-city', 'l00', 'c0')\}\\
            NoOp \{\} \{('AIRPORT', 'l10')\}\\
            NoOp \{\} \{('AIRPLANE', 'a0')\}\\
            NoOp \{\} \{('in', 'p0', 't1')\}\\
            NoOp \{\} \{('at', 'p0', 'l10')\}\\
            NoOp \{\} \{('OBJ', 'p1')\}\\
            NoOp \{\} \{('LOCATION', 'l10')\}\\
            UNLOAD-TRUCK \{'?obj': 'p0', '?truck': 't0', '?loc': 'l00'\} \{('at', 'p0', 'l00')\}\\
            NoOp \{\} \{('at', 'p0', 'l00')\}\\
            UNLOAD-TRUCK \{'?obj': 'p0', '?truck': 't0', '?loc': 'l01'\} \{('at', 'p0', 'l01')\}\\
            NoOp \{\} \{('CITY', 'c1')\}\\
            LOAD-TRUCK \{'?obj': 'p0', '?truck': 't0', '?loc': 'l00'\} \{('in', 'p0', 't0')\}\\
            NoOp \{\} \{('TRUCK', 't0')\}\\
            UNLOAD-TRUCK \{'?obj': 'p0', '?truck': 't1', '?loc': 'l10'\} \{('at', 'p0', 'l10')\}\\
            NoOp \{\} \{('at', 't0', 'l01')\}\\
            NoOp \{\} \{('LOCATION', 'l11')\}\\
            FLY-AIRPLANE \{'?airplane': 'a0', '?loc-from': 'l10', '?loc-to': 'l10'\} \{('at', 'a0', 'l10')\}\\
            NoOp \{\} \{('at', 't0', 'l00')\}\\
            FLY-AIRPLANE \{'?airplane': 'a0', '?loc-from': 'l10', '?loc-to': 'l00'\} \{('at', 'a0', 'l00')\}\\
            DRIVE-TRUCK \{'?truck': 't1', '?loc-from': 'l10', '?loc-to': 'l10', '?city': 'c1'\} \{('at', 't1', 'l10')\}\\
            NoOp \{\} \{('LOCATION', 'l01')\}\\
            DRIVE-TRUCK \{'?truck': 't0', '?loc-from': 'l01', '?loc-to': 'l01', '?city': 'c0'\} \{('at', 't0', 'l01')\}\\
            NoOp \{\} \{('CITY', 'c0')\}\\
            NoOp \{\} \{('LOCATION', 'l00')\}\\
            UNLOAD-AIRPLANE \{'?obj': 'p0', '?airplane': 'a0', '?loc': 'l00'\} \{('at', 'p0', 'l00')\}\\
            NoOp \{\} \{('in-city', 'l01', 'c0')\}\\
            NoOp \{\} \{('in', 'p0', 't0')\}\\
            LOAD-AIRPLANE \{'?obj': 'p0', '?airplane': 'a0', '?loc': 'l10'\} \{('in', 'p0', 'a0')\}\\
            NoOp \{\} \{('TRUCK', 't1')\}\\
            LOAD-TRUCK \{'?obj': 'p0', '?truck': 't1', '?loc': 'l10'\} \{('in', 'p0', 't1')\}\\
        }}\end{center}
        % \newpage
        
    LLM-EP-ACTION-SET
        \begin{center}
        \fcolorbox{black}{gray!10}{\parbox{.9\linewidth}{
            NoOp \{\} \{('at', 'a0', 'l00')\}\\
            NoOp \{\} \{('at', 't1', 'l10')\}\\
            NoOp \{\} \{('OBJ', 'p0')\}\\
            NoOp \{\} \{('in', 'p0', 'a0')\}\\
            DRIVE-TRUCK \{'?truck': 't1', '?loc-from': 'l10', '?loc-to': 'l11', '?city': 'c1'\} \{('at', 't1', 'l11')\}\\
            NoOp \{\} \{('AIRPORT', 'l00')\}\\
            NoOp \{\} \{('in-city', 'l11', 'c1')\}\\
            NoOp \{\} \{('at', 't1', 'l11')\}\\
            NoOp \{\} \{('at', 'a0', 'l10')\}\\
            \textbf{UNLOAD-TRUCK \{'?obj': 'p0', '?truck': 't1', '?loc': 'l11'\} \{('at', 'p0', 'l11')\}} \\
            NoOp \{\} \{('at', 'p0', 'l01')\}\\
            NoOp \{\} \{('in-city', 'l10', 'c1')\}\\
            NoOp \{\} \{('in-city', 'l00', 'c0')\}\\
            NoOp \{\} \{('AIRPORT', 'l10')\}\\
            NoOp \{\} \{('AIRPLANE', 'a0')\}\\
            NoOp \{\} \{('in', 'p0', 't1')\}\\
            NoOp \{\} \{('at', 'p0', 'l10')\}\\
            NoOp \{\} \{('OBJ', 'p1')\}\\
            NoOp \{\} \{('LOCATION', 'l10')\}\\
            NoOp \{\} \{('at', 'p0', 'l00')\}\\
            NoOp \{\} \{('CITY', 'c1')\}\\
            NoOp \{\} \{('TRUCK', 't0')\}\\
            NoOp \{\} \{('at', 't0', 'l01')\}\\
            NoOp \{\} \{('LOCATION', 'l11')\}\\
            NoOp \{\} \{('at', 't0', 'l00')\}\\
            NoOp \{\} \{('LOCATION', 'l01')\}\\
            NoOp \{\} \{('CITY', 'c0')\}\\
            NoOp \{\} \{('LOCATION', 'l00')\}\\
            NoOp \{\} \{('in-city', 'l01', 'c0')\}\\
            NoOp \{\} \{('in', 'p0', 't0')\}\\
            NoOp \{\} \{('TRUCK', 't1')\}\\
            LOAD-TRUCK \{'?obj': 'p0', '?truck': 't1', '?loc': 'l10'\} \{('in', 'p0', 't1')\}\\
        }}\end{center}
        % \newpage
        
    LLM-BP-ACTION-SET
        \begin{center}
        \fcolorbox{black}{gray!10}{\parbox{.9\linewidth}{
            DRIVE-TRUCK \{'?truck': 't1', '?loc-from': 'l10', '?loc-to': 'l11', '?city': 'c1'\} \{('at', 't1', 'l11')\}\\
            \textbf{UNLOAD-TRUCK \{'?obj': 'p0', '?truck': 't1', '?loc': 'l11'\} \{('at', 'p0', 'l11')\}} \\
            LOAD-TRUCK \{'?obj': 'p0', '?truck': 't1', '?loc': 'l10'\} \{('in', 'p0', 't1')\}\\
            NoOp \{\} \{('at', 'a0', 'l00')\}\\
            NoOp \{\} \{('at', 't1', 'l10')\}\\
            NoOp \{\} \{('OBJ', 'p0')\}\\
            NoOp \{\} \{('in', 'p0', 'a0')\}\\
            NoOp \{\} \{('AIRPORT', 'l00')\}\\
            NoOp \{\} \{('in-city', 'l11', 'c1')\}\\
            NoOp \{\} \{('at', 't1', 'l11')\}\\
            NoOp \{\} \{('at', 'a0', 'l10')\}\\
            NoOp \{\} \{('at', 'p0', 'l01')\}\\
            NoOp \{\} \{('in-city', 'l10', 'c1')\}\\
            NoOp \{\} \{('in-city', 'l00', 'c0')\}\\
            NoOp \{\} \{('AIRPORT', 'l10')\}\\
            NoOp \{\} \{('AIRPLANE', 'a0')\}\\
            NoOp \{\} \{('in', 'p0', 't1')\}\\
            NoOp \{\} \{('at', 'p0', 'l10')\}\\
            NoOp \{\} \{('OBJ', 'p1')\}\\
            NoOp \{\} \{('LOCATION', 'l10')\}\\
            NoOp \{\} \{('at', 'p0', 'l00')\}\\
            NoOp \{\} \{('CITY', 'c1')\}\\
            NoOp \{\} \{('TRUCK', 't0')\}\\
            NoOp \{\} \{('at', 't0', 'l01')\}\\
            NoOp \{\} \{('LOCATION', 'l11')\}\\
            NoOp \{\} \{('at', 't0', 'l00')\}\\
            NoOp \{\} \{('LOCATION', 'l01')\}\\
            NoOp \{\} \{('CITY', 'c0')\}\\
            NoOp \{\} \{('LOCATION', 'l00')\}\\
            NoOp \{\} \{('in-city', 'l01', 'c0')\}\\
            NoOp \{\} \{('in', 'p0', 't0')\}\\
            NoOp \{\} \{('TRUCK', 't1')\}\\
        }}\end{center}
\end{enumerate}

%%%%%%%%%%%%%%%%%%%%%%%%%%%%%%%%%%%%%%%%%%%%%%%%%%%%%%%%%%%%
\ignore{
\newpage
\section*{NeurIPS Paper Checklist}

%%% BEGIN INSTRUCTIONS %%%
The checklist is designed to encourage best practices for responsible machine learning research, addressing issues of reproducibility, transparency, research ethics, and societal impact. Do not remove the checklist: {\bf The papers not including the checklist will be desk rejected.} The checklist should follow the references and follow the (optional) supplemental material.  The checklist does NOT count towards the page
limit. 

Please read the checklist guidelines carefully for information on how to answer these questions. For each question in the checklist:
\begin{itemize}
    \item You should answer \answerYes{}, \answerNo{}, or \answerNA{}.
    \item \answerNA{} means either that the question is Not Applicable for that particular paper or the relevant information is Not Available.
    \item Please provide a short (1–2 sentence) justification right after your answer (even for NA). 
   % \item {\bf The papers not including the checklist will be desk rejected.}
\end{itemize}

{\bf The checklist answers are an integral part of your paper submission.} They are visible to the reviewers, area chairs, senior area chairs, and ethics reviewers. You will be asked to also include it (after eventual revisions) with the final version of your paper, and its final version will be published with the paper.

The reviewers of your paper will be asked to use the checklist as one of the factors in their evaluation. While "\answerYes{}" is generally preferable to "\answerNo{}", it is perfectly acceptable to answer "\answerNo{}" provided a proper justification is given (e.g., "error bars are not reported because it would be too computationally expensive" or "we were unable to find the license for the dataset we used"). In general, answering "\answerNo{}" or "\answerNA{}" is not grounds for rejection. While the questions are phrased in a binary way, we acknowledge that the true answer is often more nuanced, so please just use your best judgment and write a justification to elaborate. All supporting evidence can appear either in the main paper or the supplemental material, provided in appendix. If you answer \answerYes{} to a question, in the justification please point to the section(s) where related material for the question can be found.

IMPORTANT, please:
\begin{itemize}
    \item {\bf Delete this instruction block, but keep the section heading ``NeurIPS paper checklist"},
    \item  {\bf Keep the checklist subsection headings, questions/answers and guidelines below.}
    \item {\bf Do not modify the questions and only use the provided macros for your answers}.
\end{itemize}

%%% END INSTRUCTIONS %%%

\begin{enumerate}

\item {\bf Claims}
    \item[] Question: Do the main claims made in the abstract and introduction accurately reflect the paper's contributions and scope?
    \item[] Answer: \answerYes{} % Replace by \answerYes{}, \answerNo{}, or \answerNA{}.
    \item[] Justification: %\justificationTODO{}

\item {\bf Limitations}
    \item[] Question: Does the paper discuss the limitations of the work performed by the authors?
    \item[] Answer: \answerYes{} % Replace by \answerYes{}, \answerNo{}, or \answerNA{}.
    \item[] Justification: %\justificationTODO{}

\item {\bf Theory Assumptions and Proofs}
    \item[] Question: For each theoretical result, does the paper provide the full set of assumptions and a complete (and correct) proof?
    \item[] Answer: \answerYes{} % Replace by \answerYes{}, \answerNo{}, or \answerNA{}.
    \item[] Justification: %\justificationTODO{}

    \item {\bf Experimental Result Reproducibility}
    \item[] Question: Does the paper fully disclose all the information needed to reproduce the main experimental results of the paper to the extent that it affects the main claims and/or conclusions of the paper (regardless of whether the code and data are provided or not)?
    \item[] Answer: \answerYes{} % Replace by \answerYes{}, \answerNo{}, or \answerNA{}.
    \item[] Justification: %\justificationTODO{}

\item {\bf Open access to data and code}
    \item[] Question: Does the paper provide open access to the data and code, with sufficient instructions to faithfully reproduce the main experimental results, as described in supplemental material?
    \item[] Answer: \answerYes{} % Replace by \answerYes{}, \answerNo{}, or \answerNA{}.
    \item[] Justification: %\justificationTODO{}

\item {\bf Experimental Setting/Details}
    \item[] Question: Does the paper specify all the training and test details (e.g., data splits, hyperparameters, how they were chosen, type of optimizer, etc.) necessary to understand the results?
    \item[] Answer: \answerYes{} % Replace by \answerYes{}, \answerNo{}, or \answerNA{}.
    \item[] Justification: %\justificationTODO{}

\item {\bf Experiment Statistical Significance}
    \item[] Question: Does the paper report error bars suitably and correctly defined or other appropriate information about the statistical significance of the experiments?
    \item[] Answer: \answerYes{} % Replace by \answerYes{}, \answerNo{}, or \answerNA{}.
    \item[] Justification: %\justificationTODO{}

\item {\bf Experiments Compute Resources}
    \item[] Question: For each experiment, does the paper provide sufficient information on the computer resources (type of compute workers, memory, time of execution) needed to reproduce the experiments?
    \item[] Answer: \answerYes{} % Replace by \answerYes{}, \answerNo{}, or \answerNA{}.
    \item[] Justification: %\justificationTODO{}
    
\item {\bf Code Of Ethics}
    \item[] Question: Does the research conducted in the paper conform, in every respect, with the NeurIPS Code of Ethics \url{https://neurips.cc/public/EthicsGuidelines}?
    \item[] Answer: \answerYes{} % Replace by \answerYes{}, \answerNo{}, or \answerNA{}.
    \item[] Justification: %\justificationTODO{}

\item {\bf Broader Impacts}
    \item[] Question: Does the paper discuss both potential positive societal impacts and negative societal impacts of the work performed?
    \item[] Answer: \answerYes{} % Replace by \answerYes{}, \answerNo{}, or \answerNA{}.
    \item[] Justification: %\justificationTODO{}
    
\item {\bf Safeguards}
    \item[] Question: Does the paper describe safeguards that have been put in place for responsible release of data or models that have a high risk for misuse (e.g., pretrained language models, image generators, or scraped datasets)?
    \item[] Answer: \answerYes{} % Replace by \answerYes{}, \answerNo{}, or \answerNA{}.
    \item[] Justification: %\justificationTODO{}

\item {\bf Licenses for existing assets}
    \item[] Question: Are the creators or original owners of assets (e.g., code, data, models), used in the paper, properly credited and are the license and terms of use explicitly mentioned and properly respected?
    \item[] Answer: \answerYes{} % Replace by \answerYes{}, \answerNo{}, or \answerNA{}.
    \item[] Justification: %\justificationTODO{}

\item {\bf New Assets}
    \item[] Question: Are new assets introduced in the paper well documented and is the documentation provided alongside the assets?
    \item[] Answer: \answerYes{} % Replace by \answerYes{}, \answerNo{}, or \answerNA{}.
    \item[] Justification: %\justificationTODO{}

\item {\bf Crowdsourcing and Research with Human Subjects}
    \item[] Question: For crowdsourcing experiments and research with human subjects, does the paper include the full text of instructions given to participants and screenshots, if applicable, as well as details about compensation (if any)? 
    \item[] Answer: \answerYes{} % Replace by \answerYes{}, \answerNo{}, or \answerNA{}.
    \item[] Justification: %\justificationTODO{}

\item {\bf Institutional Review Board (IRB) Approvals or Equivalent for Research with Human Subjects}
    \item[] Question: Does the paper describe potential risks incurred by study participants, whether such risks were disclosed to the subjects, and whether Institutional Review Board (IRB) approvals (or an equivalent approval/review based on the requirements of your country or institution) were obtained?
    \item[] Answer: \answerYes{} % Replace by \answerYes{}, \answerNo{}, or \answerNA{}.
    \item[] Justification: %\justificationTODO{}

\end{enumerate}
}

\end{document}